\documentclass{article}

\PassOptionsToPackage{round, compress}{natbib}
 \usepackage[main, final]{neurips_2026}

\usepackage[utf8]{inputenc} 
\usepackage[T1]{fontenc}    
\usepackage{hyperref}       
\usepackage{url}            
\usepackage{booktabs}       
\usepackage{amsfonts}       
\usepackage{nicefrac}       
\usepackage{microtype}      
\usepackage{xcolor}         

\usepackage{amsmath}
\usepackage{amssymb}
\usepackage{mathtools}
\usepackage{amsthm}

\usepackage{etoc}
\usepackage{graphicx}
\usepackage{subcaption}
\usepackage{multirow}
\usepackage{hyperref}
\usepackage{colortbl}
\usepackage{float}
\usepackage{enumerate}   
\usepackage{algorithm}
\usepackage{placeins}
\usepackage{algpseudocode}
\usepackage[table]{xcolor}   
\usepackage{colortbl}
\usepackage{enumitem}   
\usepackage{amsmath,amssymb,tikz}
\usetikzlibrary{arrows.meta}

\definecolor{ink}{HTML}{243447}
\definecolor{TornadoBlue}{HTML}{3969B0}
\definecolor{muted}{HTML}{64748B}
\definecolor{myGray}{gray}{0.88}

\theoremstyle{plain}
\newtheorem{theorem}{Theorem}[section]
\newtheorem{proposition}[theorem]{Proposition}

\newtheorem{corollary}[theorem]{Corollary}

\theoremstyle{definition}

\theoremstyle{remark}
\newtheorem{remark}[theorem]{Remark}

\newcommand{\merr}[2]{#1\,{\scriptsize$\pm$\,#2}}

\definecolor{TornadoBlue}{HTML}{3969B0}
\definecolor{TornadoAccent}{HTML}{469B69} 

\hypersetup{
    colorlinks=true,
    citecolor=TornadoBlue,
    linkcolor=TornadoAccent,
    urlcolor=TornadoBlue
}

\title{TORNADO: Adaptive Latent Stochastic Transport for Calibrated Probabilistic PDE Forecasting}

\author{
\begin{tabular}[t]{c}
\textbf{Jorge Mifsut Benet$^{1}$\thanks{Corresponding author: \texttt{mifsutbenet[at]isir.upmc.fr}},\;
Armand Kassa\"i Koupa\"i$^{1}$,\;
Ramon Daniel Regueiro-Espino$^{1}$,}\\
\textbf{Nicolas Baskiotis$^{1}$, \; Patrick Gallinari$^{1,2}$} \\
\normalfont $^{1}$ Sorbonne Universit\'e, CNRS, ISIR, 75005 Paris, France \\
\normalfont $^{2}$ Criteo AI Lab, Paris, France
\end{tabular}
}

\begin{document}
\etocdepthtag.toc{mtmain}

\maketitle

\begin{abstract}

Probabilistic neural surrogates for PDEs enable ensemble forecasting, but often exhibit miscalibration when used for uncertainty quantification, especially in compute-constrained, few-step inference regimes. Existing generative approaches typically rely on fixed global stochasticity, which cannot adapt predictive uncertainty to the local dynamical state. 
We propose TORNADO, a latent probabilistic forecasting framework that formulates next-step prediction as stochastic transport between VAE encoder posteriors.  Unlike existing approaches with globally prescribed stochasticity, TORNADO learns a SDE with state-dependent diffusion, enabling the predictive uncertainty to adapt to the local latent dynamics. The model is trained using analytic conditional endpoint targets derived from a Gaussian interpolant, enabling simulation-free learning. An additional energy distance regularization term improves distributional alignment. Empirically, TORNADO improves uncertainty calibration over representative generative baselines, including stochastic interpolants, diffusion models, and flow matching, while maintaining comparable predictive accuracy. Our results show that adapting stochasticity to the latent state, rather than prescribing it globally, is an effective mechanism for improving calibration in practical neural PDE forecasting settings. Project page: \url{https://tornado-sde.github.io}.

\end{abstract}

\section{Introduction}
Neural surrogates for partial differential equations~(PDEs) approximate solution operators, enabling fast data-driven prediction of high-dimensional spatiotemporal fields \citep{Lu2019, kochkov}. Often,  solution forecasting is performed in a reduced latent space through an encode–evolve–decode pipeline, which lowers computational cost \citep{serrano2024aroma, li2025generativelatentneuralpdesolver, wu2024uncertainty}. However, their reliability can deteriorate under distribution shift \citep{koupai2024geps}, they may fail to respect known physical constraints, and small one-step errors can compound in autoregressive rollout \citep{KOVACHKI2024operator, lippe2023pderefiner}.

These effects are particularly pronounced in chaotic systems, where small perturbations to a state can produce exponentially diverging trajectories. Deterministic models trained with squared-error losses therefore often degrade over rollout, since pointwise accuracy does not ensure that the forecast preserves the statistical structure of the dynamics \citep{jiang2024training}. This motivates a conditional probabilistic formulation: given the current state, the model should represent a distribution over plausible future states rather than a single trajectory \citep{chen2024probabilisticforecastingstochasticinterpolants}. Probabilistic forecasting also provides a basis for uncertainty quantification \citep{bülte2025probabilisticneuraloperatorsfunctional}, since predictive spread can be used to assess forecast confidence. Crucially, probabilistic forecasts are only informative when their uncertainty is well calibrated, i.e., when predictive spread matches actual forecast uncertainty.

Generative models provide a natural route to modeling high-dimensional forecast distributions, motivating the use of diffusion models, flow matching, and stochastic interpolants for probabilistic PDE forecasting and operator learning \citep{Price2024weather,Li2025LatentPDEFlowMatching,zhou2025reframinggenerativemodelsphysical,hou2025cfolearningcontinuoustimepde}. However, sampling along a generative bridge requires repeated network evaluations, with computational cost increasing with the number of sampling steps. This cost is particularly restrictive in autoregressive ensemble forecasting, motivating few-step methods.

Additionally, sample diversity is not the same as calibrated uncertainty. In conditional forecasting, generative time-steppers may become underdispersed, with samples collapsing toward a conditional mean instead of covering the range of plausible futures, especially in few-step settings \citep{dhariwal2021diffusion, ho2021classifierfree, bortoli2025distributional}. This often results in overconfident generative surrogates, which can be particularly harmful for complex and chaotic systems \citep{kassaienma}. Classical stochastic interpolants (SI) inject randomness through a prescribed inference-time diffusion coefficient, but this stochasticity is state-agnostic. Since physical predictability varies across states and regimes, calibrated uncertainty requires adaptive stochasticity.

We introduce TORNADO (laTent pOsteRioR traNsport with ADaptive diffusiOn), a framework for calibrated probabilistic forecasting in latent-space PDE surrogates. Given consecutive PDE states, TORNADO models the one-step transition between their variational autoencoder (VAE) encoder posteriors using a learned It\^{o} SDE. Our central contribution is to equip this stochastic transport with state-dependent diffusion. In contrast to fixed-schedule generative baselines, this learned diffusion acts as an adaptive uncertainty mechanism, allowing predictive spread to vary across states and dynamical regimes, and enabling calibrated, heteroscedastic forecasts. To instantiate this idea, we introduce the following key methodological contributions:
\begin{enumerate}[label=(\roman*)]
\item \textbf{Posterior-to-posterior latent forecasting.}
We formulate PDE stepping as an explicit distribution-matching problem between the VAE encoder posteriors at consecutive times. We theoretically justify this as a representation-consistent target, and empirically find that it mitigates the overconfident mean-matching behavior often observed in few-step generative time-steppers.

\item \textbf{Analytic local endpoint supervision.}
We introduce a simulation-free local supervision objective for the endpoint transport without differentiating through an SDE solver. This is achieved by defining a Gaussian interpolant over an auxiliary time $\tau$, from which we derive analytic conditional endpoint targets for local supervision.

\item \textbf{Adaptive, state-dependent diffusion.}
We learn both drift and diffusion fields for the latent SDE by matching the conditional endpoint mean and variance, augmented with an energy-distance term. This turns the diffusion field into a learned state-dependent uncertainty mechanism, enabling heteroscedastic probabilistic predictions that adapt across trajectories.
\end{enumerate}

\section{Background}
\subsection{Problem setting}
We focus on time-dependent PDEs  defined on a spatial domain $\Omega \subset \mathbb{R}^d$ over a time interval $[0, T]$. Each instance is defined by an initial condition $\textbf{u}^0(x)$, boundary conditions $\mathbf{b} \in L^2(\partial \Omega \times [0, T], \mathbb{R}^{d_b})$, a forcing term $\mathbf{f} \in L^2(\Omega \times [0, T], \mathbb{R}^{d_f})$, and/or PDE coefficients $\mathbf{c}$. The governing system is:
\begin{align}
    \mathcal{N}\!\left[ \textbf{u}; \textbf{c}, \textbf{f} \right](x,t) &= 0, && (x, t) \in \Omega \times (0, T], \\
    \mathcal{B}\!\left[ \textbf{u}; \textbf{b} \right](x,t) &= 0, && (x, t) \in \partial \Omega \times [0, T], \\
    \textbf{u}(x, 0) &= \textbf{u}^0(x), && x \in \Omega,
\end{align}
where $\mathcal{N}$ is a (possibly nonlinear) differential operator, $\mathcal{B}$ encodes the boundary conditions, and $\mathbf{u} : [0, T] \times \Omega \rightarrow \mathbb{R}^{d_u}$ denotes the solution field. We denote $\textbf{u}(x,t) = u_t$ for simplicity, where the subscript $t$ denotes time step indices separated by some $\Delta t$. We train a model $\mathcal{G}$ as a next-step forecaster: given the current state $u_t$, it defines a conditional distribution over the next state. At inference, we evaluate it as a probabilistic predictor by drawing multiple samples $\{\hat u_{t+1}^{(k)}\}_{k=1}^K \sim \mathcal{G}(\cdot\,|\,u_t)$, yielding an ensemble of candidates (and, under rollout, an ensemble of trajectories).


\subsection{Diffusion, Flows, and Stochastic Interpolants}

Transport-based generative models (e.g., normalizing flows and diffusion models) generate samples from a complex target density $\rho_1$ by continuously transporting samples drawn from a simple prior density $\rho_0$. This transport is defined by an ordinary or stochastic differential equation (ODE/SDE). SI provide a common view of these methods: training amounts to learning the vector field (or velocity) that reproduces a prescribed probability path between arbitrary $\rho_0$ and $\rho_1$ \cite{albergo2023stochasticinterpolants}.

Both flow matching (FM) and SI learn a velocity field $f_\theta$. In FM, sampling proceeds by integrating the corresponding ODE, whereas in SI one may instead use an SDE realization in which the injected noise is prescribed by an auxiliary-time diffusion schedule $g(\tau)$. For example, for the Euler-Maruyama SDE solver, a step is computed as:
\begin{equation}
      z_{\tau+\Delta \tau} =
  z_\tau
  + \Delta\tau\, f_\theta(z_\tau,\tau)
  + \sqrt{\Delta\tau}\; g(\tau)\odot \xi,
  \label{eq:em}
\end{equation}

with $\xi \sim \mathcal{N}(0,I)$. A key limitation of classical SI for PDE forecasting is the use of this \emph{global} time-dependent noise schedule $g(\tau)$. Since $g(\tau)$ is only a function of $\tau$,  the injected stochasticity is state-agnostic. In the few-step regime targeted here, this makes it difficult to adapt predictive spread to the local dynamics. We therefore learn a state-dependent diffusion amplitude so that uncertainty can adapt directly to the local regime.


\section{Method}

TORNADO parameterizes a latent Itô SDE over an auxiliary time $\tau\in[0,1]$ with learned drift and state-dependent diffusion. To train this stochastic step model, we need a tractable supervision signal for both drift and diffusion, avoiding the cost of pathwise differentiation through the SDE solver. We first describe the inference procedure to make explicit how the learned SDE generates the next latent state ($\S$~\ref{ssec:inference}). We then justify the encoder posterior of the next snapshot as the supervision target for the latent predictor ($\S$\ref{ssec:setup}). To obtain such supervision, we exploit the Gaussianity of the encoder posteriors and introduce an auxiliary-time Gaussian interpolant ($\S$~\ref{ssec:targets}). The resulting analytic conditional targets make training simulation-free and also support an energy-distance regularizer on sampled endpoint updates ($\S$~\ref{ssec:ed}).

\subsection{Inference by SDE integration}
\label{ssec:inference}

\begin{figure*}[tb!]
    \centering
    \includegraphics[width=\textwidth]{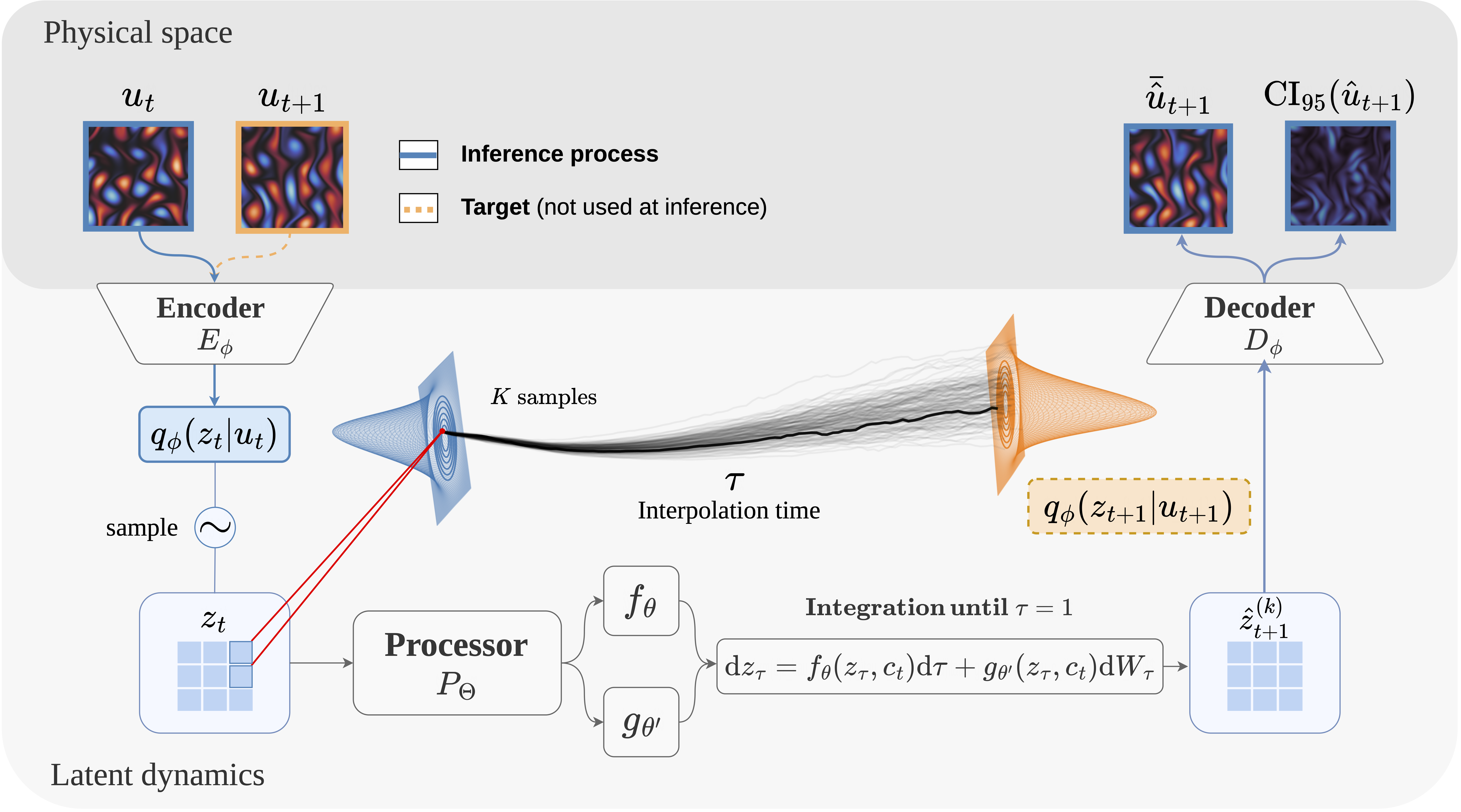}
    \caption{Inference pipeline of TORNADO. The encoder $E_\phi$ maps the current PDE state $u_t$ to a latent posterior $q_\phi(z_t\mid u_t)$, from which a latent sample $z_t$ is drawn. Starting from the encoded latent state, we draw $K$ independent realizations of the learned latent SDE over the auxiliary time $\tau\in[0,1]$, using drift $f_\theta$ and diagonal diffusion $g_{\theta'}$ predicted by $P_\Theta$. The decoder $D_\phi$ maps the resulting latent ensemble to a predictive distribution over $u_{t+1}$, yielding a mean $\bar{\hat{u}}_{t+1}$ and a 95\% confidence interval $\text{CI}_{95}(\hat{u}_{t+1})$.}
    \label{fig:tornado}
\end{figure*}

Building on the SDE formulation of SI, we generate the next latent state by integrating a learned Itô SDE over an auxiliary interpolation time $\tau\in[0,1]$:
\begin{equation}
  \mathrm{d}z_\tau
  = f_\theta(z_\tau,\tau)\,\mathrm{d}\tau
  + g_{\theta'}(z_\tau,\tau)\,\mathrm{d}W_\tau,
  \label{eq:lsfm_sde}
\end{equation}

where $W_\tau$ is a standard Wiener process. The drift $f_\theta$ and the state-dependent diffusion amplitude
$g_{\theta'}$ are parameterized by neural networks. Unlike classical SI that prescribe a global
diffusion schedule, TORNADO learns both drift and diffusion directly from data, with endpoint supervision derived from
consecutive latent posteriors.

The auxiliary time $\tau$ distinguishes this formulation from simpler one-step probabilistic predictors. We show in Appendix~\ref{ap:one-step} that deterministic $(\mu,\sigma)$ regression, MC-Dropout, and direct moment matching produce miscalibrated samples and faster error accumulation on our benchmarks.

At inference time, generating $\hat u_{t+1}$ from $u_t$ proceeds as follows. 

\begin{center}
\begin{tikzpicture}[
  font=\sffamily,
  text=ink,
  flow/.style={
    -{Stealth[length=2.4mm,width=1.8mm]},
    draw=TornadoBlue,line width=1.05pt
  },
  state/.style={
    font=\Large,inner sep=0pt,
    text height=3.8mm,text depth=1.5mm
  },
  dimension/.style={font=\small,text=muted,inner sep=1pt},
  operator/.style={font=\normalsize,text=TornadoBlue,inner sep=1pt},
  caption/.style={font=\small,text=muted,inner sep=1pt}
]

\node[state] at (0,0)   {$u_t$};
\node[state] at (2.8,0) {$z_t$};
\node[state] at (5.6,0) {$\hat z_{t+1}$};
\node[state] at (8.4,0) {$\hat u_{t+1}$};

\node[caption] at (2.8,-.80)
  {$z_t\sim q_\phi(\,\cdot\mid u_t)$};

\node[dimension] at (0,-.40)   {$\in\mathbb{R}^{d_u}$};
\node[dimension] at (2.8,-.40) {$\in\mathbb{R}^{d_z}$};
\node[dimension] at (5.6,-.40) {$\in\mathbb{R}^{d_z}$};
\node[dimension] at (8.4,-.40) {$\in\mathbb{R}^{d_u}$};

\draw[flow] (.65,0) -- (2.15,0)
  node[midway,above=2pt,operator] {$E_\phi$}
  node[midway,below=2pt,caption] {encode};

\draw[flow] (3.45,0) -- (4.95,0)
  node[midway,above=2pt,operator] {SDE}
  node[midway,below=2pt,caption] {evolve};

\draw[flow] (6.25,0) -- (7.75,0)
  node[midway,above=2pt,operator] {$D_\phi$}
  node[midway,below=2pt,caption] {decode};

\end{tikzpicture}
\end{center}
We encode $u_t$ into the posterior $q_\phi(z_t\mid u_t)$ and draw an initial sample $z_t$. Independent realizations of the learned SDE produce an ensemble $\{\hat z_{t+1}^{(k)}\}_{k=1}^K$, which is decoded to physical space as $\{\hat u_{t+1}^{(k)}\}_{k=1}^K$. For multi-step forecasting, each ensemble member is propagated autoregressively as the next initial condition, with one SDE realization per member at each step. This yields a predictive ensemble over future states.

We integrate the SDE numerically via Euler--Maruyama~(EM) for $N$ steps with step $\Delta\tau=1/N$:                       
\begin{equation}
      z^{(k)}_{n+1}
  =
  z^{(k)}_n
  + \Delta\tau\, f_\theta(z^{(k)}_n,\tau_n)
  + \sqrt{\Delta\tau}\, g_{\theta'}(z^{(k)}_n,\tau_n)\odot \xi_n,
  \label{eq:em_tornado}
\end{equation}

for $n=0,\dots,N-1$, with $\xi_n \sim\mathcal{N}(0,I)$. We refer to the number of integration steps $N$ as the number of forward evaluations (NFE). The inference procedure is summarized in Algorithm~\ref{alg:inference} (Appendix~\ref{ap:algs}) and illustrated in Figure~\ref{fig:tornado}. We study the impact of integrating with higher order solvers in Appendix~\ref{ap:solvers}.

\subsection{Objective: latent posterior transport}
\label{ssec:setup}
A common setting for neural PDE surrogates is to learn the time evolution model in a reduced latent space, where PDE states are first compressed into a lower-dimensional representation. Concretely, we use a VAE, where the encoder $E_\phi$ maps PDE solution fields $u \in \mathbb{R}^{d_u}$ to a diagonal Gaussian posterior
\[
q_\phi (z | u) = \mathcal{N}(\mu_\phi(u), \sigma^2_\phi (u)), \;\;z\in \mathbb{R}^{d_z}, \; d_z \ll d_u,
\]

and the decoder $D_\phi$ defines a conditional likelihood $p_\phi (u|z)$. 
Apart from allowing for reduced compute with the compressed representations, the VAE turns each individual snapshot $u$ into a local continuous distribution $q_\phi(z|u)$, from which we can sample different latent representations of the same field. Given a pair of consecutive time frames $(u_t, u_{t+1})$, encoding both elements yields two local posteriors $q_\phi(z_t|u_t)$ and  $q_\phi(z_{t+1}|u_{t+1})$. 

\paragraph{Encoder posterior as target.} Since $z_{t+1}$ is unobserved, the only supervision available for the next latent state comes from the encoder posterior
$q_\phi(z_{t+1}\mid u_{t+1})$. Combined with the data dynamics, this induces a target distribution over $z_{t+1}$ given
$u_t$, obtained by averaging the encoder posterior over likely next states:
\[
\tilde p(z_{t+1}\mid u_t)
:= \mathbb{E}_{u_{t+1}\sim p_{\text{data}}(\cdot\mid u_t)}\big[q_\phi(z_{t+1}\mid u_{t+1})\big].
\]

The following result shows that this encoder-induced distribution is the natural latent target for a probabilistic step model. Let  $p_\theta(z_{t+1}\mid u_t)$ denote a candidate conditional latent step model. 

\begin{equation}
\mathcal{L}(\theta)
:=\mathbb{E}_{(u_t,u_{t+1})\sim p_{\text{data}}}
\big[\mathrm{KL}\big(q_\phi(z\mid u_{t+1}) \,\|\, p_\theta(z\mid u_t)\big)\big].
\label{eqn:posterior-to-predictor-main}
\end{equation}

We use a forward-KL reference objective because, when averaged over possible next states, its minimizer is exactly the encoder-averaged distribution $\tilde p(\cdot\mid u_t)$:

\begin{proposition}
\label{prop:posterior-minimization}
    The objective in Eq.~\eqref{eqn:posterior-to-predictor-main} is minimized when
\[
p_\theta(z_{t+1}\mid u_t)=\tilde p(z_{t+1}\mid u_t),
\]
i.e., when the step model matches the encoder-induced latent target.
\end{proposition}

This identifies
$\tilde p(z_{t+1}\mid u_t)$ as the appropriate latent supervision target. The proof of Proposition~\ref{prop:posterior-minimization} is provided in Appendix \ref{ap:vae_theory}. This ties the calibration of latent probabilistic predictors to the calibration of the underlying VAE posterior, a dependence we study empirically in Appendix~\ref{ap:vae-beta}.  We next construct a tractable local supervision scheme for the drift and diffusion dynamics.

\subsection{Gaussian interpolant and conditional endpoint targets}
\label{ssec:targets}

To obtain training samples, we construct an  interpolant between independent endpoint samples $z_0\sim q_\phi(z_t| u_t)$ and $z_1\sim q_\phi(z_{t+1}| u_{t+1})$ similarly to SI:
\begin{equation}
  z_\tau = (1-\tau)z_0 + \tau z_1 + \sqrt{\gamma(\tau)}\,\xi,
  \qquad \xi\sim\mathcal{N}(0,I),
  \label{eq:interpolant}
\end{equation}

where $\gamma(\tau)=\sigma_{\gamma}\tau(1-\tau)$ and $\sigma_{\gamma}$ is a tunable hyperparameter. We study its effect empirically in Appendix~\ref{ap:sigma_scale_calibration} and analyze limiting regimes in Appendix~\ref{ap:sigma_scale_theory}. Overall, $\sigma_{\gamma}>0$ is needed to obtain well-conditioned supervision, while large values can degrade calibration.

Learning the diffusion from endpoint pairs alone is underdetermined, as many drift--diffusion pairs can induce similar endpoint distributions, while supervising diffusion through full SDE trajectories would require differentiating through the solver at substantial cost. We therefore approximate the remaining evolution to the endpoint with a local one-step transition conditioned on the bridge
state $(z_\tau,\tau)$:
\begin{equation}
  \hat{z}_1 \mid z_\tau = z_\tau + (1-\tau)\, f_\theta(z_\tau,\tau)+
  \sqrt{1-\tau}\, g_{\theta'}(z_\tau,\tau)\odot \varepsilon, \quad  \varepsilon\sim\mathcal{N}(0,I).
  \label{eq:onestep_kernel}
\end{equation}

Here, the drift determines the remaining conditional mean displacement to the endpoint via $(1-\tau)\, f_\theta(z_\tau,\tau)$, while the diffusion determines the corresponding spread via $(1-\tau)\, g_{\theta'}(z_\tau,\tau)^2$. The natural supervision targets are therefore the conditional endpoint mean and covariance of $z_1 | z_\tau$.
\begin{equation}
\begin{split}
  \mathcal{L}_{\text{drift}}
  &= \mathbb{E}_{\tau,z_\tau}
  \Big[\big\|(1-\tau)\, f_\theta(z_\tau,\tau) - (\mu_{\text{target}}-z_\tau)\big\|_2^2\Big],\\
  \mathcal{L}_{\text{var}}
  &= \mathbb{E}_{\tau,z_\tau}
  \Big[\big\|(1-\tau)\, g_{\theta'}(z_\tau,\tau)^2 - \Sigma_{\text{target}}\big\|_2^2\Big].
\end{split}
\label{eqn:losses}
\end{equation}

This construction provides local endpoint supervision rather than a constraint on the full continuous-time marginal path. Accordingly, SDE integration is used as a finite-step sampler. We set $(\mu_{\text{target}},\Sigma_{\text{target}})=(\mu_{1\mid\tau},\Sigma_{1\mid \tau})$, whose conditional endpoint moments admit the closed-form expression below:

\begin{proposition}[Conditional endpoint law under the Gaussian interpolant]
    \label{prop:transport-main}
    Under the above setup, let $\mu_{1|\tau}$ and $\Sigma_{1|\tau}$ denote the conditional mean and covariance of $z_1$ given $z_\tau$.  Then the conditional moments are given explicitly by: 
    \begin{equation*}
        \begin{split}
            \mu_{1|\tau} & = \mu_1 + \tau \Sigma_1\Sigma_{\tau}^{-1}(z_{\tau} - \mu_{\tau}),\\
            \Sigma_{1|\tau} &= \Sigma_1 - \tau^2 \Sigma_1 \Sigma_{\tau}^{-1}\Sigma_1,
        \end{split}
    \end{equation*}
    where $\mu_{\tau}=(1-\tau)\mu_0 + \tau \mu_1$ and $\Sigma_{\tau} = (1-\tau)^2 \Sigma_0+\tau^2 \Sigma_1 + \gamma (\tau)I$.
\end{proposition}
\noindent
Proposition~\ref{prop:transport-main} follows from standard Gaussian conditioning; derivation provided in
Appendix~\ref{ap:target_theory}.


\subsection{Distributional regularization via energy distance (ED)}
\label{ssec:ed}

The drift and variance losses match the conditional endpoint moments directly, but they supervise the
two components separately. We additionally use ED as a sample-level regularizer, comparing samples  $\hat{z}_1^{(k)}$  from the one-step transition \eqref{eq:onestep_kernel} and samples ${z}^{(\ell)}_{\text{target}}$ from the target conditional Gaussian $\mathcal{N}(\mu_{\text{target}},\Sigma_{\text{target}})$:
\begin{equation}
  \mathcal{L}_{\text{ED}}
  =
  \mathbb{E}_{\tau,z_\tau}\Big[\operatorname{ED}\big(\{\hat{z}_1^{(k)}\}_{k=1}^K,\{{z}^{(\ell)}_{\text{target}}\}_{\ell=1}^L\big)\Big],
  \label{eq:ed_loss}
\end{equation}
with $\operatorname{ED}(P,Q) =2\,\mathbb{E}\|X-Y\|-\mathbb{E}\|X-X'\|-\mathbb{E}\|Y-Y'\|,
 \quad X,X'\!\sim P,\;Y,Y'\!\sim Q\ \text{i.i.d}.$

Our final objective is
\begin{equation}
  \mathcal{L}
  =
  \mathcal{L}_{\text{drift}}
  + \lambda_{\text{var}}\mathcal{L}_{\text{var}}
  + \lambda_{\text{ED}}\mathcal{L}_{\text{ED}},
  \label{eq:loss}
\end{equation}
where $\lambda_{\text{var}}$ and $\lambda_{\text{ED}}$ are ramped up from $0$ after a drift warm-up stage. The training regime is summarized in Algorithm~\ref{alg:training} (Appendix~\ref{ap:algs}). A study of the impact of each loss component and their weights is carried out in Appendix~\ref{ap:loss-scheme}.

\section{Experiments}
We evaluate TORNADO’s ability to produce calibrated predictive distributions under compute-constrained, few-step inference. We compare against standard generative baselines on four PDE datasets using four metrics computed in physical space capturing mean accuracy, distributional quality, calibration, and spread. Our empirical goal is to test whether learning state-dependent diffusion improves calibration at fixed inference budget while remaining competitive in predictive accuracy.


\subsection{Experimental setting} \label{sec:exp_set}
\paragraph{Datasets}
\textbf{Synthetic data.} We use APEBench \citep{apebench} with default settings to generate 2D Kolmogorov flow, Kuramoto--Sivashinsky~(KS), Burgers, and Gray--Scott reaction--diffusion~(GS) datasets on a $160\times160$ grid, with 1024 training and 256 test trajectories each (GS: 256/64).
\textbf{Real-world data.} We use measured flame imaging from the Combustion scenario of RealPDEBench \citep{hu2026realpdebench} and ERA5 atmospheric reanalysis \citep{era5} obtained through WeatherBench \citep{weatherbench}.
The combustion dataset contains 30 recorded trajectories of 2001 $128\times128$ frames, each at a unique (NH$_3$ blend, equivalence ratio) condition; 26 trajectories are used for training and 4 for testing on unseen conditions.
For ERA5, we use 6-hourly $121\times240$ fields at $1.5^\circ$ resolution: 2-m temperature $(T_{2m})$, 10-m zonal and meridional wind $(U_{10},V_{10})$, and 500 hPa geopotential $(Z_{500})$, with calendar-year splits of 2010--2015 for training, 2016 for validation, and 2017--2018 for testing.
Further details are provided in Appendix~\ref{ap:data}.

\paragraph{Baselines}
As a generative framework, TORNADO's direct counterparts are diffusion, flow matching, and stochastic interpolants. We use those three as baselines to ensure that differences in performance can be attributed to the transport and uncertainty mechanisms rather than to model scale or task-specific architectural optimization. All baselines share the same trained VAE, conditioning, latent-processor backbone, and inference budget (measured in NFE), differing in auxiliary-time construction, training objective, and sampler.

\begin{itemize}
    \item  \textbf{Diffusion~(DDIM).}
Discrete noising with $N{=}100$ steps, trained to predict Gaussian noise and sampled with DDIM \citep{song2020ddim}.

\item \textbf{Flow Matching~(FM).}
Noise-to-data conditional FM \citep{lipman2023flow} with a linear interpolant and Euler sampling.

\item \textbf{Stochastic Interpolants~(SI).} \citep{albergo2023stochasticinterpolants}
Continuous-time adaptation of \citet{zhou2025reframinggenerativemodelsphysical}, sampled with EM.
\end{itemize}

Implementation details and comparisons with direct one-step probabilistic baselines appear in Appendices~\ref{ap:baselines} and~\ref{ap:one-step}, respectively.

\paragraph{Metrics}
The experiments characterize probabilistic surrogates along three axes: (i) accuracy of the central prediction, (ii) quality of the predictive distribution, and (iii) calibration, separating reliability from trivial over/under-dispersion. Accordingly, we report Mean Squared Error~(MSE) for mean accuracy, Continuous Ranked Probability Score~(CRPS) as a strictly proper score for distributional quality, Root Mean Squared Calibration Error~(RMSCE) to quantify calibration, and Spread-Skill ratio~(SSR) as a complementary diagnostic to detect over/under-dispersion. We also report $\lvert \log (\mathrm{SSR})\rvert$ so that lower values indicate better performance, consistent with the directionality of the other metrics, for clarity. All metrics are computed in physical space and averaged over spatial grid points, channels, batch elements, and over the reported prediction horizon (except on ERA5, where metrics use normalized units and cos(lat) weighting). Further details in Appendix~\ref{ap:metrix}.

\paragraph{Architecture and experimental setting.}
\label{ssec:arch_opt}
\textbf{VAE:}
All methods operate in the same latent space given by a Deep Compression Autoencoder (DCAE) \citep{dcae} trained with a standard $\beta$-VAE objective \citep{higgins2017betavae}. We train one VAE instance for each dataset. 
\textbf{Processor:}
TORNADO and all baselines share a Transformer trunk $P_\Theta$ (DiT-style \citep{peebles2023scalablediffusionmodelstransformers}). The trunk's output branches into a drift head $f_\theta$ and a diffusion head $g_{\theta'}$. The drift head consists of a final per-token projection followed by an unpatchify operation. The diffusion head, instantiated only for TORNADO, uses the same decoder preceded by an additional transformer block to parameterize the state-dependent diffusion amplitude; it is detached (\texttt{stopgrad}) from the shared trunk by default, so its gradients do not increase the drift model's representation capacity. More details are in Appendix~\ref{app:impl}.

\paragraph{Training protocol.}
We adopt one-step teacher-forced training as a controlled simplification shared by all methods; this isolates differences in probabilistic transport and calibration without conflating them with rollout-training effects. This choice is made for comparability rather than necessity: TORNADO is not limited to teacher forcing, and we additionally report an autoregressive extension in Appendix~\ref{ap:AR}. All methods are trained on randomly sampled consecutive pairs $(u_t,u_{t+1})$. For methods that sample an interpolation time, we draw $\tau \sim \mathcal{U}(\tau_{\min}, 1-\tau_{\min})$ with $\tau_{\min}=10^{-3}$ to avoid numerical issues near the endpoints.

\paragraph{Evaluation protocol and compute budget.}
We evaluate probabilistic forecasts under autoregressive (AR) rollout with a fixed ensemble size $K=16$ and focus on compute-constrained sampling with small NFE. Unless stated otherwise, we report results at NFE=5. We choose NFE=5 as a representative low-step regime: it reflects the compute constraints of AR ensemble forecasting, while giving all iterative baselines enough integration steps for a meaningful comparison. Additional $K$ and NFE sweeps are deferred to Appendix~\ref{ap:ensemble-size} and \ref{ap:nfe} respectively. Rollout horizons are specified per experiment to account for dataset-dependent trajectory lengths and error accumulation.

\subsection{Performance of predictive distributions}
\label{sec:rollout}
Table \ref{tab:megatable} reports interval metrics over rollout steps 1-20 , except ERA5 (1–40), across the four synthetic PDE and two real-world datasets at fixed compute budget (NFE=5). 
Across all four synthetic datasets, TORNADO achieves the lowest calibration error at fixed budget, with RMSCE reductions ranging from roughly 2× to over 5× relative to the baselines on representative cases such as Kolmogorov and Burgers. It also yields substantially better spread–skill agreement, with $|\log(\mathrm{SSR})|$ reduced by around 7× on KS and Gray–Scott relative to SI, and even more relative to DDIM and FM. These gains come while maintaining competitive MSE and CRPS in the few-step regime targeted. 

Flow matching is consistently strong in mean accuracy, but it is systematically miscalibrated. Its large $|\log(\mathrm{SSR})|$ values, together with large RMSCE, indicate under-dispersed predictive ensembles, i.e., the model concentrates too much probability mass around the ensemble mean. This aligns with prior observations that flow-matching-style training can behave primarily as a conditional mean matcher in low-step forecasting regimes, rather than faithfully representing the full predictive distribution. By contrast, the remaining baselines do not exhibit a uniform behavior across scenarios. DDIM and SI show mixed behavior across PDEs, suggesting that fixed schedules/noise parameterizations interact strongly with the dataset’s intrinsic stochasticity and error growth. 

TORNADO achieves the lowest RMSCE on both real-world datasets. On Combustion, it also leads in MSE and CRPS. SI’s slightly better SSR should be interpreted in light of its higher predictive MSE. On ERA5, TORNADO nearly matches SI's leading MSE and CRPS while improving RMSCE by approximately 10\% and achieving the best spread--skill agreement.

\newcommand{\scinum}[2]{#1\times 10^{#2}}
\definecolor{RowGray}{RGB}{247,247,247}
\definecolor{BestGreen}{RGB}{232,246,234}
\definecolor{SecondBlue}{RGB}{232,240,255}

\newcommand{\best}[1]{\cellcolor{BestGreen}\textbf{#1}}
\newcommand{\second}[1]{\cellcolor{SecondBlue}#1}
\definecolor{BestGreenText}{RGB}{70,155,105}
\definecolor{SecondBlueText}{RGB}{70,120,195}

\begin{table*}[!t]
    \centering
    \caption{Interval evaluation at NFE${=}5$ over each dataset's horizon: steps 1--20, except ERA5 (1--40). Mean $\pm$ std over the test set, except Combustion, where $\pm$ is the spread across the four held-out conditions. ERA5 aggregates its four variables in normalised, $\cos(\mathrm{lat})$-weighted units. \textbf{Best} and \underline{second-best} per dataset and metric are highlighted (\textcolor{BestGreenText}{green}/\textcolor{SecondBlueText}{blue}). }

    \setlength{\tabcolsep}{6pt}
    \renewcommand{\arraystretch}{1.08}
    \resizebox{\columnwidth}{!}{%
    \begin{tabular}{llcccc}
        \toprule
        \textbf{Dataset} & \textbf{Model}
        & MSE $(10^{-3})\,(\downarrow)$
        & CRPS $(10^{-2})\,(\downarrow)$
        & RMSCE $(10^{-2})\,(\downarrow)$
        & $|\log(\mathrm{SSR})|\,(\downarrow)$ \\
        \midrule
        \midrule
        
        \multirow{4}{*}{\textbf{Kolmogorov}} 
        & DDIM    & \merr{178}{4}     & \merr{18.84}{0.14} & \second{\underline{\merr{9.52}{0.05}}}   & \second{\underline{\merr{0.628}{0.005}}} \\
        & FM      & \best{\textbf{\merr{107}{3}}} & \second{\underline{\merr{16.33}{0.15}}} & \merr{20.33}{0.03} & \merr{1.494}{0.006} \\
        & SI      & \merr{220}{5}     & \merr{22.43}{0.14} & \merr{10.56}{0.06} & \merr{0.653}{0.005} \\
        & TORNADO & \second{\underline{\merr{120}{2}}} & \best{\textbf{\merr{14.53}{0.12}}} & \best{\textbf{\merr{3.95}{0.06}}} & \best{\textbf{\merr{0.268}{0.006}}}\\
        \midrule

        \multirow{4}{*}{\textbf{KS}} 
        & DDIM    & \merr{33.88}{0.14}  & \merr{11.12}{0.02}  & \merr{15.051}{0.015} & \merr{0.8773}{0.0014} \\
        & FM     & \second{\underline{\merr{9.138}{0.015}}} & \second{\underline{\merr{5.152}{0.005}}} & \merr{5.483}{0.010} & \merr{0.2984}{0.0007} \\
        & SI      & \merr{9.604}{0.009} & \merr{5.220}{0.003} & \second{\underline{\merr{3.378}{0.009}}} & \second{\underline{\merr{0.1417}{0.0003}}} \\
        & TORNADO & \best{\textbf{\merr{7.824}{0.012}}} & \best{\textbf{\merr{4.760}{0.004}}} & \best{\textbf{\merr{2.509}{0.007}}} & \best{\textbf{\merr{0.0209}{0.0007}}} \\
        \midrule

        \multirow{4}{*}{\textbf{Burgers}} 
        & DDIM    & \merr{0.789}{0.006} & \merr{1.523}{0.004} & \merr{11.00}{0.04} & \merr{0.575}{0.003} \\
        & FM      & \best{\textbf{\merr{0.611}{0.004}}} & \best{\textbf{\merr{1.383}{0.004}}} & \merr{10.80}{0.03} & \merr{0.647}{0.002} \\
        & SI      & \second{\underline{\merr{0.676}{0.004}}} & \merr{1.496}{0.002} & \second{\underline{\merr{7.31}{0.03}}} & \second{\underline{\merr{0.409}{0.002}}} \\
        & TORNADO & \merr{0.764}{0.007} & \second{\underline{\merr{1.462}{0.005}}} & \best{\textbf{\merr{2.55}{0.02}}} & \best{\textbf{\merr{0.003}{0.005}}} \\
        \midrule

        \multirow{4}{*}{\textbf{GS}} 
        & DDIM    & \merr{0.710}{0.006}  & \merr{1.226}{0.004}  & \merr{7.943}{0.009}  & \merr{0.5121}{0.0018} \\
        & FM      & \second{\underline{\merr{0.5176}{0.0008}}} & \second{\underline{\merr{1.0692}{0.0008}}} & \merr{8.250}{0.009} & \merr{0.5598}{0.0011} \\
        & SI      & \merr{0.5238}{0.0015} & \merr{1.0695}{0.0012} & \second{\underline{\merr{4.40}{0.02}}} & \second{\underline{\merr{0.2683}{0.0007}}} \\
        & TORNADO & \best{\textbf{\merr{0.4163}{0.0012}}} & \best{\textbf{\merr{0.9489}{0.0015}}} & \best{\textbf{\merr{2.476}{0.016}}} & \best{\textbf{\merr{0.0372}{0.0014}}} \\

         \midrule
         \midrule

        \multirow{4}{*}{\textbf{Combustion}} 
        & DDIM    & \merr{1.67}{1.51} & \second{\underline{\merr{0.812}{0.627}}} & \merr{13.18}{0.95} & \merr{2.395}{0.712} \\
        & FM      & \second{\underline{\merr{1.65}{1.40}}} & \merr{0.852}{0.590} & \merr{13.63}{1.12} & \merr{1.332}{0.241} \\
        & SI      & \merr{1.88}{1.20} & \merr{0.954}{0.430} & \second{\underline{\merr{9.52}{1.32}}} & \best{\textbf{\merr{0.526}{0.113}}} \\
        & TORNADO & \best{\textbf{\merr{1.47}{1.26}}} & \best{\textbf{\merr{0.775}{0.512}}} & \best{\textbf{\merr{8.91}{1.80}}} & \second{\underline{\merr{0.549}{0.219}}} \\

                \midrule

        \multirow{4}{*}{\textbf{ERA5}} 
        & DDIM    & \merr{288.5}{22.8} & \merr{23.37}{1.05} & \merr{15.39}{1.11} & \merr{0.650}{0.041} \\
        & FM      & \merr{292.5}{24.5} & \merr{23.19}{1.12} & \merr{12.91}{1.32} & \merr{0.576}{0.042} \\
        & SI      & \best{\textbf{\merr{265.0}{20.2}}} & \best{\textbf{\merr{21.13}{0.91}}} & \second{\underline{\merr{6.39}{0.86}}} & \second{\underline{\merr{0.266}{0.033}}} \\
        & TORNADO & \second{\underline{\merr{266.0}{20.8}}} & \second{\underline{\merr{21.16}{0.89}}} & \best{\textbf{\merr{5.77}{0.75}}} & \best{\textbf{\merr{0.254}{0.033}}} \\
        
        \bottomrule
    \end{tabular}%
    }
    \label{tab:megatable}
    \vspace{-4pt}
\end{table*}

\begin{figure}[b]
    \centering
    \includegraphics[width=0.98\linewidth]{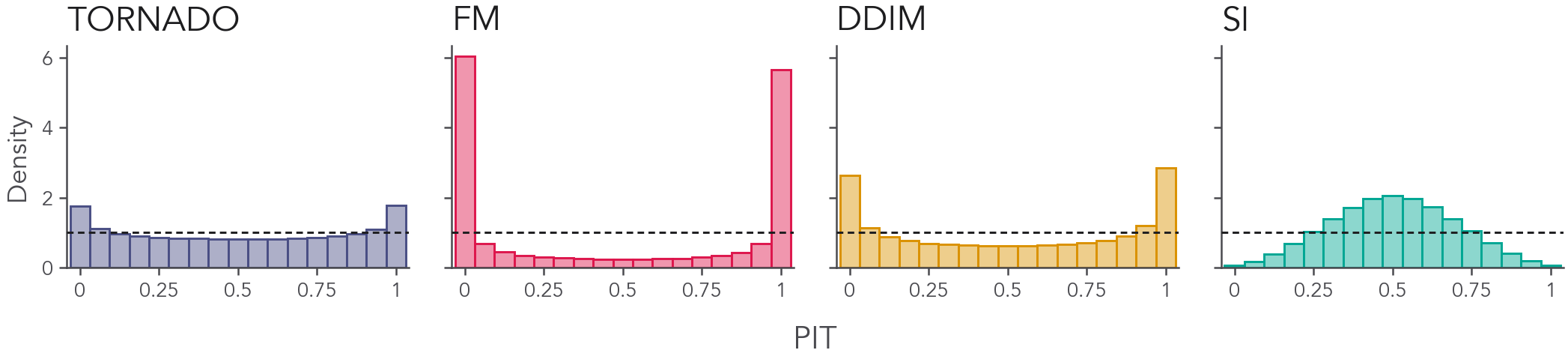}
    \caption{PIT histograms on Kolmogorov flow NFE${=}5$, $K{=}16$, 20 rollout
steps}
\label{fig:pits_main}
\end{figure} 

\subsection{Calibration diagnostics}

RMSCE quantifies calibration error and $|\log\mathrm{SSR}|$ spread--skill mismatch, but neither reveals whether ensembles are over- or underdispersed. We therefore examine probability integral transform (PIT) histograms. At each spatial location, the predictive ensemble’s empirical CDF evaluated at the ground truth gives the fraction of predictions at or below it. For calibrated forecasts, the ground truth falls below each predicted percentile with the expected frequency, yielding uniformly distributed PIT values. Bell- and U-shaped distributions indicate excessive and insufficient spread, respectively.
Figure~\ref{fig:pits_main} shows PIT histograms pooled over spatial locations, trajectories, and rollout steps 1--20 on Kolmogorov flow ($K=16$, NFE${=}5$). Despite similar RMSCE, SI is bell-shaped while DDIM is U-shaped, with FM showing an even stronger concentration at the extremes. TORNADO is closest to uniform, with only moderate excess mass at the boundaries.
More details in Appendix~\ref{ap:pits}

\subsection{Analysis over long rollouts} \label{sec:long}

\begin{figure*}[!t]
    \centering
    \includegraphics[width=0.99\linewidth]{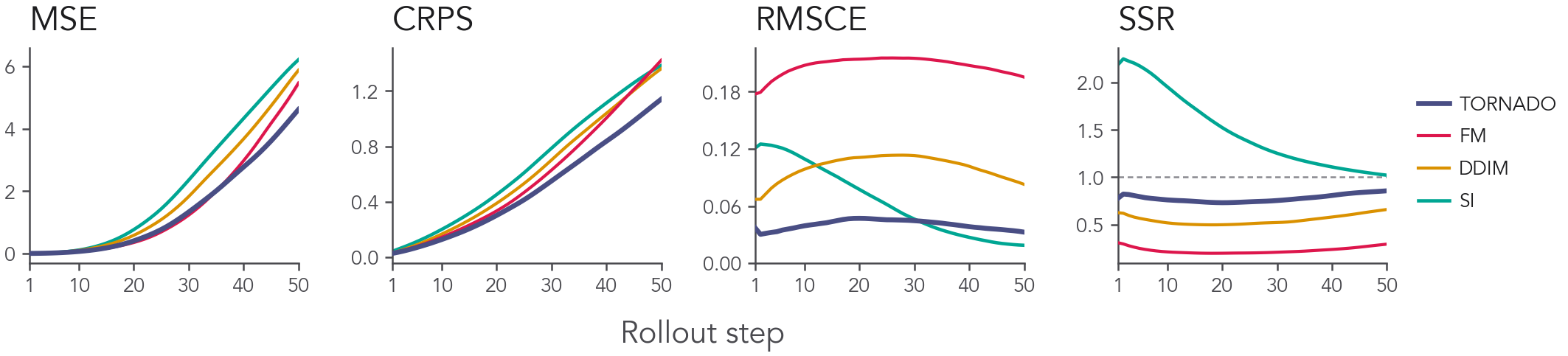}
    \caption{Kolmogorov flow rollout metrics vs.\ horizon $t$ at fixed compute (NFE=5): MSE, CRPS, RMSCE, and SSR. Lower is better for all metrics except SSR (ideal value 1).}
    \label{fig:nfe5}
\end{figure*}

\begin{figure*}[!t]
    \centering
    \includegraphics[width=\linewidth]{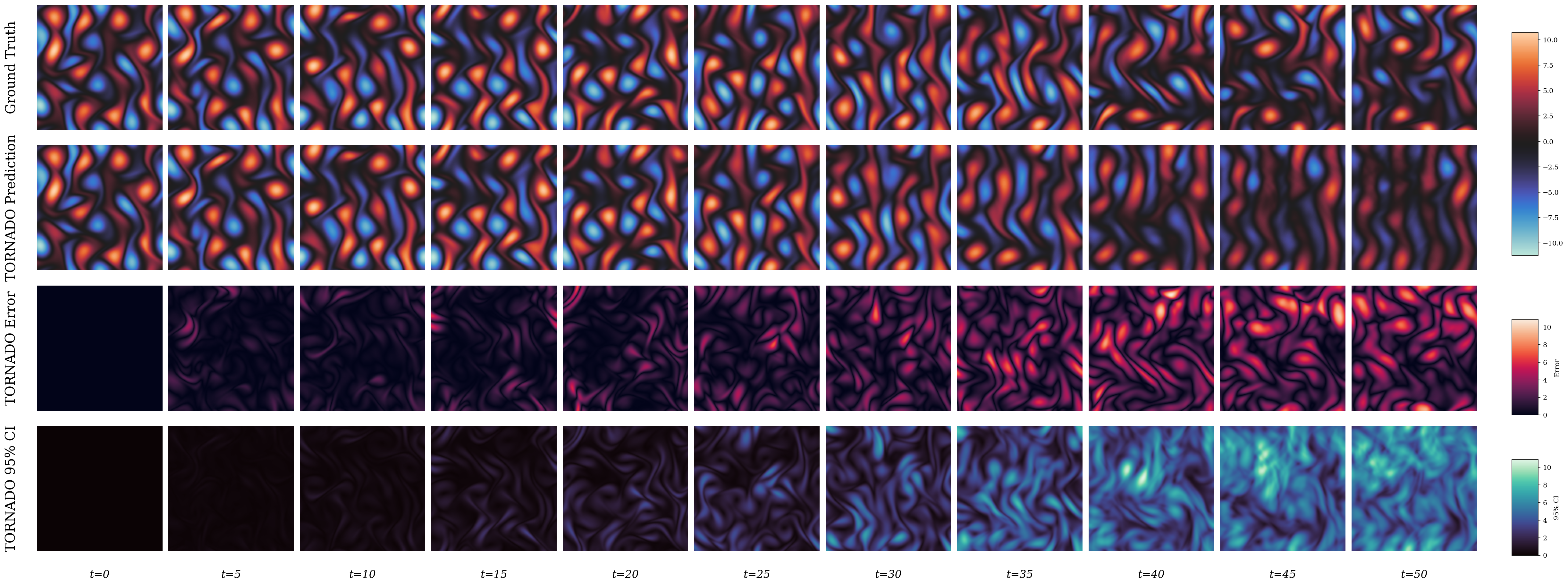}
    \caption{Rollout visualization for $T=50$ on  the Kolmogorov Flow dataset. Rows top to bottom: ground truth trajectory, TORNADO mean prediction, pixel-wise mean prediction error, pixel-wise predictive ensemble 95\% confidence interval.}
    \label{fig:viz_TORNADO}
\end{figure*}

Figure~\ref{fig:nfe5} shows per time-step rollout metrics up to $T=50$ (NFE=5). 
MSE exhibits a clear elbow around $t\approx20$, after which errors grow more rapidly. We still report MSE because it marks the onset of error amplification, but beyond that point it becomes a limited proxy for forecast quality in chaotic regimes. We report in Appendix~\ref{ap:enstrophy} a complementary long-horizon evaluation on the Kolmogorov Flow dataset based on total enstrophy, a more meaningful statistic once pointwise metrics become less informative, and find that TORNADO remains closest to the ground-truth distribution.
Calibration metrics tell a different story. Before the elbow, TORNADO retains the strongest uncertainty quality, with lower RMSCE and CRPS, and SSR closer to 1. Later, SI appears more competitive because its initial overspread is partially offset by the accumulated MSE, bringing SSR closer to 1. This should not be interpreted as SI learning a better uncertainty model at long horizons. Once the rollout leaves the stable regime, calibration metrics become increasingly influenced by compounding forecast error.
Figure~\ref{fig:viz_TORNADO} illustrates one rollout trajectory over 50 steps on the Kolmogorov dataset. As forecast error accumulates, TORNADO widens its predicted 95\% confidence interval accordingly, keeping it of comparable magnitude to the realized error over rollout.



\subsection{Ablation on adaptive diffusion} \label{ab:noise}

To isolate the effect of learning a \emph{state-dependent} diffusion term $g$, we perform an inference ablation. With the same trained drift and same NFE=5, we compare the distributional quality of the ensembles produced by performing the rollout with different integration techniques:
\begin{enumerate}[label=(\roman*)]
    \item \textbf{Euler:} deterministic ($g=0$),
    \item \textbf{EM - Constant amplitude} ($g = \bar{g_{\theta'}}$), where $\bar{g_{\theta'}}$ is the empirical average value of $g_{\theta'}(z_\tau, \tau)$ over the validation set, 

     \item \textbf{EM - Fixed schedule} ($g(\tau) =  \sqrt{\gamma(\tau)}$), where $\gamma(\tau)= \sigma_\text{scale}\tau(1-\tau)$ as in the interpolant,

     \item  \textbf{EM - TORNADO} ($g = g_{\theta'}(z_\tau, \tau)$).

\end{enumerate}

\begin{table}[H]
    \centering
    \caption{Learned state-dependent adaptive diffusion (TORNADO) vs. other integration methods for a same drift, matched NFE=5, averaged over rollout time $T=33$ for the Kolmogorov Flow dataset.}
    \begin{tabular}{lcccc}
        \toprule
        Method & MSE $\downarrow$ & CRPS $\downarrow$ & RMSCE $\downarrow$ & $|\log(\mathrm{SSR})|$ $\downarrow$ \\
        \midrule
        Euler & 0.439 & 0.351 & 0.286 & 9.674 \\
        Constant Amplitude & 2.701 & 0.810 & 0.085 & 0.602 \\
        Fixed Schedule & 0.488 & 0.291 & 0.067 & 0.361 \\
        TORNADO & \textbf{0.371} & \textbf{0.241} & \textbf{0.039} & \textbf{0.264} \\
        \bottomrule
        \end{tabular}%
    
    \label{tab:integrator}
\end{table}
Table~\ref{tab:integrator} reports metrics on the Kolmogorov Flow test set for a shared learned drift. The deterministic Euler method achieves competitive mean accuracy but the lack of stochasticity yields under-dispersed, overconfident ensembles with poor calibration. Injecting diffusion with a constant amplitude or with a fixed schedule remains suboptimal: they cannot adapt the noise level to the local state, leading to mismatched dispersion and degraded distributional scores. In contrast, learning a state-dependent adaptive diffusion term allows for more calibrated predictive distributions, with spread better aligned to forecast error and stable mean accuracy over the rollout.

\paragraph{Additional ablations and experiments:}
The Appendix provides additional ablations on diffusion loss components and weights (Ap.~\ref{ap:loss-scheme}), the impact of VAE calibration (Ap.~\ref{ap:vae-beta}), posterior quality and its effect on rollouts (Ap.~\ref{ap:vae-rollout}), higher-order solvers (Ap.~\ref{ap:solvers}), gradient stopping at the diffusion head (Ap.~\ref{ao:detach}), compute time (Ap.~\ref{ap:compute}), the impact of $\sigma_\gamma$ (Ap.~\ref{ap:sigma_scale_calibration}), and ensemble size $K$ (Ap.~\ref{ap:ensemble-size}). Additional experiments cover autoregressive training (Ap.~\ref{ap:AR}), comparisons with one-step baselines (Ap.~\ref{ap:one-step}), long-horizon statistical evaluation  (Ap.~\ref{ap:enstrophy}), calibration and sharpness beyond RMSCE (Ap.~\ref{ap:pits}), error versus variance allocation (Ap.~\ref{ap:error-variance}), ERA5 per-variable results (Ap.~\ref{ap:era5_pervar}), and the impact of NFE (Ap.~\ref{ap:nfe}).

\section{Conclusion}

TORNADO addresses calibrated probabilistic PDE forecasting in the compute-constrained, few-step regime, where fixed stochasticity can inadequately capture dynamical variability.  Its learned state-dependent diffusion adapts predictive uncertainty to the local regime, consistently improving calibration over representative generative baselines at matched inference cost while maintaining competitive accuracy. Controlled ablations show that learned state-dependent diffusion improves calibration over constant or globally scheduled alternatives, and TORNADO also outperforms standard stochastic interpolants with fixed diffusion in calibration. These results highlight adaptive stochasticity as a key ingredient for reliable probabilistic neural PDE surrogates.

\paragraph{Limitations} The quality of the learned uncertainty is  tied to the latent supervision target induced by the VAE posterior. Appendix \ref{ap:vae-beta} shows that better latent calibration improves reliability, motivating more expressive, better-calibrated representations. Additionally, our objective supervises local endpoint moments rather than explicitly enforcing consistency with the interpolant’s full continuous-time marginal evolution. While our NFE and solver ablations support its empirical effectiveness, further characterizing how local supervision and numerical discretization jointly shape the resulting endpoint distribution remains an interesting direction for future work.

\section*{Acknowledgements}
We acknowledge the financial support provided by DL4CLIM (ANR-19-CHIA-0018-01), DEEPNUM (ANR-21-CE23-0017-02), PHLUSIM (ANR-23-CE23-0025-02), MELISSA (ANR-24-CE23-7140-01), PEPR Sharp (ANR-23-PEIA0008, ANR, FRANCE 2030), and  cluster IA  PostGenAI@Paris (ANR-23-IACL-0007).

This project was provided with computer and storage resources
by GENCI at IDRIS thanks to the grants 2025-AD011015481, 2025-A0191016890, 2026-A0211016890 on the supercomputer Jean Zay’s V100/A100/H100 partitions.

\newpage

\bibliography{refs}
\bibliographystyle{plainnat}


\newpage

\appendix

\etocdepthtag.toc{mtappendix}

\etocsettagdepth{mtmain}{none}
\etocsettagdepth{mtappendix}{subsection}

\begin{center}
  {\Large \textbf{Appendix Contents}}
\end{center}
\vspace{1em}

\etocsettocstyle{}{} 
\tableofcontents

\newpage

\section{Theoretical Analysis of Latent Posterior Transport} 
\subsection{Background: Gaussian conditioning identity}
We use the standard conditioning formula for jointly Gaussian random variables. If $(X,Y)$ are jointly Gaussian with means $(\mu_X,\mu_Y)$ and covariances $\text{Var}(X)$, $\text{Var}(Y)$, $\text{Cov}(Y,X)$, then
\begin{equation*}
    Y \mid X = x \sim \mathcal{N}(\mu_Y+\text{Cov}(Y,X)\text{Var}(X)^{-1}(x-\mu_X),\text{Var}(Y)-\text{Cov}(Y,X)\text{Var}(X)^{-1}\text{Cov}(X,Y)).
\end{equation*}

This identity is all we need to derive the closed-form conditional targets in Proposition~\ref{prop:transport} (Similar formulas also appear in Gaussian Schrödinger-bridge references \cite{bunne2023schrodingerbridgegaussianmeasures}). 

\subsection{Main Results}
\label{ap:target_theory}
Let $q_{\phi}(z_t \mid u_t)=\mathcal{N}(\mu_0,\Sigma_0)$ and $q_{\phi}(z_{t+1}\mid u_{t+1})=\mathcal{N}(\mu_1,\Sigma_1)$ denote the diagonal Gaussian posteriors of consecutive PDE states, with $\Sigma_0, \Sigma_1 \succ 0$. Define the stochastic interpolant:
\begin{equation*}
    z_{\tau} = (1-\tau) z_0 + \tau z_1 + \sqrt{\gamma (\tau)} \xi, \; \tau \in (0,1),
\end{equation*}
where $z_0 \sim \mathcal{N}(\mu_0,\Sigma_0)$,  $z_1 \sim \mathcal{N}(\mu_1,\Sigma_1)$, and $\xi \sim \mathcal{N}(0,I)$ are mutually independent, and $\gamma (\tau) = \sigma_{\gamma} \tau (1-\tau)$ for some $\sigma_{\gamma} >0$. We sample $z_0 \sim q_{\phi}(z_t \mid u_t)$ and $z_1 \sim q_{\phi}(z_{t+1} \mid u_{t+1})$ independently, hence $\text{Cov}(z_0,z_1)=0$ conditional on $(u_t,u_{t+1})$. The goal of this section is to justify the use as training objectives the terms described in Eq.~\eqref{eqn:losses}.

Assumption (diagonal covariances). $\Sigma_0,\Sigma_1$ are diagonal (VAE posteriors) and $\gamma(\tau)I$ is isotropic, hence all covariances below are diagonal and operations are element-wise.

\begin{proposition}[Conditional endpoint law under the Gaussian interpolant]
    \label{prop:transport}
    Under the above setup, let $\mu_{1 \mid \tau}$ and $\Sigma_{1 \mid \tau}$ denote the conditional mean and covariance of $z_1$ given $z_\tau$.  Then the conditional moments are given explicitly by: 
    \begin{equation*}
        \begin{split}
            \mu_{1\mid \tau} & = \mu_1 + \tau \Sigma_1\Sigma_{\tau}^{-1}(z_{\tau} - \mu_{\tau}),\\
            \Sigma_{1 \mid \tau} &= \Sigma_1 - \tau^2 \Sigma_1 \Sigma_{\tau}^{-1}\Sigma_1,
        \end{split}
    \end{equation*}
    where $\mu_{\tau}=(1-\tau)\mu_0 + \tau \mu_1$ and $\Sigma_{\tau} = (1-\tau)^2 \Sigma_0+\tau^2 \Sigma_1 + \gamma (\tau)I$.
\end{proposition}

\begin{proof}
    Since $(z_0,z_1,\xi)$ are mutually independent Gaussians, the interpolant $z_{\tau}$ is Gaussian with:
    $$\mathbb{E}[ z_{\tau}]=\mathbb{E}\left[(1-\tau)z_0+\tau z_1 + \sqrt{\gamma(\tau)}\xi
    \right]= (1-\tau)\mathbb{E}[z_0]+\tau \mathbb{E}[z_1] + \sqrt{\gamma(\tau)} \mathbb{E}[\xi]=(1-\tau) \mu_0 + \tau \mu_1,$$
    $$\text{Var}( z_{\tau})=\text{Var}\left((1-\tau)z_0+\tau z_1 + \sqrt{\gamma(\tau)}\xi
    \right)= (1-\tau)^2\text{Var}(z_0)+\tau^2 \text{Var}(z_1) + \gamma(\tau) \text{Var}(\xi)= $$
    $$
    (1-\tau)^2 \Sigma_0 + \tau^2 \Sigma_1 + \gamma (\tau) I =: \Sigma_{\tau}.$$
    Since the joint distribution of $(z_{\tau},z_1)$ is Gaussian, by linearity of covariance we have that:
    $$
    \text{Cov}(z_1,z_{\tau})=\text{Cov}(z_1,(1-\tau)z_0+\tau z_1+ \sqrt{\gamma(\tau)}\xi)= \\
    $$
    $$(1-\tau)\text{Cov}(z_1,z_0)+\tau \text{Cov}(z_1,z_1)+\sqrt{\gamma(\tau)}\text{Cov}(z_1,\xi)=\tau \Sigma_1.
    $$
    By the standard Gaussian conditioning formula, for jointly Gaussian $(X,Y)$ with $Y \mid X \sim \mathcal{N}(\mu_{Y \mid X},\Sigma_{Y \mid X})$:
    $$\mu_{Y\mid X}=\mu_Y + \text{Cov}(Y,X)\text{Var}(X)^{-1}(X-\mu_X), \quad \Sigma_{Y \mid X} = \Sigma_{Y} -\text{Cov}(Y,X)\text{Var}(X)^{-1}\text{Cov}(X,Y).$$
    Applying this with $Y=z_1$ and $X=z_{\tau}$:
    \begin{equation}
    \begin{split}
    \mu_{1
    \mid \tau}&=\mu_1 + \tau \Sigma_1\Sigma_{\tau}^{-1}(z_{\tau}-\mu_{\tau}), \\
    \Sigma_{1\mid \tau}&=\Sigma_1 - \tau \Sigma_1 \Sigma_{\tau}^{-1}\tau \Sigma_1 = \Sigma_1 - \tau^{2} \Sigma_1 \Sigma_{\tau}^{-1} \Sigma_1.
    \end{split}
    \label{eqn:condmoments}
    \end{equation}
    By construction, the VAE posteriors have diagonal covariance, and $\sqrt{\gamma(\tau)}\xi$ has covariance $\gamma(\tau)I$, hence diagonal. Since sums, products, and inverses of diagonal matrices are diagonal, all covariance matrices in the above expressions are diagonal, and thus products and inverses are element-wise.
\end{proof}

Define $s_{1\mid\tau}:=\mathrm{diag}(\Sigma_{1\mid\tau})\in\mathbb{R}^d$ and let $\Delta \tau:=1-\tau$.

\begin{remark}[Moment targets for the one-step kernel]
Conditioned on $z_\tau$, the one-step update has mean $z_\tau+\Delta\tau f_\theta(z_\tau,\tau)$ and covariance $\Delta\tau\,g_{\theta'}(z_\tau,\tau)^2$.
Thus the squared-error objectives in Eq.~\eqref{eqn:losses} are minimized when these moments match $(\mu_{1\mid\tau},\Sigma_{1\mid\tau})$, i.e.,
$f^*(z_\tau,\tau)=(\mu_{1\mid\tau}-z_\tau)/\Delta \tau$ and $g^*(z_\tau,\tau)^2=s_{1\mid\tau}/\Delta \tau$.
The minimizer of $\mathcal{L}_{\text{var}}$ is unique in $g^2$, while $g$ is only identifiable up to sign (coordinate-wise). Uniqueness in $g$ can be enforced by parameterizing $g \geq 0$ (e.g., $g=\text{softplus}(\tilde{g})$). 
\end{remark}

\begin{remark}[Connection to Gaussian conditioning]
The conditional moments in Proposition~\ref{prop:transport} follow from the standard jointly-Gaussian conditioning identity, here the required moments $(\mu_\tau,\Sigma_\tau,\text{Cov}(z_1,z_\tau))$
are induced directly by the interpolant construction.
\end{remark}

\subsection{Role of the Interpolant Noise Scale} \label{ap:sigma_scale_theory}

The hyperparameter $\sigma_{\gamma}$ in $\gamma(\tau) = \sigma_{\gamma} \tau(1-\tau)$ controls the informativeness of the stochastic interpolant $z_\tau$ about the endpoint $z_1$. Under the setup of Subsection~\ref{ap:target_theory}, with $\Sigma_0, \Sigma_1 \succ 0$, recall that
\begin{equation*}
    z_{\tau} = (1-\tau) z_0 + \tau z_1 + \sqrt{\gamma (\tau)} \xi, \quad \tau \in (0,1),
\end{equation*}
where $z_0 \sim \mathcal{N}(\mu_0,\Sigma_0)$, $z_1 \sim \mathcal{N}(\mu_1,\Sigma_1)$, and $\xi \sim \mathcal{N}(0,I)$ are mutually independent.

From Proposition~\ref{prop:transport}, the marginal variance of the interpolant is
\begin{equation*}
    \Sigma_{\tau} = (1-\tau)^2 \Sigma_0 + \tau^2 \Sigma_1 + \sigma_{\gamma} \tau(1-\tau) I,
\end{equation*}
and the conditional variance of the endpoint given the interpolant is
\begin{equation*}
    \Sigma_{1\mid \tau} = \Sigma_1 - \tau^2 \Sigma_1 \Sigma_{\tau}^{-1} \Sigma_1.
\end{equation*}

We now analyze how $\sigma_{\gamma}$ affects the conditional variance $\Sigma_{1\mid \tau}$ for a fixed $\tau \in (0,1)$.

\begin{corollary}[Effect of $\sigma_{\gamma}$ and $\tau$ on conditional endpoint uncertainty]
\label{cor:sigma_tau_effects}
Recall $\Sigma_{1\mid\tau}=\Sigma_1-\tau^2\Sigma_1\Sigma_\tau^{-1}\Sigma_1$ with
$\Sigma_\tau=(1-\tau)^2\Sigma_0+\tau^2\Sigma_1+\sigma_{\gamma}\tau(1-\tau)I$.
Assume the diagonal setting, so all operations are element-wise.

\begin{enumerate}
\item \textbf{Varying $\sigma_{\gamma}$ (fixed $\tau\in(0,1)$).}
The map $\sigma_{\gamma}\mapsto \Sigma_{1\mid\tau}(\sigma_{\gamma})$ is monotonically increasing in PSD order.
Moreover,
\[
\lim_{\sigma_{\gamma}\to 0^+}\Sigma_{1\mid\tau}
= \Sigma_1-\frac{\tau^2\Sigma_1^2}{(1-\tau)^2\Sigma_0+\tau^2\Sigma_1},
\qquad
\lim_{\sigma_{\gamma}\to +\infty}\Sigma_{1\mid\tau}=\Sigma_1.
\]

\item \textbf{Varying $\tau$ (fixed $\sigma_{\gamma}>0$).}
\[
\lim_{\tau\to 0^+}\Sigma_{1\mid\tau}=\Sigma_1,
\qquad
\lim_{\tau\to 1^-}\Sigma_{1\mid\tau}=0.
\]
\end{enumerate}
\end{corollary}

\begin{proof}
Since all covariance matrices are diagonal, products and inverses are element-wise. We work coordinate-wise without loss of generality.

\textbf{1) Varying $\sigma_{\gamma}$ (fixed $\tau\in(0,1)$).}
We have
\begin{equation*}
    \frac{\partial \Sigma_{\tau}}{\partial \sigma_{\gamma}} = \tau(1-\tau) I \succ 0 \quad \text{for } \tau \in (0,1).
\end{equation*}
Thus $\Sigma_{\tau}$ is strictly increasing in $\sigma_{\gamma}$, which implies $\Sigma_{\tau}^{-1}$ is strictly decreasing. Since
\begin{equation*}
    \Sigma_{1 \mid \tau} = \Sigma_1 - \tau^2 \Sigma_1 \Sigma_{\tau}^{-1} \Sigma_1,
\end{equation*}
and the subtracted term $\tau^2 \Sigma_1 \Sigma_{\tau}^{-1} \Sigma_1$ decreases as $\Sigma_{\tau}^{-1}$ decreases, we conclude that $\Sigma_{1\mid \tau}$ is increasing in $\sigma_{\gamma}$.

As $\sigma_{\gamma} \to 0^+$, we have $\Sigma_{\tau} \to (1-\tau)^2 \Sigma_0 + \tau^2 \Sigma_1$. Direct substitution yields the result.

As $\sigma_{\gamma}\to+\infty$, for each coordinate $i$,
$\Sigma_{\tau,i}=(1-\tau)^2\Sigma_{0,i}+\tau^2\Sigma_{1,i}+\sigma_{\gamma}\tau(1-\tau)\to+\infty$,
so $\Sigma_{\tau,i}^{-1}\to 0$ and hence $\tau^2\Sigma_{1,i}^2\Sigma_{\tau,i}^{-1}\to 0$.
Therefore $\Sigma_{1\mid\tau,i}\to\Sigma_{1,i}$, i.e. $\Sigma_{1\mid\tau}\to\Sigma_1$.

\textbf{2) Varying $\tau$ (fixed $\sigma_{\gamma}>0$).}
 Rewrite the conditional variance as:
 $$
    \Sigma_{1 \mid \tau} = \Sigma_1 - \tau^2\frac{ \Sigma_1^2}{(1-\tau)^2 \Sigma_0 + \tau^2 \Sigma_1 + \sigma_{\gamma} \tau(1-\tau) I}.
$$
As $\tau \to 0^+$, the denominator remains bounded away from 0 (it converges to $\Sigma_0 \succ 0$) while the prefactor $\tau^2\rightarrow 0^+$, so the second term vanishes and $\Sigma_{1 \mid \tau} \to \Sigma_1$.

Rewrite the conditional variance as:
 $$
    \Sigma_{1 \mid \tau} = \Sigma_1 - \frac{\tau^2 \Sigma_1^2}{\Sigma_\tau} = \frac{\Sigma_1 \Sigma_\tau - \tau^2 \Sigma_1^2}{\Sigma_\tau}
    = \frac{\Sigma_1 \left( (1-\tau)^2 \Sigma_0 + \sigma_{\gamma} \tau(1-\tau) I \right)}{\Sigma_\tau}.
$$
The numerator contains the factor $(1-\tau)$, so it vanishes as $\tau \to 1^-$. The denominator approaches $\Sigma_1 \succ 0$. Therefore $\Sigma_{1 \mid \tau} \to 0$.
\end{proof}

\section{Posterior-as-target objective: a principled justification} \label{ap:vae_theory}
This section provides the formal justification for Proposition~\ref{prop:posterior-minimization} in the main text.
Intuitively, because latent states are unobserved, the only available supervision signal for $z_{t+1}$ is the encoder
posterior $q_\phi(z_{t+1}\mid u_{t+1})$. When combined with the data dynamics over $u_{t+1}$ given $u_t$, this induces an
\emph{encoder-averaged} (or \emph{aggregated}) target distribution over the next latent state conditioned on $u_t$. We show
that minimizing the expected KL objective in Eq.~\eqref{eqn:posterior-to-predictor-main} learns this induced target.

\paragraph{Setup.}
Let $p_{\text{data}}(u_t,u_{t+1})$ denote the (empirical) joint distribution of consecutive PDE states, and let
$q_\phi(z\mid u)$ be a fixed encoder density with respect to a base measure $\mu(dz)$ on $\mathbb{R}^d$.
Define the induced joint distribution over $(u_t,z_{t+1})$ by composing the data dynamics with the encoder kernel:
\begin{equation}
\label{eq:induced_joint}
\tilde p(u_t,z_{t+1})
:=\int p_{\text{data}}(u_t,u_{t+1})\,q_\phi(z_{t+1}\mid u_{t+1})\,du_{t+1}.
\end{equation}
Equivalently, for $p_{\text{data}}$-almost every $u_t$, the induced conditional density is
\begin{equation}
\label{eq:induced_conditional}
\tilde p(z_{t+1}\mid u_t)
=\int p_{\text{data}}(u_{t+1}\mid u_t)\,q_\phi(z_{t+1}\mid u_{t+1})\,du_{t+1}.
\end{equation}

\paragraph{Training objective.}
We train a conditional step model $p(z_{t+1}\mid u_t)$ by minimizing the expected KL divergence
\begin{equation}
\label{eq:posterior_to_predictor_app}
\mathcal{L}(p)
:=\mathbb{E}_{(u_t,u_{t+1})\sim p_{\text{data}}}
\Big[\mathrm{KL}\big(q_\phi(z\mid u_{t+1}) \,\|\, p(z\mid u_t)\big)\Big].
\end{equation}

For clarity, Eq.~\eqref{eq:posterior_to_predictor_app} is the same expected KL objective as Eq.~\eqref{eqn:posterior-to-predictor-main}. Expanding the KL and dropping terms independent of $p$ yields a cross-entropy term $-\mathbb{E}_{u_t} \int \tilde{p}(z\mid u_t) \log p (z \mid u_t) \mu (dz)$, where $\tilde{p}$ is defined in Eq.~\eqref{eq:induced_conditional}.

\paragraph{Technical conditions.}
Assume $\mathcal{L}(p)<\infty$ and restrict to conditional densities $p(\cdot\mid u_t)$ w.r.t.\ $\mu$ such that
$p(z\mid u_t)>0$ whenever $\tilde p(z\mid u_t)>0$ (to avoid infinite KL terms). These conditions are standard for KL-based
projection arguments.

\begin{proposition}[KL projection onto the encoder-induced target]
\label{prop:posterior-minimization-appendix}
The functional $\mathcal{L}(p)$ in Eq.~\eqref{eq:posterior_to_predictor_app} is minimized by
\[
p^*(z\mid u_t)=\tilde p(z\mid u_t)
\]
for $p_{\text{data}}$-almost every $u_t$, uniquely up to $(p_{\text{data}}\times \mu)$-null sets.
\end{proposition}

\begin{proof}
    Expand the KL divergence:
    \begin{equation*}
            \text{KL}\bigl(q_\phi(z \mid u_{t+1})\| p(z \mid u_t)\bigr) = \mathbb{E}_{z \sim q_\phi(\cdot \mid u_{t+1})}\Bigl[\log q_\phi(z \mid u_{t+1}) - \log p(z \mid u_t) \Bigr].
    \end{equation*}
    Substituting into Eq.~\eqref{eq:posterior_to_predictor_app} yields

    \[
\mathcal{L}(p)
=
\underbrace{\mathbb{E}_{(u_t,u_{t+1})\sim p_{\text{data}}}
\mathbb{E}_{z \sim q_\phi(\cdot \mid u_{t+1})}\!\left[\log q_\phi(z \mid u_{t+1})\right]}_{\text{independent of $p$}}
-
\mathbb{E}_{(u_t,u_{t+1})\sim p_{\text{data}}}
\mathbb{E}_{z \sim q_\phi(\cdot \mid u_{t+1})}\!\left[\log p(z \mid u_t)\right].
\]
Using the definition of the induced conditional $\tilde p(\cdot\mid u_t)$ in Eq.~\eqref{eq:induced_conditional} and
Fubini/Tonelli to exchange integrals (justified under $\mathcal{L}(p)<\infty$), we obtain
\[
\mathbb{E}_{(u_t,u_{t+1})\sim p_{\text{data}}}
\mathbb{E}_{z \sim q_\phi(\cdot \mid u_{t+1})}\!\left[\log p(z \mid u_t)\right]
=
\mathbb{E}_{u_t\sim p_{\text{data}}}\left[
\int \tilde p(z\mid u_t)\log p(z\mid u_t)\,\mu(dz)\right].
\]
Therefore,
\[
\mathcal{L}(p)=-\mathbb{E}_{u_t\sim p_{\text{data}}}\left[\int \tilde p(z\mid u_t)\log p(z\mid u_t)\,\mu(dz)\right]+\text{const}.
\]

Using the identity
\[
\int \tilde p(z\mid u_t)\log p(z\mid u_t)\,\mu(dz)
=
-\mathrm{KL}\big(\tilde p(\cdot\mid u_t)\,\|\,p(\cdot\mid u_t)\big)
+\underbrace{\int \tilde p(z\mid u_t)\log \tilde p(z\mid u_t)\,\mu(dz)}_{\text{independent of $p$}},
\]
we conclude that minimizing $\mathcal{L}(p)$ is equivalent to minimizing
$\mathbb{E}_{u_t\sim p_{\text{data}}}\big[\mathrm{KL}(\tilde p(\cdot\mid u_t)\,\|\,p(\cdot\mid u_t))\big]$.
This is achieved iff $p(\cdot\mid u_t)=\tilde p(\cdot\mid u_t)$ for $p_{\text{data}}$-almost every $u_t$, with uniqueness
up to $(p_{\text{data}}\times \mu)$-null sets.
\end{proof}

\section{Algorithms} \label{ap:algs}
\begin{algorithm}[h]
\caption{TORNADO training on one-step posterior pairs}
\label{alg:training}
\small
\begin{algorithmic}[1]
\Require Consecutive states $(u_t,u_{t+1})$, posteriors $q_\phi$, parameters $\Theta=(\theta,\theta')$
\State Sample $z_0\sim q_\phi(z_t\mid u_t)$ and $z_1\sim q_\phi(z_{t+1}\mid u_{t+1})$
\State Sample $\tau\sim\mathcal{U}(\tau_{\min},1-\tau_{\min})$ and $\xi\sim\mathcal{N}(0,I)$
\State Form $z_\tau=(1-\tau)z_0+\tau z_1+\sqrt{\gamma(\tau)}\,\xi$
\State Compute $(\mu_{\text{target}},\Sigma_{\text{target}})$ using Proposition~\eqref{prop:transport-main}
\State Compute $\mathcal{L}_{\text{drift}}$ and $\mathcal{L}_{\text{var}}$ using \eqref{eqn:losses}
\State Draw $\{\hat z_1^{(k)}\}_{k=1}^K$ from \eqref{eq:onestep_kernel}
\State Draw $\{\hat z_{1,\text{target}}^{(\ell)}\}_{\ell=1}^L$ from $\mathcal{N}(\mu_{\text{target}},\Sigma_{\text{target}})$
\State Compute $\mathcal{L}_{\text{ED}}$ using \eqref{eq:ed_loss} and total loss \eqref{eq:loss}
\State Update $\Theta$ by gradient descent (drift warm-up; ramp $\lambda_{\text{var}},\lambda_{\text{ED}}$)
\end{algorithmic}
\end{algorithm}

\begin{algorithm}[h]
\caption{TORNADO inference (one-step forecast and rollout)}
\label{alg:inference}
\small
\begin{algorithmic}[1]
\Require Current state $u_t$, $(E_\phi,D_\phi)$, learned $(f_\theta,g_{\theta'})$, steps $N$, samples $K$
\State Sample $z_0 \sim q_\phi(z_t\mid u_t)$. Generate $K$ copies $\{z_0^{(k)}\}_{k=1}^K$
\State $\Delta\tau \gets 1/N$
\For{$k=1$ \textbf{to} $K$}
  \State $z \gets z_0^{(k)}$
  \For{$n=0$ \textbf{to} $N-1$}
    \State $\tau_n \gets n\,\Delta\tau$, $\varepsilon_n\sim\mathcal{N}(0,I)$
    \State $z \gets z + \Delta\tau\, f_\theta(z,\tau_n) + \sqrt{\Delta\tau}\, g_{\theta'}(z,\tau_n)\odot \varepsilon_n$
  \EndFor
  \State $\hat z_{t+1}^{(k)} \gets z$
\EndFor
\State Decode $\{\hat u_{t+1}^{(k)}\}_{k=1}^K \gets D_\phi(\{\hat z_{t+1}^{(k)}\}_{k=1}^K)$
\State \textit{(Optional)} Autoregressive rollout: set $z_0^{(k)}\leftarrow \hat z_{t+1}^{(k)}$ and repeat
\end{algorithmic}
\end{algorithm}

\clearpage

\section{Related Work}
\label{ap:related}
\paragraph{Generative models for PDE surrogates.}

Generative models learn (possibly conditional) distributions over high-dimensional data, enabling sampling and uncertainty-aware prediction. While they have driven major advances in vision and language, there is increasing interest in using them as neural surrogates for spatiotemporal physical systems governed by PDEs.
\textbf{Conditional diffusion models} \cite{ho2020ddpm,song2020ddim} learn to sample from a target distribution by iteratively denoising a sequence of corrupted variables, typically predefined by a fixed forward noising process. Conditioning can be imposed on context such as past states, parameters, or forcings, producing conditional samplers for PDE solution fields.
In PDE surrogate modeling for forward processes, diffusion models are commonly used in two ways. 
(i) \emph{Refinement / correction}: a diffusion model acts as a learned stochastic corrector that corrects trajectories produced by a deterministic surrogate by denoising, improving long-horizon stability under autoregressive rollout \cite{lippe2023pderefiner,pedersen2025thermalizerstableautoregressiveneural,serrano2024aroma}. 
(ii) \emph{Direct generative surrogate}: the diffusion model itself predicts future solution states by sampling $u_{t+\Delta t}$ conditioned on history, parameters, or observations \cite{huang2024diffusionpde,jacobsen2024cocogenphysicallyconsistentconditionedscorebased,shysheya2024on,Kohl2023ACDM,Price2024weather}. Concurrently, \textbf{Conditional flow matching} \citep{lipman2023flow} learns a time-dependent vector field $v_\theta(x,t\,|\,c)$ whose induced ODE transports a simple base distribution (usually a standard Gaussian) to the data distribution, enabling fast sampling via probability flow ODE integration. This provides a diffusion-adjacent alternative with deterministic dynamics and has recently been explored for PDE surrogate generation \cite{baldan2025pbfm,Li2025LatentPDEFlowMatching, li2025generativelatentneuralpdesolver, koupai2025efficientgenerativetransformeroperators}, while providing a simulation free alternative to Continuous Normalizing flows \cite{chen2019neuralordinarydifferentialequations} without the need to integrate the adjoint for backpropagation. \citet{rout2026fastflexibleprobabilisticforecasting} combine learned initial-state perturbations with deterministic flow-matching propagation. FuXi-ENS~\citep{Zhong2025} uses a VAE-based model to generate flow-dependent Gaussian perturbations at each autoregressive forecast step, followed by a deterministic forecasting network. Both approaches learn ensemble variability through perturbation generation, whereas TORNADO learns the diffusion amplitude within the latent stochastic transport.

Finally, \textbf{stochastic interpolants (SIs)} \cite{albergo2023stochasticinterpolants} unify diffusion- and flow-based perspectives by learning transports along stochastic paths connecting two endpoint distributions. In practice, SIs learn a drift field (often via conditional regression of an interpolant) and generate samples by integrating an It\^{o} SDE whose diffusion schedule is specified by the user. This formulation is particularly natural for PDE surrogates that map one state distribution to the next (``data-to-data''), and has been adopted in recent work on generative physical forecasting \cite{zhou2025reframinggenerativemodelsphysical,hou2025cfolearningcontinuoustimepde}. \cite{zhou2025reframinggenerativemodelsphysical} benchmarks SIs against diffusion and flow baselines on multiple physics and climate datasets, while \cite{hou2025cfolearningcontinuoustimepde} proposes spline-based probability paths for continuous-time forecasting and a simulation-free alternative to Neural ODE training. Extensions imposing physical structure or constraints on SI-based surrogates have also been explored for turbulent flows 

\vspace{-5pt}

\cite{mücke2025physicsawaregenerativemodelsturbulent}.
\paragraph{Uncertainty quantification in neural PDE solvers} 

Quantifying predictive uncertainty in neural PDE surrogates remains challenging, 
and existing approaches can be broadly categorized by their methodological strategy.

\textit{Bayesian and ensemble methods.} 
Classical strategies adapt Bayesian inference to neural operators.  The Bayesian Neural Operator~(BNO)~\cite{magnani2022approximate} employs a Laplace approximation of the weight posterior, while deep ensembles~\cite{lakshminarayanan2017simplescalablepredictiveuncertainty} remain a strong baseline albeit computationally expensive.  DiverseNO~\cite{mouli2024using} reduces this cost for Fourier Neural Operators~(FNO)~\cite{li2020fourier} by generating multiple outputs from a single forward pass through the final Fourier layer. DINOZAUR~\cite{matveev2025light} takes a different approach, replacing the 
dense FNO multiplier with a heat-kernel-inspired diffusion parameterization that 
requires only one learnable time parameter per channel; by placing priors on these 
parameters and performing variational inference, it yields spatially correlated 
uncertainty estimates with drastically reduced parameter count.
BARNN~\cite{coscia2025barnn} extends Bayesian treatment to autoregressive architectures 
by treating both network weights and hidden states as stochastic.

\textit{Latent-space and bottleneck methods.} 
An alternative strategy incorporates uncertainty within a compressed representation. IB-UQ~\cite{guo2024ib} uses an information bottleneck to recover predictive means and variances, while LE-PDE-UQ~\cite{wu2024uncertainty} introduces an encode-process-decode architecture where uncertainty is modeled in the latent processor and decoded via a separate head. These methods reduce the dimensionality of the uncertainty estimation problem 
but typically rely on parametric assumptions (e.g., Gaussian outputs).

\textit{Post-hoc calibration.} 
Rather than modifying the model architecture, \cite{gopakumarcalibrated} applies conformal prediction to physics residual errors, providing distribution-free coverage guarantees. However, this requires a held-out calibration set and does not improve the underlying predictive distribution.

\textit{Proper scoring approaches.} 
The Probabilistic Neural Operator~(PNO)~\cite{bülte2025probabilisticneuraloperatorsfunctional} provides theoretical guarantees for training neural operators using the Energy Score (ES), directly optimizing a strictly proper scoring rule. This aligns with our evaluation methodology, though PNO operates in physical space and does not exploit temporal structure.

Most existing methods operate in physical space, incurring costs that scale with the spatial resolution of the PDE discretization. Moreover, few approaches explicitly leverage the sequential structure of autoregressive forecasting, where consecutive states are often similar.

\section{Dataset details} \label{ap:data}

\subsection{Synthetic data}

The synthetic PDE trajectory data were generated with APEBench \cite{apebench}, a benchmark suite to evaluate autoregressive neural PDE emulators. All datasets were generated on 2D spatial grids of size $[160,160]$, with 1024 trajectories for train and 256 for test (except Gray-Scott, where it is 256/64) using the suite defaults. Concretely:
\begin{itemize}
    \item \textbf{Kolmogorov Flow:} 2D incompressible Navier-Stokes on a periodic square domain of extent $2\pi$, with viscosity $\nu=10^{-2}$ (Re$=100$ in the scenario convention) and linear drag $-0.1$, driven by a sinusoidal forcing of amplitude $1.0$ at injection mode $k=4$; integrated with $\Delta t=0.1$ using $20$ solver substeps ($\Delta t_{\text{solver}}=0.005$) and $500$ warmup steps. We generate 100 steps train and test.
    \item \textbf{Kuramoto-Sivashinsky (KS):}  parameterized through the suite difficulty coefficients with integrated diffusion and hyper-diffusion $(\gamma_2,\gamma_4)=(-1.2,-15.0)$ and a gradient term $\delta=-6.0$; $500$ warmup steps are used.  We generate 50 steps train and test.
    \item \textbf{Burgers:} 2D Burgers single-channel scenario on a unit domain (extent $1.0$) with outer step $\Delta t=0.1$ (one substep), convection coefficient $-0.125$ and viscosity $3\times 10^{-4}$ (no external forcing).  We generate 25 steps train and test.
    \item \textbf{Gray-Scott (GS):} two-channel reaction--diffusion system on a unit domain (extent $1.0$) with feed/kill rates $(F,k)=(0.04,0.06)$ and diffusivities $(D_1,D_2)=(2\times10^{-5},1\times10^{-5})$, integrated with $\Delta t=10.0$ using $10$ solver substeps ($\Delta t_{\text{solver}}=1.0$).  We generate 100 steps train and test.
\end{itemize}

\subsection{Real-world data}
In order to test whether the calibration gains transfer to observational or experimentally measured systems, where measurement noise, partial observation, and model mismatch differ from clean simulator data, we carry out the main experiment also on two real world datasets.  We use measured flame imaging from the Combustion scenario of RealPDEBench \citep{hu2026realpdebench} and ERA5 atmospheric reanalysis \citep{era5} obtained through WeatherBench \citep{weatherbench}.

\subsubsection{RealPDEBench}

RealPDEBench \citep{hu2026realpdebench} is a benchmark for scientific ML that integrates real-world measurements with paired numerical simulations. We focus on one of their proposed tasks, real-world training, which consists of training and evaluating on real, observational, noisy data.

We use the Combustion dataset, consisting of cross-sectional data of 3D swirl-stabilized NH$_3$/CH$_4$/air flames, measured as real-world OH$^*$ chemiluminescence intensity $I$. Table~\ref{tab:combustion} compiles its specifications. It contains 30 trajectories of 2001 $128\times128$ frames, each at a distinct combination of NH$_3$ blend and equivalence ratio. The train--test split is 26/4 by trajectory, so no test condition is seen during training: the model must generalize to an unseen (blend, $\phi$) pair, whose individual values it has only encountered in other combinations. All models are conditioned on the pair (NH$_3$ blend, $\phi$) through adaLN.

\begin{table}[t]
\centering
\caption{RealPDEBench Combustion, experimental subset: dataset specifications.}
\label{tab:combustion}
\begin{tabular}{ll}
\toprule
Trajectories            & 30 (one per operating condition) \\
Frames per trajectory   & 2001 \\
Spatial resolution      & $128\times128$ \\
Channels                & 1 (OH$^*$ chemiluminescence intensity) \\
Conditioning            & NH$_3$ blend $\in\{0,20,40,60,80\}\%$; \\
                        & $\phi \in \{0.75, 0.85, 0.9, 0.95, 1.0, 1.05, 1.1, 1.15, 1.2, 1.25, 1.3\}$ \\
Split (by trajectory)   & 26 train / 4 test \\
Held-out conditions (NH$_3$, $\phi$) & $(0\%, 1.10)$, $(20\%, 1.05)$, $(40\%, 1.10)$, $(80\%, 1.15)$ \\
Normalisation           & $\log(1+x)$ followed by a scalar $z$-score, fitted on training trajectories \\
\bottomrule
\end{tabular}
\end{table}

The convention proposed by RealPDEBench is to evaluate forecasts on short spatiotemporal windows sampled from the long trajectories. We use non-overlapping 21-frame windows: the first frame is the initial condition and the model predicts the remaining 20 autoregressively. Each held-out trajectory gives 95 windows, 380 in total.

\subsubsection{WeatherBench ERA5}

ERA5 \citep{era5} is the ECMWF atmospheric reanalysis: a gridded state estimate produced by assimilating observations into a numerical weather prediction model. We use the $1.5^\circ$ WeatherBench-2 distribution \citep{weatherbench}. Table~\ref{tab:era5} compiles its specifications. Where Combustion is a small, noisy, single-channel measurement, ERA5 is a large four-channel state on a periodic sphere, so together they probe the same calibration question in two very different regimes.

Splits are by whole calendar year: 2010--2015 for training, 2016 for validation and 2017--2018 for testing. We take a subset of GraphCast's fields \citep{graphcast}: three surface variables and one atmospheric variable. This is not a full atmospheric state and we claim no complete forecasting system, but the four are heterogeneous by design: a thermodynamic scalar, a vector pair whose components must stay correlated, and the circulation field that is the standard verification variable for medium-range skill. Because they differ by two orders of magnitude, each channel is $z$-scored with training-year statistics. The same applies at evaluation, aggregate metrics are computed in normalized units and weighted by $\cos(\text{latitude})$. Conditioning is on four date features, $[\sin,\cos]$ of day-of-year and of time-of-day, injected through adaLN.

\begin{table}[h]
\centering
\caption{WeatherBench-2 ERA5 at $1.5^\circ$: dataset specifications.}
\label{tab:era5}
\begin{tabular}{ll}
\toprule
Grid                  & $121\times240$ (latitude $\times$ longitude), poles included, longitude periodic \\
Cadence               & 6-hourly \\
Channels              & 4 ($T_{2m}$, $U_{10}$, $V_{10}$, $Z_{500}$) \\
Frames                & 8764 train / 1464 validation / 2920 test \\
Split (by year)       & 2010--2015 / 2016 / 2017--2018 \\
Encoder padding       & $121 \rightarrow 128$ latitudes: circular in longitude, replicate at \\
                      & the poles; cropped back after decoding \\
Latent grid           & $16\times30\times8$ \\
Conditioning          & $[\sin, \cos]$ of day-of-year and time-of-day (4 features) \\
Normalisation         & per-channel $z$-score, fitted on the training years \\
\bottomrule
\end{tabular}
\end{table}

Evaluation uses non-overlapping 41-frame windows: the first frame is the initial condition and the model predicts the remaining 40 autoregressively, a 10-day forecast at 6-hourly resolution.
Each test year gives 35 windows, 70 in total.
\newpage
\section{Baselines} \label{ap:baselines}
\textbf{Conditional Diffusion Models}
We use a standard conditional diffusion model that defines a discrete noising process indexed by $\tau \in \{0, \dots, N-1\}$, with $N=100$ noising steps. The target $z_1=E_\phi(u_{t+1})$ is noised using a linear schedule as
\[
    z_\tau = \sqrt{\bar{\alpha}_\tau}z_1 + \sqrt{1-\bar{\alpha}_\tau} \varepsilon\,, \;\; \varepsilon\sim \mathcal{N}(0,I),
\]   
with $\alpha_\tau = 1 - \beta_\tau$ and $\bar{\alpha}_\tau = \prod_{j=0}^\tau \alpha_j$. The network is trained to predict $\varepsilon$ by minimizing:
\[
    \mathbb{E}\big[\lVert \varepsilon - \varepsilon_\theta([z_0,z_\tau], \tau)\rVert^2_2\big].
\]

At inference, we sample following the protocol from \textbf{Denoising Diffusion Implicit Models (DDIM)} \cite{song2020ddim} non-Markovian deterministic sampling where we can control the number of denoising steps to compare NFEs.

\textbf{Conditional Flow Matching (FM)} \cite{lipman2023flow}
We implement a standard noise-to-data Conditional Flow Matching baseline defining a linear interpolator in continuous interpolation time $\tau \in [\tau_{min},1-\tau_{min}]$:
\[
    z_\tau = (1-\tau) z_1 + \tau \varepsilon \,, \;\; \varepsilon\sim \mathcal{N}(0,I).
\]
We sample $\tau \sim \operatorname{Unif}(0,1)$ and supervise the velocity target $v^*=\dot{z}_\tau=\varepsilon - z_1$ and minimize:
\[
    \mathbb{E}\big[\lVert v^* - f_\theta([z_0,z_\tau], \tau)\rVert^2_2\big].
\]
At inference, we start from a sample $z_{\tau=1}\sim\mathcal{N}(0,I)$ and integrate the auxiliary-time ODE from $\tau=1 \rightarrow0$ (prediction for next-step posterior sample $z_1$) with the Euler method, where we can choose the number of integrator steps (NFE).

\textbf{Stochastic Interpolant (SI)} \cite{albergo2023stochasticinterpolants}
We follow the implementation from \cite{zhou2025reframinggenerativemodelsphysical} but changing
the definition of the interpolation time to be continuous in $\tau \in [\tau_{min},1-\tau_{min}]$,
like in the FM baseline. We define the stochastic interpolant between the two latent posterior
distributions of consecutive timesteps as
\[
    I(\tau) = (1 - \tau)z_0 + \tau z_1 + \gamma(\tau) W_\tau,
\]
with $z_0\sim q_\phi(z_t | u_t)$, $z_1\sim q_\phi(z_{t+1} | u_{t+1})$, $\gamma(\tau) = 1-\tau$, and
$W_\tau$ a standard Wiener process, so that $W_\tau\sim \mathcal{N}(0,\tau I)$. The stochastic
perturbation therefore has scale $(1-\tau)\sqrt{\tau}$ and vanishes at both endpoints, respecting
the SI boundary conditions. By It\^o's lemma, $I(\tau)$ satisfies
\[
\mathrm{d}I(\tau) = \big[z_1 - z_0 + \dot\gamma(\tau)W_\tau\big]\mathrm{d}\tau
                  + \gamma(\tau)\,\mathrm{d}W_\tau ,
\]
and since $W_\tau$ is not differentiable, no chain-rule term in $\sqrt{\tau}$ arises. We train the
model to match the drift $b(\tau, z_0, z_1) = z_1 - z_0 + \dot\gamma(\tau)W_\tau$ by minimizing
Eq.~\ref{eq:si_loss}.
\begin{equation}
\mathcal{L}_{\theta}=\mathbb{E}_{\tau,z_0,z_1,W_\tau}\!\left[\left\|v_\theta(I(\tau),\tau)-b(\tau, z_0, z_1)\right\|_2^2\right].
\label{eq:si_loss}
\end{equation}
At inference, similarly to FM, we integrate the drift with tuneable NFE, using Euler--Maruyama to
solve $\mathrm{d}z = v_\theta\,\mathrm{d}\tau + g(\tau)\,\mathrm{d}W_\tau$ with diffusion schedule
$g(\tau)=\gamma(\tau)$. For a general $g$ the drift requires the correction
$\tfrac{1}{2}(g^2-\gamma^2)\nabla\!\log\rho_\tau$; with $g=\gamma$ this term vanishes.

\section{Metrics} \label{ap:metrix}
\paragraph{Mean Square Error (MSE):} we compute the mean square error of the ensemble mean with respect to the ground truth as a measure of accuracy:

\[
\mathrm{MSE} := \mathbb{E}\big[\|\bar u_{t+1}-u_{t+1}\|_2^2\big].
\]

\paragraph{Continuous Ranked Probability Score (CRPS):} it is a widely used strictly proper scoring rule \cite{Gneiting2007} which evaluates the full predictive distribution beyond first and second moments. We report the empirical CRPS computed per grid entry from the ensemble,
\[
\mathrm{CRPS}(\hat F, y) = \mathbb{E}_{X\sim \hat F}|X-y|
-\tfrac{1}{2}\mathbb{E}_{X,X'\sim \hat F}|X-X'|,
\]
and averaged over all entries. Here $\hat F$ denotes the empirical distribution supported on $\{\hat u^{(k)}_{t+1}\}_{k=1}^{K}$ and $y=u_{t+1}$.

\paragraph{Root Mean Square Calibration Error (RMSCE):} calibration refers to how much the predicted uncertainty of a model corresponds to the actual uncertainty of the prediction \cite{guo2017calibrationmodernneuralnetworks}. We quantify calibration computing probability integral transform (PIT) values computed per grid entry. For a calibrated prediction, the PIT should be distributed uniformly on $[0,1]$. 

Let $p_m$ be $M=50$ evenly-spaced points in $[0,1]$ and let $\widehat G(p)$ be the empirical CDF of PIT values.
We report the root mean-squared calibration error:
\[
\mathrm{RMSCE}_{\mathrm{PIT}} := \Big(\tfrac{1}{M}\sum_{m=1}^{M}\big(\widehat G(p_m)-p_m\big)^2\Big)^{1/2}.
\]

\paragraph{Spread-Skill Ratio (SSR):} Because RMSCE measures coverage at a finite set of nominal levels and does not directly penalize overly wide intervals, we additionally report SSR as a sharpness/spread diagnostic:
\[
\mathrm{SSR} := \frac{\sqrt{\mathbb{E} \big[ \mathrm{Var(\hat u^{(k)}_{t+1})\big]} }}{\sqrt{\mathrm{MSE}(\bar u_{t+1},u_{t+1})}}.
\]
Ideally, $\mathrm{SSR} \approx 1$,  values $> 1$ indicate overspread and values $< 1$, underspread. We report $\lvert \log (\mathrm{SSR})\rvert$ so that lower is better, as in the other metrics, for clarity.

\section{Implementation details} \label{app:impl}


\noindent\textbf{Model sizes.}
\begin{center}
\setlength{\tabcolsep}{6pt}
\renewcommand{\arraystretch}{1.05}
\begin{tabular}{l r}
\toprule
  \textbf{Type} & \textbf{Params} \\
\midrule
DiT (Diffusion Transformer) & 15.9M \\
Convolutional VAE (encoder--decoder) & 20.5M \\
\bottomrule
\end{tabular}
\end{center}

\noindent\textbf{VAE}

The autoencoder uses a  Deep Compression Autoencoder (DCAE) style backbone \cite{dcae} trained with a standard $\beta$-VAE objective \cite{higgins2017betavae}. Defined with latent dimensionality 4 channels and an $8\times$ spatial downsampling (from $160\times160$ to $20\times20$ for APEBench, $16\times16$ for Combustion and ERA5 uses 8 latent channels, giving $16\times30\times8$ ). %

\noindent\textbf{DiT}

The latent stepper is a Diffusion Transformer \citep{peebles2023scalablediffusionmodelstransformers} operating on the $20\times20$ latent grid (one token per spatial site, i.e., $1\times1$ patch projection). For Real Combustion and ERA5 it operates on the $16\times16$ and $16\times30$ grids above, with all other DiT hyperparameters unchanged. %
We use 12 transformer blocks with hidden width 256 and 16 self-attention heads; conditioning is injected through a 256-dimensional embedding (used to condition on $\tau$). Inputs are the current $z_\tau$ and start-point sample $z_0$ concatenated as $[z_0,z_\tau]$. %

When training TORNADO, we implement a separate diffusion-prediction branch and stop gradients from this branch into the shared transformer trunk; additionally, the diffusion branch includes an extra transformer block before its output projection. %

\noindent\textbf{Optimization and schedules. (DiT)}

We train the dynamics model with Adam at learning rate $10^{-4}$ and a step-wise exponential decay schedule (StepLR with step size 10 and multiplicative factor $0.99$).
\paragraph{DiT training}
We use $K=L=16$ for the ED samples, $\lambda_\text{ED}$ and $\lambda_\text{var}$ are ramped from 0 to $(0.1,1)$ respectively during 200 epochs after a 1000 epoch warmup. TORNADO and baselines are trained matched on number of epochs (5000). 

\clearpage

\section{Additional ablations} \label{ab:addi}
\subsection{Ablation on diffusion losses}
\label{ap:loss-scheme}
Here, we test the importance of each of the diffusion loss terms. First, we train several instances of TORNADO with combinations of  $\lambda_\text{var} \in \{0.0,1.0\}$ and $\lambda_\text{ED} \in \{0.0,0.1,1.0\}$ for the diffusion loss terms in Equation \ref{eq:loss}. We report on Table \ref{tab:loss_ablation} the averaged inference results at rollout horizon $T=10$ for each of the trained instances at matched NFE=5 for the Kolmogorov Flow data set. 

\begin{table}[H]
\centering
\caption{Diffusion-loss ablation on Kolmogorov Flow at NFE=5. Lower is better.}
\begin{tabular}{lcccc}
\toprule
Loss setting & MSE $\downarrow$ & CRPS $\downarrow$ & RMSCE $\downarrow$ & $|\log(\mathrm{SSR})|$ $\downarrow$ \\
\midrule
$\lambda_{ED}=0, \lambda_{var}=1$   & 0.0245 & 0.0766 & 0.1004 & 0.6240 \\
$\lambda_{ED}=0.1, \lambda_{var}=0$   & 0.0244 & 0.0744 & 0.0420 & 0.1520 \\
$\lambda_{ED}=1, \lambda_{var}=0$     & 0.0388 & 0.0929 & 0.0583 & \textbf{0.1273} \\
$\lambda_{ED}=0.1, \lambda_{var}=1$ & \textbf{0.0196} & \textbf{0.0662} & \textbf{0.0309} & 0.2335 \\
\bottomrule
\end{tabular}%
\label{tab:loss_ablation}
\end{table}

Moment matching alone ($\lambda_\text{ED}=0.0, \lambda_\text{var}=1.0$) produces predictions that are not well calibrated due to under-dispersion, whereas supervising on strong ED only ($\lambda_\text{ED}=1.0, \lambda_\text{var}=0.0$) degrades the mean accuracy of the ensemble. We find that combining both terms achieves the best overall performance under this trade-off, supporting the use of ED as a stabilizing regularizer for learning diffusion while retaining accurate mean dynamics.

\begin{figure}[h!]
    \centering
    \begin{subfigure}[b]{0.48\textwidth}
        \centering
        \includegraphics[width=\linewidth]{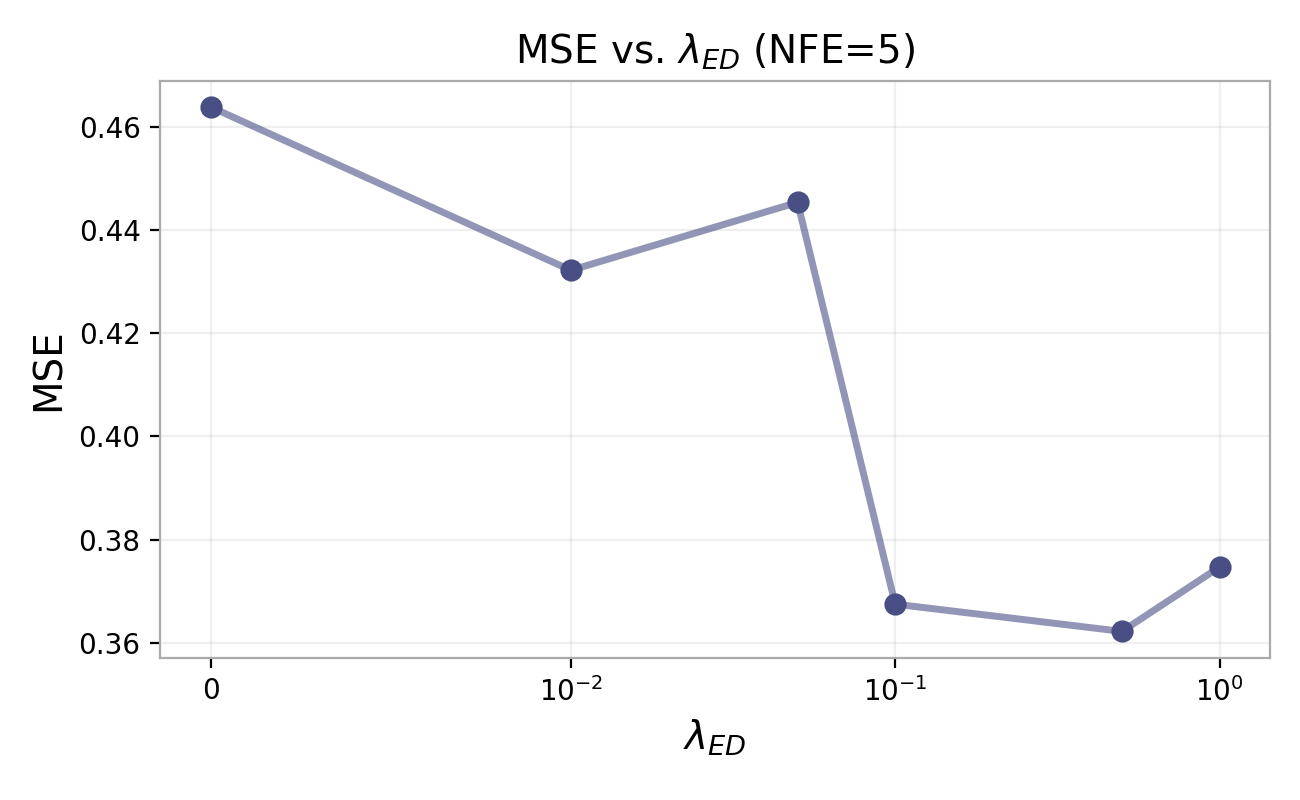}

    \end{subfigure}
    \hfill 
    \begin{subfigure}[b]{0.48\textwidth}
        \centering
        \includegraphics[width=\linewidth]{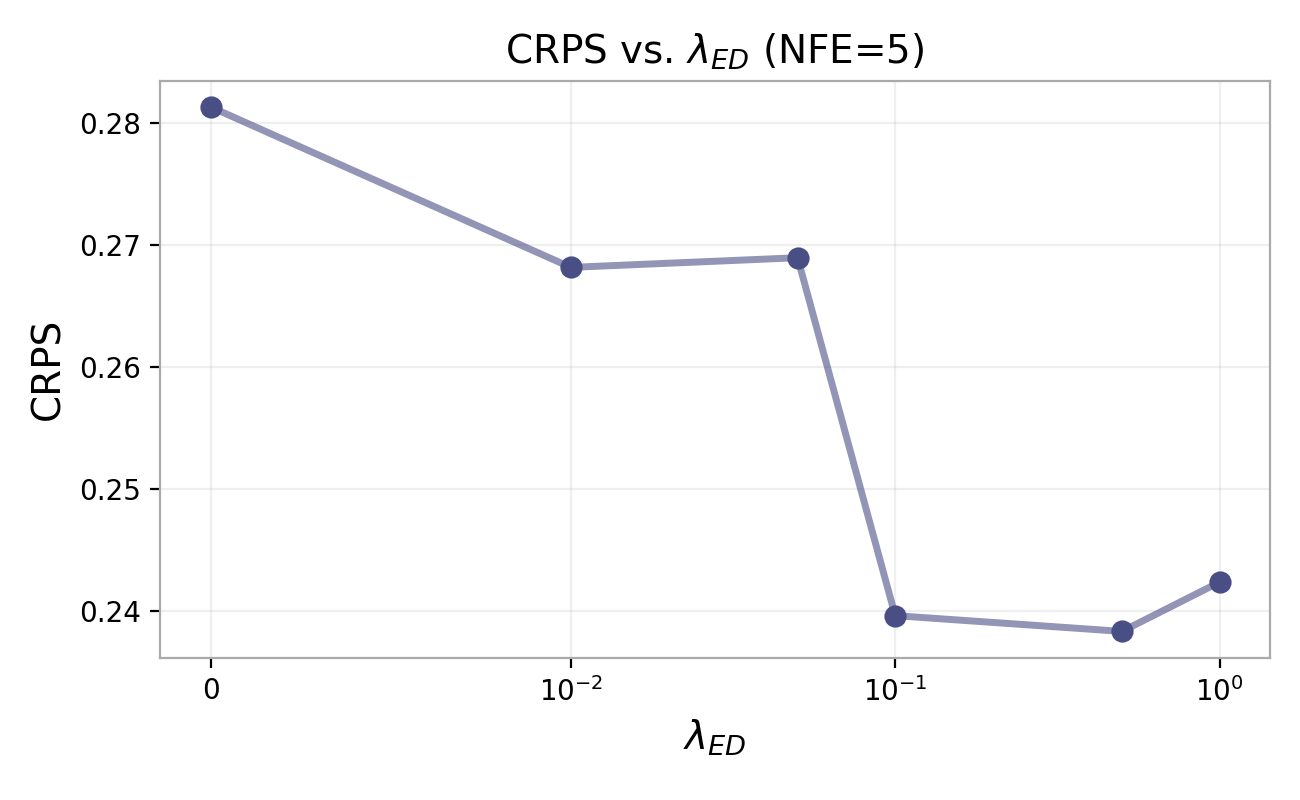}

    \end{subfigure}
    
    \vspace{1em} 

    \begin{subfigure}[b]{0.48\textwidth}
        \centering
        \includegraphics[width=\linewidth]{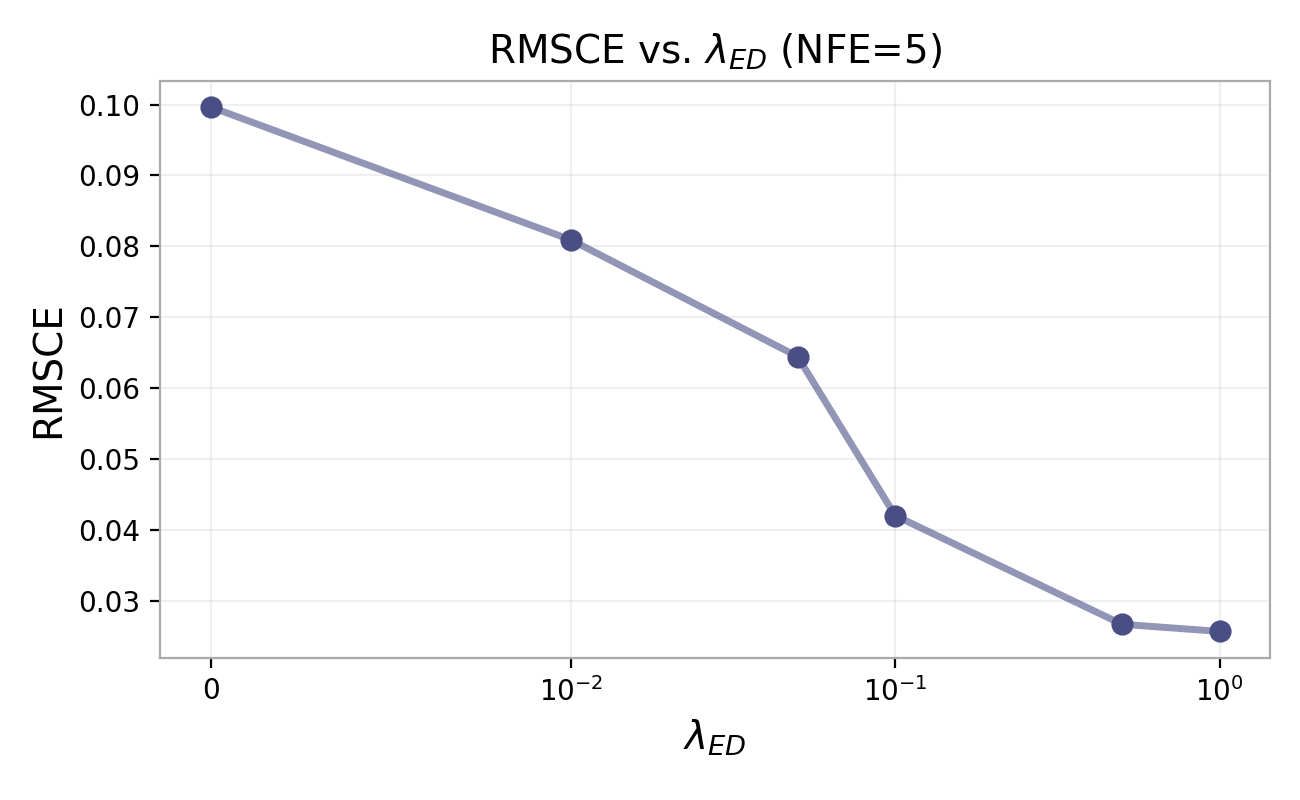}

    \end{subfigure}
    \hfill
    \begin{subfigure}[b]{0.48\textwidth}
        \centering
        \includegraphics[width=\linewidth]{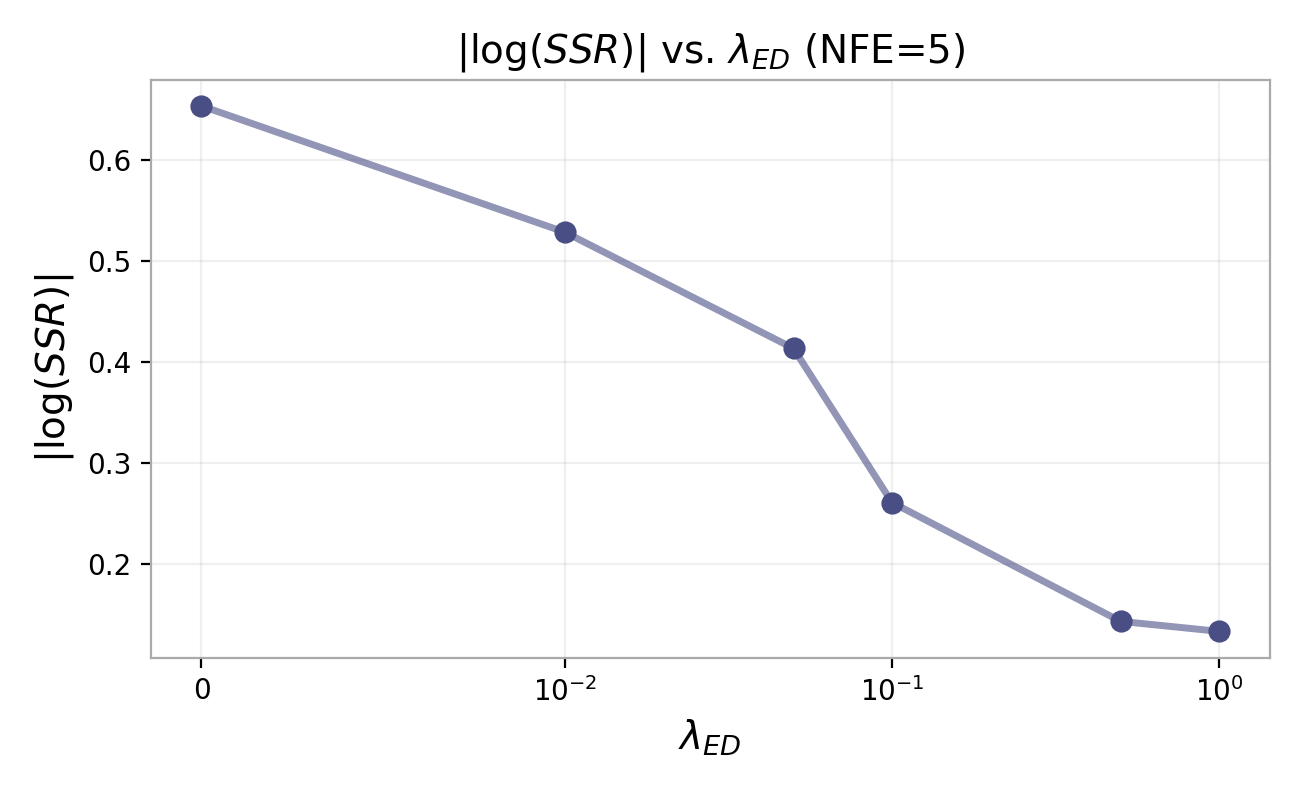}

    \end{subfigure}

    \caption{Hyperparameter sensitivity sweep for $\lambda_{ED} \in [0, 1.0]$. The method exhibits a stability plateau for $\lambda_{ED} \in [0.1, 0.5]$ where calibration is minimized without degrading accuracy.}
    \label{fig:ablation_grid}
\end{figure}

To further investigate the sensitivity to the ED regularization, we fix $\lambda_{\text{var}} = 1$ and train the model on a finer grid of $\lambda_{\text{ED}}$ values. Figure~\ref{fig:ablation_grid} presents the averaged  results with a rollout time of $T=33$ for models trained with $\lambda_{\text{var}}=1$ and $\lambda_{\text{ED}}\in[0,1]$. 

The results reveal two distinct performance regimes separated by a sharp break between $\lambda_{\text{ED}}=0.05$ and $0.1$ consistently across all metrics. Performance on MSE and CRPS plateaus after this point showing worsening results for $\lambda_{\text{ED}}=1.0$ indicating the onset of over-regularization. Gains in RMSCE and SSR performance start flattening after the $\lambda_{\text{ED}} = 0.1$ point. This study suggests that, while superior performance could be obtained with $\lambda_{\text{ED}} = 0.5$, our choice of $\lambda_{\text{ED}} = 0.1$ is not a narrowly tuned optimum, but rather the critical entry point into the effective regularization regime.

\subsection{Impact of VAE calibration}
\label{ap:vae-beta}

 To explore how the calibration of the VAE posterior affects the calibration of TORNADO predictions, we trained four VAE instances with $\beta\in\{10^{-7},10^{-6},10^{-5},10^{-4}\}$ and one TORNADO model per VAE under identical settings for the Kolmogorov dataset. We test the average RMSCE TORNADO's one step predictive ensemble of size $K=16$. We compare it to the average VAE reconstruction RMSCE. To obtain the predictive ensemble we encode each frame and sample the posterior $K=16$ times and decoding each sample. Results are in Table \ref{tab:vae}.

\vspace{0.25cm}
\begin{table}[h]
\centering
\caption{Calibration results for the VAE's reconstruction task and TORNADO's one-step prediction task averaged over the validation set for different values of $\beta$}
\begin{tabular}{lcc}
\toprule
$\beta$ & VAE RMSCE $\downarrow$ & TORNADO 1-step RMSCE $\downarrow$ \\
\midrule
$10^{-7}$ & 0.284 & 0.098 \\
$10^{-6}$ & \textbf{0.100} & \textbf{0.073} \\
$10^{-5}$ & 0.169 & 0.148 \\
$10^{-4}$ & 0.218 & 0.201 \\
\bottomrule
\label{tab:vae}
\end{tabular}
\end{table}

Indeed, the gains depend on the quality of the VAE posterior, and an appropriate trade-off must be achieved. Two failure modes can be identified.
When $\beta$ is too small, the posterior collapses toward a Dirac distribution (autoencoder regime), leaving TORNADO with no meaningful variance target to match. Nevertheless, the learned stochasticity still improves calibration compared to the raw VAE posterior.
Conversely, when $\beta$ is too large, the posterior becomes overly smooth, inducing over-dispersed targets and degrading calibration.

Interestingly, in all regimes, TORNADO manages to improve the one step forecast ensemble calibration with respect to the underlying VAE reconstruction calibration baseline, suggesting that the latter does not represent a lower bound for the calibration of TORNADO. 

Importantly, however, we observe that the VAE setting yielding the best-calibrated latent posterior also leads to the best-calibrated predictor. This suggests that TORNADO effectively inherits and refines the uncertainty structure of the latent representation when the latter is well balanced.

\subsection{Impact of posterior quality on rollouts}
\label{ap:vae-rollout}

We repeat the sweep of Appendix~\ref{ap:vae-beta} with independently trained VAEs and TORNADO models
for the same values of $\beta$, and evaluate 20-step rollouts on Kolmogorov flow (NFE${=}5$, $K{=}16$).
Table~\ref{tab:vae_beta_sweep} reports the
reconstruction error and KL divergence of each VAE, and the rollout MSE, RMSCE and sharpness of the
corresponding TORNADO model.

\begin{table}[h]
\centering
\caption{VAE regularization sweep on Kolmogorov flow: latent statistics of each VAE and 20-step rollout
metrics of the TORNADO model trained on it (NFE${=}5$, $K{=}16$).}
\label{tab:vae_beta_sweep}
\begin{tabular}{cccccc}
\toprule
$\beta$ & Recon.\ vRMSE $\downarrow$ & KL (nats/sample) & Rollout MSE $\downarrow$ & Rollout RMSCE $\downarrow$ & Sharpness $\downarrow$ \\
\midrule
$10^{-7}$ & 0.0018 & 11031.9 & 0.0216 & 0.0514 & 0.0907 \\
$10^{-6}$ & 0.0024 & 5376.7  & 0.2060 & 0.0393 & 0.4301 \\
$10^{-5}$ & 0.0084 & 2004.0  & 0.2177 & 0.0287 & 0.3467 \\
$10^{-4}$ & 0.0357 & 635.4   & 2.0723 & 0.0388 & 1.7371 \\
\bottomrule
\end{tabular}
\end{table}

As expected, increasing $\beta$ raises the reconstruction error and lowers the KL divergence. At $\beta=10^{-7}$,
TORNADO attains the lowest rollout MSE, but its ensembles are the narrowest and its calibration error
exceeds that obtained for $\beta\in\{10^{-6},10^{-5}\}$, consistent with the near-deterministic posterior
discussed in Appendix~\ref{ap:vae-beta}. At $\beta=10^{-4}$, the rollout MSE increases by an order of
magnitude and the ensemble spread by a factor of four relative to $\beta=10^{-6}$, while RMSCE remains
low. The low calibration error is then a consequence of the inflated spread, so RMSCE must be read
together with accuracy and sharpness (Appendix~\ref{ap:pits}).

The range $\beta\in[10^{-6},10^{-5}]$ combines low reconstruction error, low calibration error and
moderate spread. This agrees with Appendix~\ref{ap:vae-beta}, where the best one-step calibration is
obtained at $\beta=10^{-6}$: both sweeps identify the same regime. In practice, we recommend a
logarithmic sweep over $\beta$, retaining the range in which reconstruction remains accurate and
calibration and sharpness are preserved. All methods in the main text use $\beta=10^{-6}$; this value
is not tuned per method.

\subsection{Comparison of higher order solvers}
\label{ap:solvers}
To discern the impact of the choice of ODE or SDE solver on the final results, we test TORNADO and FM with higher-order solvers: Heun (for FM) and a derivative-free Milstein scheme \citep{Kloeden2010-pi} (for TORNADO), alongside their standard counterparts Euler and EM. Results\footnote{This solver comparison uses a different TORNADO checkpoint from Table~\ref{tab:megatable}.} for Kolmogorov at $T=20$ are in Table \ref{tab:solvers}. 

\vspace{0.25cm}
\begin{table}[h]
\centering
\caption{Results for Kolmogorov at $T=20$  with different ODE/SDE solvers for TORNADO (Euler-Maruyama and Milstein) and FM (Euler and Heun)}
\begin{tabular}{lcccc}
\toprule
Method & MSE $\downarrow$ & CRPS $\downarrow$ & RMSCE $\downarrow$ & $|\log(\mathrm{SSR})|$ $\downarrow$ \\
\midrule
TORNADO-EM       & 0.155 & 0.167 & \textbf{0.053} & \textbf{0.376} \\
TORNADO-Milstein & 0.155 & 0.167 & \textbf{0.053} & 0.376 \\
FM-Euler         & \textbf{0.107} & \textbf{0.163} & 0.203 & 1.500 \\
FM-Heun          & \textbf{0.106} & 0.161 & 0.195 & 1.404 \\
\bottomrule
\label{tab:solvers}
\end{tabular}
\end{table}

Higher-order solvers do not change the conclusions: FM-Heun improves slightly over FM-Euler yet remains worse than TORNADO-EM, and Milstein yields identical results to EM. This is consistent with a smooth learned diffusion field, where higher-order stochastic corrections are negligible.

\subsection{Ablation on gradient stopping}
\label{ao:detach}
To motivate the design choice of stopping the diffusion head's gradients from propagating through the shared trunk, we report on Table \ref{tab:detach_ablation} the average performance at NFE=5 and $T=10$ for models trained on the Kolmogorov Flow with and without adding a detach (gradient stopping) to the diffusion head's input.

\begin{table}[h]
\centering
\caption{Gradient-stop ablation (detach diffusion head) on Kolmogorov Flow at NFE=5,
averaged over rollout horizon $T=10$. Lower is better.}
\label{tab:detach_ablation}
\begin{tabular}{lcccc}
\toprule
Setting & MSE $\downarrow$ & CRPS $\downarrow$ & RMSCE $\downarrow$ & $|\log(\mathrm{SSR})| \downarrow$ \\
\midrule
w/ detach  & 0.0195 & 0.0662 & 0.0309 & 0.2334 \\
w/o detach & 0.0203 & 0.0672 & 0.0304 & 0.2296 \\
\bottomrule
\end{tabular}
\end{table}

Detaching the diffusion head gradients from the shared trunk has negligible effect on both accuracy and calibration  at NFE=5 on Kolmogorov Flow, suggesting that the drift learning is largely unaffected by whether diffusion gradients backpropagate into the trunk. We therefore treat the diffusion head as a modular add-on and keep gradient stopping as a conservative default to reduce potential gradient interference, without affecting performance. We leverage this property by only training the diffusion head once the drift learning has converged, reducing the training cost significantly by limiting the sample based ED computations.

\subsection{Computing time comparison}
\label{ap:compute}

Admittedly, the implementation of TORNADO entails the addition of an extra component to the shared trunk model used by the baselines. Namely, we require the use of an extra diffusion head that models the diffusion term of our SDE. On Table \ref{tab:compute_cost} we report the average physical time step transition time ($t \rightarrow t+1$) in function of the number of integrator steps (NFE). The time is measured after a GPU warmup of 50 steps, averaged over 200 steps and 3 repetitions. To compare with other experiments, the measurement is taken for a batch size of 32 and an ensemble size of 16 with the Kolmogorov Flow data set. Measurements are taken on a NVIDIA H100 80GB HBM3.
\begin{table}[H]
\centering
\caption{Compute cost (ms per forecast step) at batch size $B=32$ and ensemble size $K=16$. Lower is better.}

\begin{tabular}{lccc}
\toprule
Method & NFE=2 & NFE=5 & NFE=10 \\
\midrule
DDIM    & 555.00  & 1387.70 & 2773.38 \\
FM      & 555.16  & 1387.80 & 2775.52 \\
SI      & 554.70  & 1387.57 & 2775.01 \\
TORNADO & 602.41  & 1505.95 & 3011.94 \\
\bottomrule
\end{tabular}%

\label{tab:compute_cost}
\end{table}
Overall, the addition of the diffusion head incurs on a compute time around $8.5\%$ higher than that of the drift model. We deem this added cost acceptable in trade-off for the improved calibration of the predictions.

\subsection{Effect of $\sigma_\gamma$ on calibration}
\label{ap:sigma_scale_calibration}

The stochastic interpolant in Eq. \ref{eq:interpolant} injects Gaussian noise with variance $\gamma(\tau) = \sigma_\gamma\tau(1-\tau)$. This hyperparameter affects both the sampled intermediate states $z_{\tau}$ and the closed-form conditional targets $\mu_{1\mid \tau}$ and $\Sigma_{1 \mid \tau}$ used for supervision. In particular, it controls how informative $z_{\tau}$ is about the endpoint $z_1$ through the conditional covariance $\Sigma_{1 \mid \tau}$.

\paragraph{Theoretical observation (see Appendix~\ref{ap:sigma_scale_theory}, Corollary~\ref{cor:sigma_tau_effects}.)}
For fixed $\tau\in(0,1)$, increasing $\sigma_{\gamma}$ monotonically increases the conditional endpoint covariance
$\Sigma_{1\mid\tau}$, interpolating from a low-noise regime to a noise-dominated regime where
$\Sigma_{1\mid\tau}\to \Sigma_1$ as $\sigma_{\gamma}\to\infty$.
Moreover, for any fixed $\sigma_{\gamma}>0$, we have $\Sigma_{1\mid\tau}\to\Sigma_1$ as $\tau\to 0^+$ and
$\Sigma_{1\mid\tau}\to 0$ as $\tau\to 1^-$. 
This suggests a direct mechanism by which $\sigma_{\gamma}$ influences predictive dispersion: increasing $\sigma_{\gamma}$ increases $\Sigma_{1\mid \tau}$, reducing the contraction of uncertainty about the endpoint induced by conditioning on $z_{\tau}$. Motivated by this, we empirically evaluate calibration and dispersion as $\sigma_{\gamma}$ varies.

\begin{figure}[ht]
    \centering
    \includegraphics[width=0.6\linewidth]{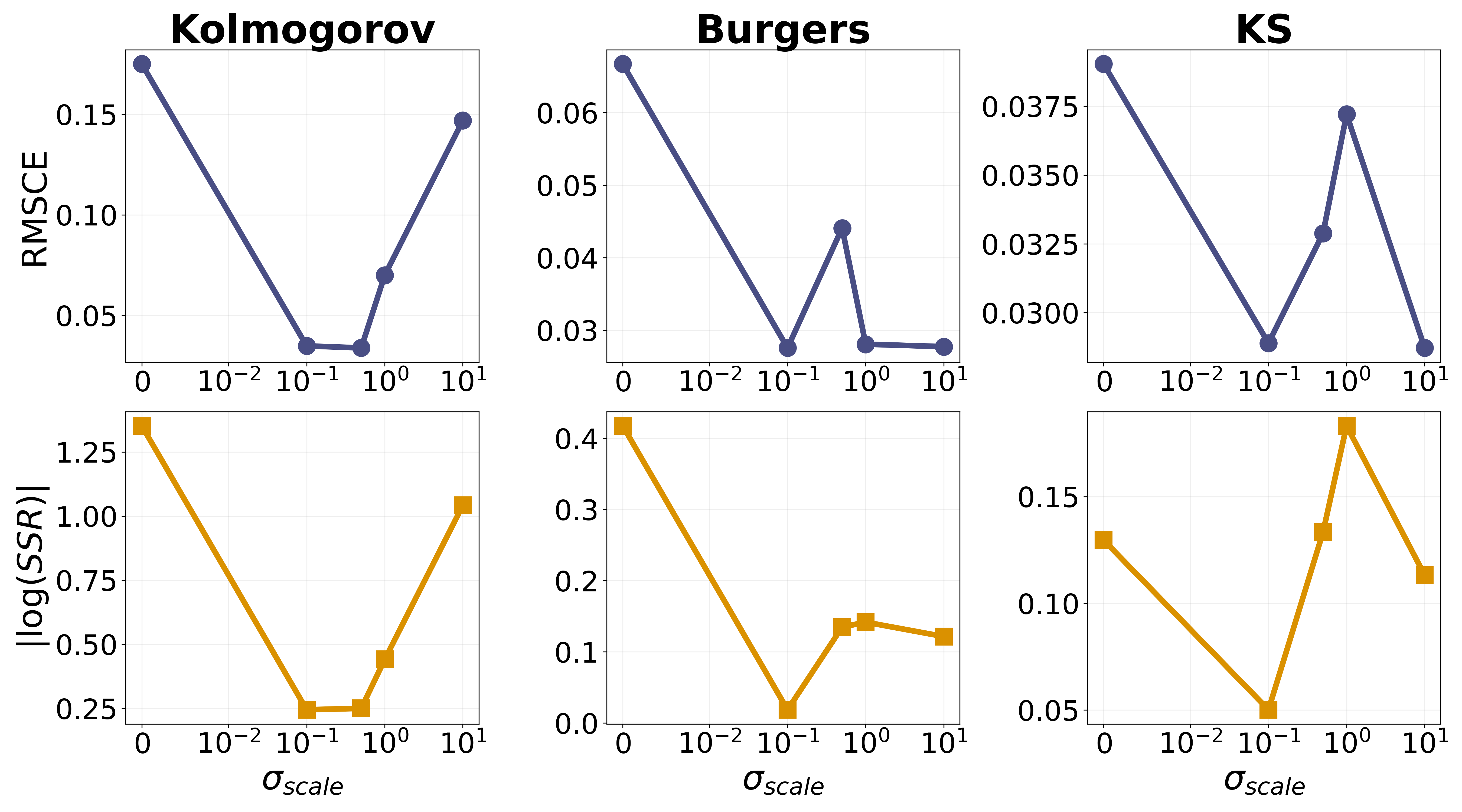}
    \caption{RMSCE and $\lvert \log (\text{SSR})\rvert$ at $\sigma_\gamma \in \{0,0.1,0.5,1,10.0\}$ for the Kolmogorov Flow, Burgers and KS datasets averaged over 10 rollout steps.}
    \label{fig:noise_ab}
\end{figure}

Figure \ref{fig:noise_ab} reports averaged RMSCE and $\lvert \log (\text{SSR}) \rvert$ over 10 rollout steps, spatial grid, and batch elements for predictions of models trained with $\sigma_\gamma \in \{0,0.1,0.5,1,10.0\}$. The effect of $\sigma_\gamma$ is asymmetrical across PDEs; however, a consistent trend can be observed: at $\sigma_\gamma=0$, models exhibit poor calibration and under-dispersion. At larger values of $\sigma_\gamma$, while calibration improves, SSR indicates over-spread predictions. Overall, the experimental findings are consistent with the theoretical results and moderate noise levels provide the best trade-off, we chose $\sigma_\gamma=0.1$ in all experiments.

\subsection{Ensemble size $K$}
\label{ap:ensemble-size}
\begin{figure}[h]
    \centering
    \includegraphics[width=0.7\linewidth]{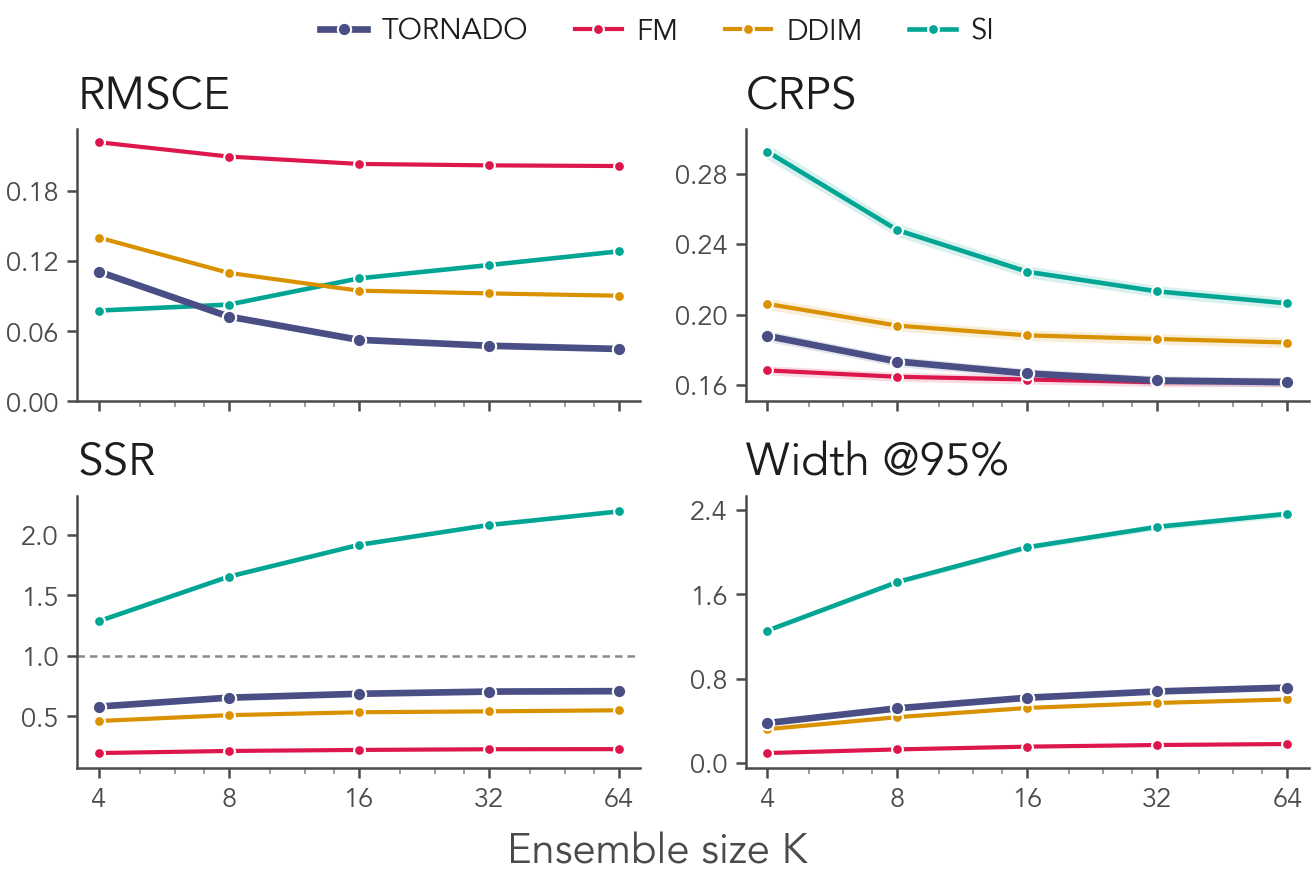}
    \caption{Calibration, predictive skill, and spread across ensemble
    sizes. Lower RMSCE and CRPS are better; the target spread--skill
    ratio is one.}
    \label{fig:ensemble_size}
\end{figure}

We explore the impact of the inference ensemble size $K$ choice. In Figure~\ref{fig:ensemble_size} we sweep
$K \in \{4, 8, 16, 32, 64\}$ on Kolmogorov flow at NFE${=}5$ over 20
autoregressive steps, reusing the trained models and the forecasting protocol
unchanged, only the number of sampled members varies. Increasing the ensemble size improves TORNADO's calibration, with RMSCE
falling from about 0.11 at $K=4$ to 0.045 at $K=64$. TORNADO has the lowest RMSCE from $K=8$ onwards, with most of the improvement occurring by $K=16$, showing it is a reasonable choice in the trade off between performance and cost. Its CRPS also decreases, approaching FM's at larger ensemble sizes. The spread--skill ratio moves towards one but remains below it,
while the 95\% intervals widen moderately. DDIM improves more modestly,
and FM retains the narrowest intervals and highest calibration error.
SI's CRPS improves with larger ensembles, but its calibration worsens
as its intervals widen and its spread--skill ratio rises further above one.

\section{Additional experiments} \label{app:ad-exp}
\subsection{Extension to Auto-Regressive training}
\label{ap:AR}

While the main paper adopts one-step teacher-forced training as a controlled simplification, TORNADO is not restricted to this setting. We additionally train autoregressive variants that unroll the latent dynamics over horizons $H\in\{2,4\}$ during training. Starting from the posterior of the current frame, the model is recursively rolled forward and re-anchored at each step using the encoder posterior of the predicted state, while the same local drift, variance, and energy-distance objectives are applied at every step with a decaying weight over the rollout horizon. This extends TORNADO beyond teacher forcing while preserving its local conditional supervision. As shown in Table~\ref{tab:ar_training}, the autoregressive variants retain the gains of the base method and further improve calibration, with $H=4$ also giving the strongest overall performance.

\begin{table}[h]
\centering
\caption{Autoregressive extension of TORNADO on the Kolmogorov Flow dataset. Extending training beyond teacher forcing preserves the benefits of the base method and further improves calibration, with $H=4$ achieving the best overall performance.}
\begin{tabular}{lcccc}
\toprule
Method / training regime & MSE $\downarrow$ & CRPS $\downarrow$ & RMSCE $\downarrow$ & $|\log(\mathrm{SSR})|$ $\downarrow$ \\
\midrule
\textbf{TORNADO (TF)}       & 0.120 & 0.145 & 0.0395 & 0.268 \\
\textbf{TORNADO-AR} $(H=2)$ & 0.139 & 0.157 & 0.0368 & 0.225 \\
\textbf{TORNADO-AR} $(H=4)$ & \textbf{0.111} & \textbf{0.141} & \textbf{0.0318} & \textbf{0.180} \\
\bottomrule
\end{tabular}

\label{tab:ar_training}
\end{table}

\subsection{Comparison to one-step baselines}
\label{ap:one-step}

We additionally consider three simple latent baselines that bypass auxiliary-time transport and predict the next latent state directly in a single step. \textbf{MC Dropout} predicts the next latent state with a standard MSE loss and relies on stochastic dropout at inference to produce ensemble uncertainty. \textbf{Gaussian NLL} takes the current posterior statistics $(\mu_t,\sigma_t)$ as input and predicts the mean and diagonal standard deviation of a Gaussian next-state distribution, trained by Gaussian negative log-likelihood against a sample from the next VAE posterior. \textbf{VAE Matching} uses the same parameterization, but is trained to match the mean and variance of the next VAE posterior directly through a Gaussian KL. Unlike TORNADO, these baselines do not introduce an auxiliary-time stochastic transport between consecutive latent posteriors. The results for the Kolmogorov dataset averaged on a 20 step rollout are shown in Table~\ref{tab:one-step}

\begin{table}[h]
    \centering
    \caption{Comparison with direct one-step latent baselines on the Kolmogorov Flow dataset.}
    \begin{tabular}{lcccc}
    \toprule
    Method & MSE $\downarrow$ & CRPS $\downarrow$ & RMSCE $\downarrow$ & $|\log(\mathrm{SSR})|$ $\downarrow$ \\
    \midrule
    \textbf{TORNADO (NFE=5)} & \textbf{0.120} & \textbf{0.145} & \textbf{0.039} & \textbf{0.268} \\
    NLL                      & 0.326 & 0.243 & 0.192 & 2.344 \\
    VAE matching             & 0.687 & 0.360 & 0.199 & 2.628 \\
    MC Dropout               & 0.321 & 0.250 & 0.075 & 0.447 \\
    \bottomrule
    \end{tabular} 
    \label{tab:one-step}
\end{table}

Across all three baselines, TORNADO yields better calibrated predictive distributions and more stable autoregressive rollouts, showing that direct one-step latent uncertainty prediction is not sufficient to match the benefits of adaptive stochastic transport.  We do not consider approaches such as a deep ensemble comparison because they would introduce substantially different training, storage, and parameter budgets from our controlled comparison.

\subsection{Long-horizon statistical evaluation via total enstrophy}
\label{ap:enstrophy}

At long rollout horizons, pointwise trajectory agreement becomes increasingly difficult to interpret in chaotic systems. Following the long-horizon statistical evaluation protocol of \citet{Zighed_Thome_Gallinari_Sayadi_2026}, we therefore compare the distributions of a coarse scalar statistic over fixed rollout windows. For the Kolmogorov flow, total enstrophy is a natural choice, as it measures the global intensity of vorticity fluctuations and remains meaningful beyond the horizon where pointwise comparisons become difficult to interpret. We define the total enstrophy of a snapshot as

\[
Z(\omega_t)=\sum_{i,j,c} \omega_t(i,j,c)^2.
\]

For each window of size 20 in $T=100$ rollouts, we  compare the empirical distributions of predicted and ground-truth enstrophy values over the test set using the 1D Wasserstein distance.

\begin{figure}[h]
    \centering
    \includegraphics[width=0.8\linewidth]{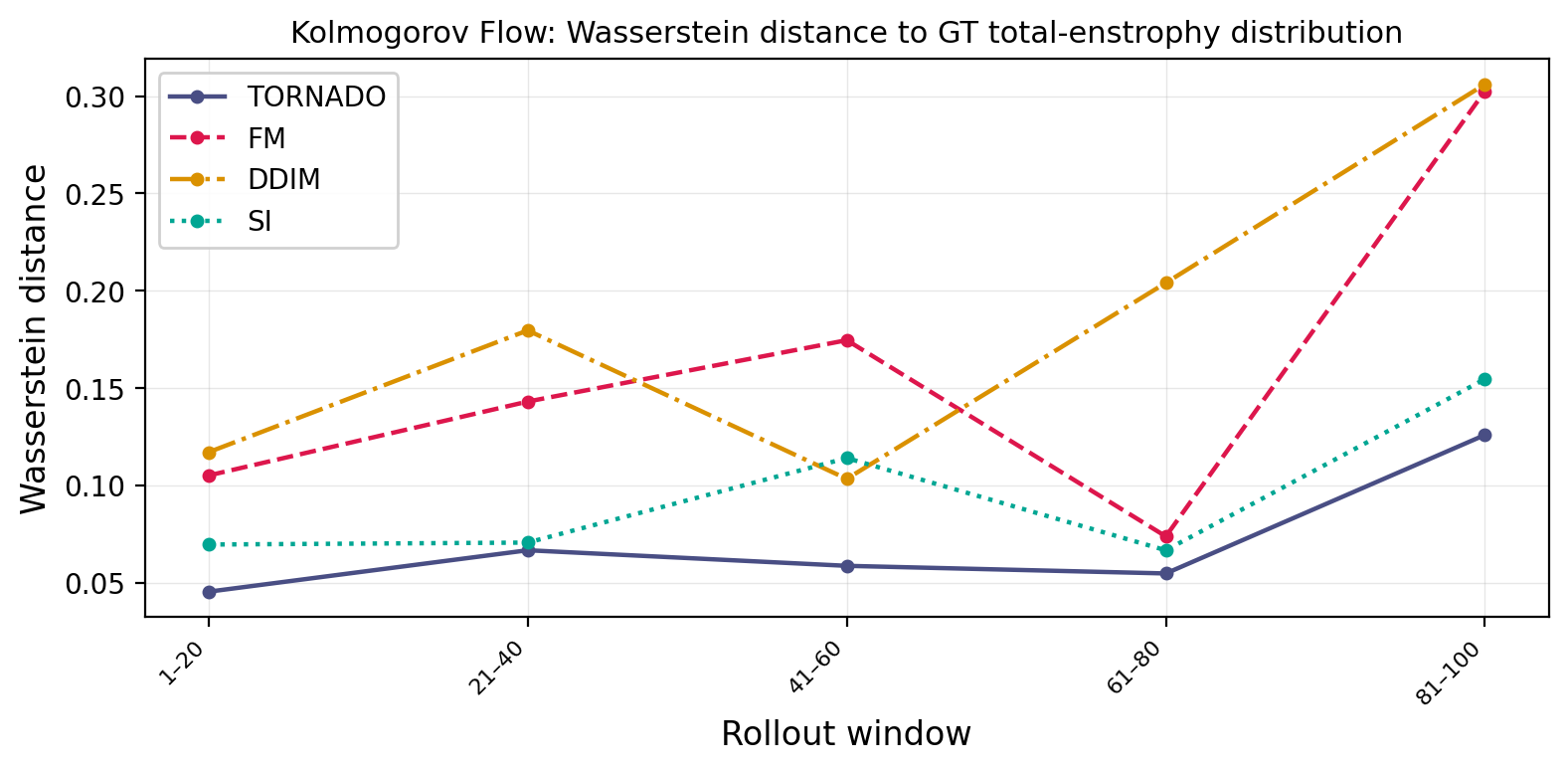}
    \caption{1D Wasserstein distance between the predicted and ground-truth total-enstrophy distributions over the rollout windows of length 20, aggregated over the test set for Kolmogorov Flow over 100 step trajectories. Lower is better.}
    \label{fig:enstrophy_wasserstein}
\end{figure}

Figure~\ref{fig:enstrophy_wasserstein} reports the 1D Wasserstein distance between the predicted and ground-truth total-enstrophy distributions across 20-step rollout windows. TORNADO achieves the smallest distance in all windows, while SI is generally the next closest method and FM and DDIM exhibit larger statistical drift, particularly at later horizons. Under this coarse statistical evaluation, TORNADO is therefore the most consistent method over long rollouts on the Kolmogorov dataset.

While the Wasserstein analysis quantifies distributional agreement aggregated over the test set, it does not reveal how each method's predictive band tracks the ground truth along an individual rollout. Calibrated probabilistic forecasting requires not only that the marginal distribution of physical statistics match in aggregate, but that the predictive band actually contain the realized trajectory at the rate implied by its nominal coverage. 
\begin{figure}[h!]
    \centering
    \includegraphics[width=0.48\linewidth]{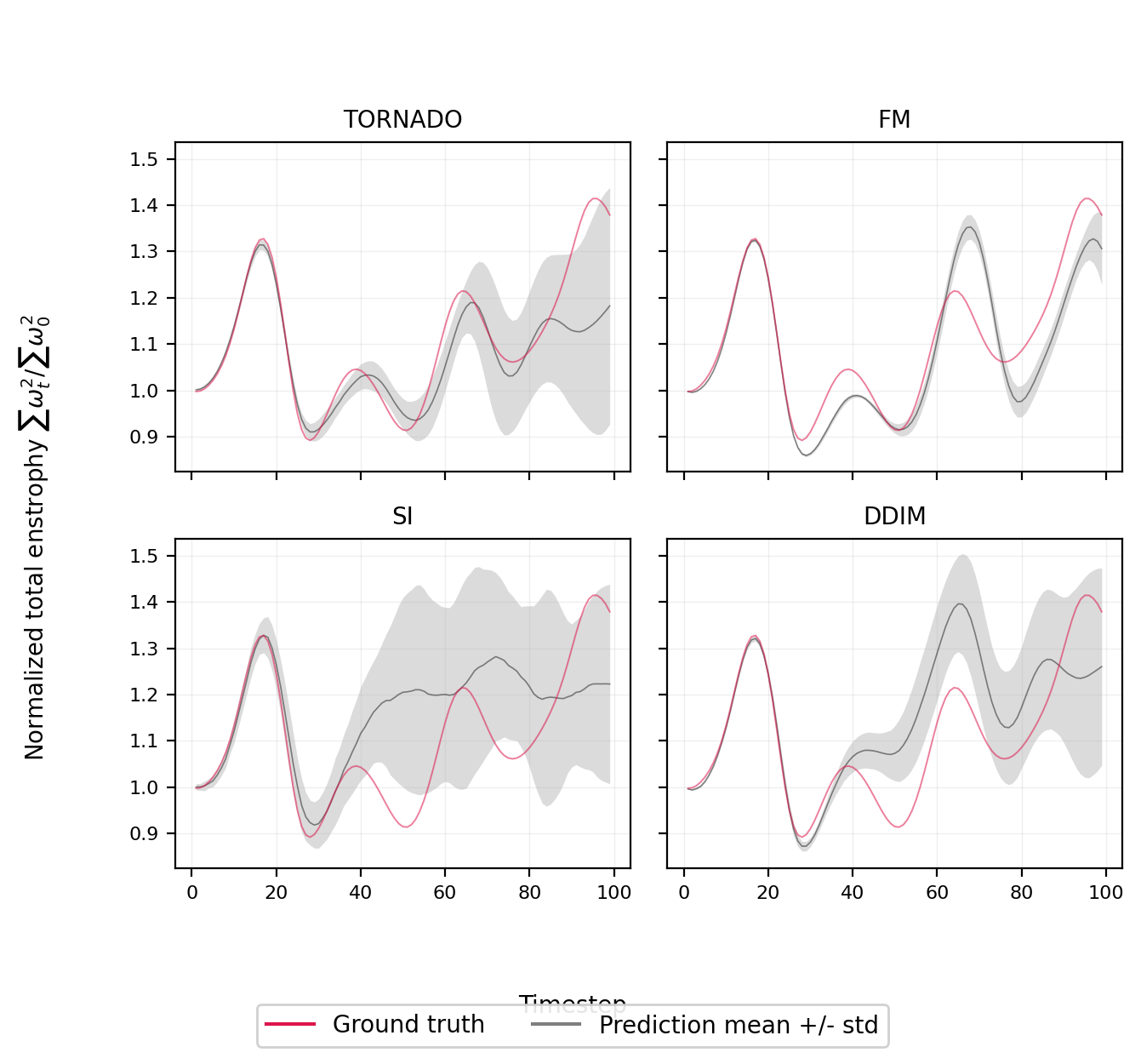}
    \hfill
    \includegraphics[width=0.48\linewidth]{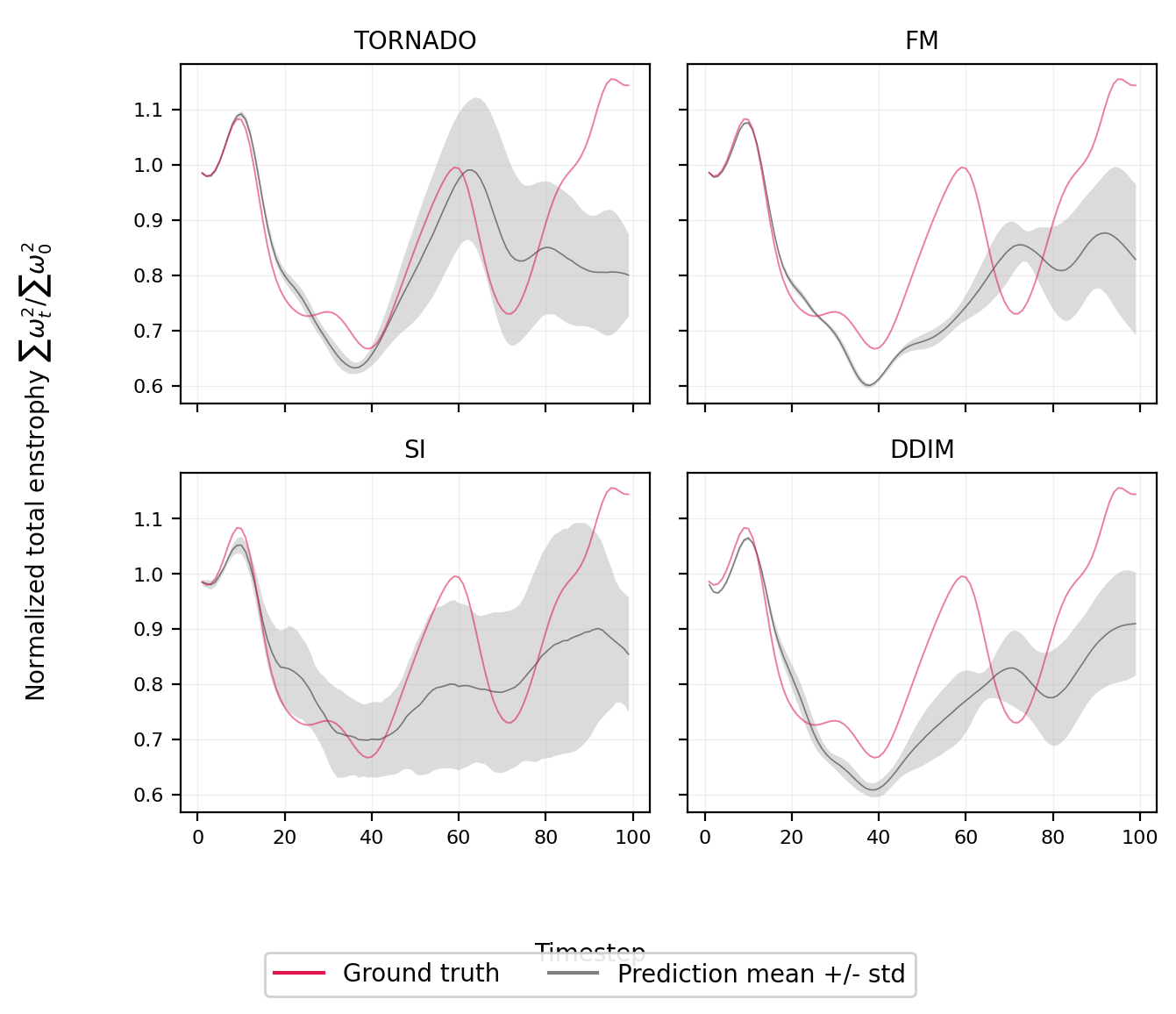}
    \caption{Normalized total enstrophy over two representative Kolmogorov flow rollouts. Red curves show the ground truth; gray curves and shaded regions show the predictive ensemble mean and one standard deviation for each method.}
    \label{fig:ens_traj}
\end{figure}
As a qualitative complement, in Figure~\ref{fig:ens_traj}  we examine the evolution of normalized total enstrophy $Z(t)/Z(0)$ along one illustrative ground-truth trajectory together with the predictive ensemble mean and standard deviation for each method. TORNADO tracks the evolution of this coarse statistic closely over the first half of the rollout, and as its mean prediction begins to drift, the ensemble spread widens accordingly. FM also follows the early dynamics well, but remains comparatively underdispersed once it departs from the ground-truth trajectory. SI exhibits the opposite behavior, with a broader spread already present from the start. DDIM departs from the ground-truth statistics earlier and remains less well aligned throughout the rollout.

\subsection{Calibration beyond RMSCE}
\label{ap:pits}

RMSCE and $|\log(\mathrm{SSR})|$ collapse some aspects of calibration into one scalar. On one hand, good calibration and SSR can be achieved at low accuracy by widening the predictive intervals. While these metrics were reported alongside MSE and CRPS, which constitute a measure of accuracy and sharpness, the analysis can be made more explicit by looking at sharpness measures, like the width of the predictive ensemble. On the other hand, by collapsing the RMSCE into a scalar and taking the log absolute value of the SSR, we lose the information on whether miscalibration comes from underspreading (U-shaped PIT, SSR $< 1$) or overspreading (bell-shaped PIT, SSR $> 1$).

\begin{figure}[h]
    \centering
    \includegraphics[width=0.7\linewidth]{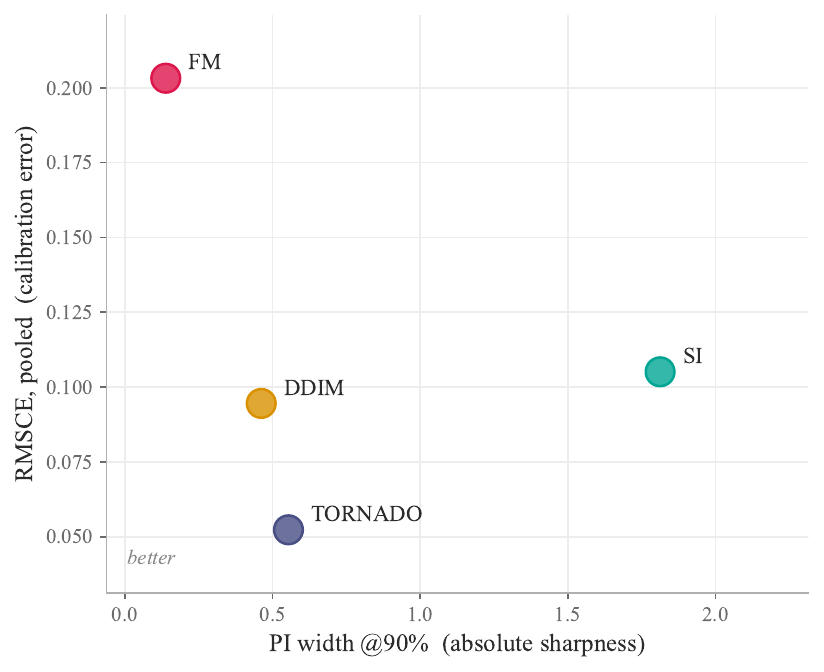}
    \caption{20 step pooled calibration error versus mean 90\% predictive interval width on Kolmogorov flow.}
    \label{fig:pareto}
\end{figure}

\paragraph{Absolute sharpness.}
SSR measures spread relative to error: a value near one indicates that the two match, but does not imply narrow intervals. We therefore report mean predictive interval width alongside calibration error for Kolmogorov flow with NFE${=}5$, $K{=}16$, and 20 rollout steps. Table~\ref{tab:sharpness_calibration} reports 95\% interval widths and coverage, while Figure~\ref{fig:pareto} plots pooled RMSCE against mean 90\% interval width. 

TORNADO has the lowest calibration error, and its 95\% intervals are $3.3\times$ narrower than SI's. Its calibration advantage therefore does not come from wider intervals. FM has the narrowest intervals, but they cover only 28.4\% of observations at the nominal 95\% level, and its calibration error is the largest. TORNADO's coverage is higher at 80.0\%, though still below the 95\% target. Interval width must therefore be read alongside coverage. Figure~\ref{fig:pareto} shows no clear trade-off between calibration and sharpness: TORNADO lies toward the lower-left corner, while the baselines vary across both axes.

\begin{table}[h]
\centering
\caption{Calibration, sharpness, and predictive skill on Kolmogorov flow.}
\label{tab:sharpness_calibration}
\begin{tabular}{lccccc}
\toprule
Method & RMSCE $\downarrow$ & PI width@95\% & cov.@95\% & SSR $\rightarrow 1$ & CRPS $\downarrow$ \\
\midrule
DDIM    & 0.095         & 0.521 & 0.632 & 0.533          & 0.188 \\
FM      & 0.203         & 0.156 & 0.284 & 0.225          & 0.163 \\
SI      & 0.106         & 2.037 & 0.990 & 1.921          & 0.224 \\
TORNADO & \textbf{0.039} & 0.624 & 0.800 & \textbf{0.765} & \textbf{0.145} \\
\bottomrule
\end{tabular}
\end{table}

\paragraph{Per-level calibration.}
We assess calibration using the rank of the ground truth among the $K$
predictive samples, normalised to $[0,1]$, as an empirical probability integral
transform (PIT). For a calibrated ensemble, these ranks are uniformly
distributed over the available rank positions. We also examine observed
coverage $\hat{G}(p)$ at each nominal level $p$. The RMSCE defined in
Appendix~\ref{ap:metrix} summarizes the root-mean-square deviation of this
coverage curve from the diagonal.

\begin{figure}[h]
    \centering
    \includegraphics[width=0.7\linewidth]{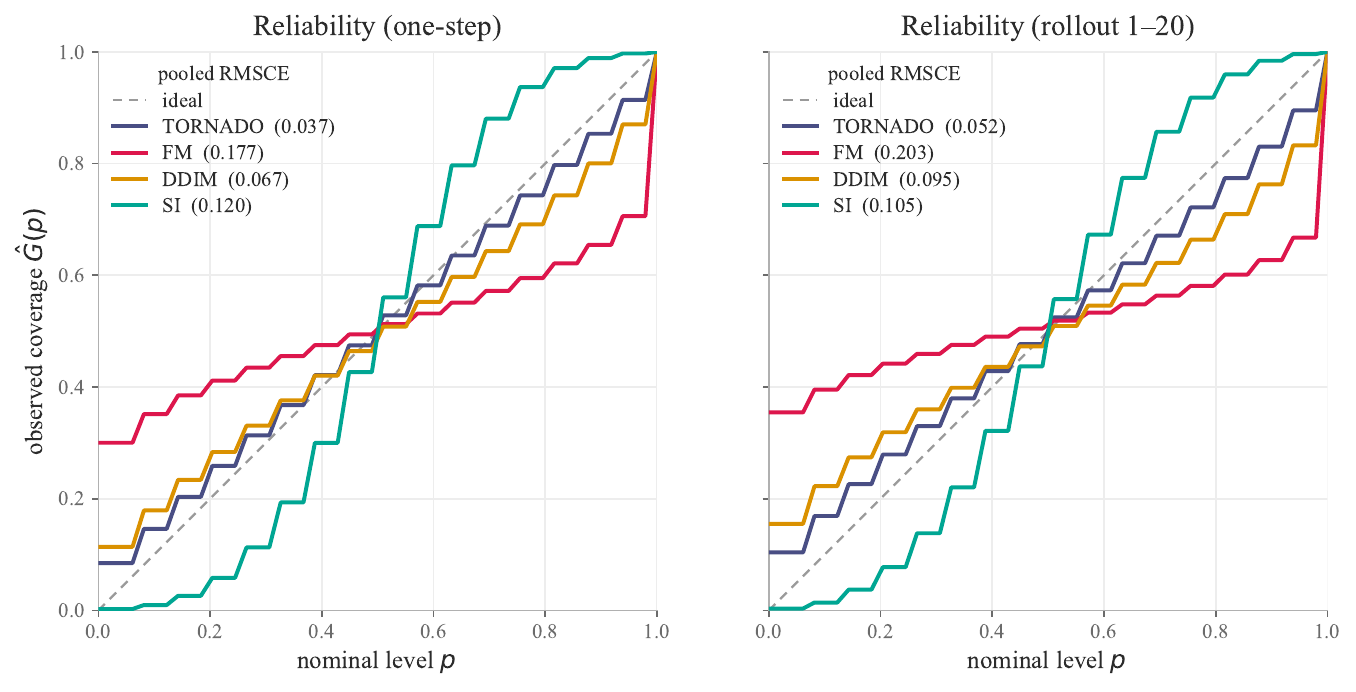}
    \caption{Reliability diagrams on Kolmogorov flow (NFE${=}5$, $K{=}16$), one-step
(left) and over the 20-step rollout (right). Curves above the diagonal
over-cover, below under-cover.}
    \label{fig:reliability}
\end{figure}

RMSCE measures the size of calibration errors but not their direction.
A U-shaped rank histogram indicates insufficient ensemble spread, while
a center-peaked histogram indicates excessive spread. The corresponding
coverage curves show whether predictive intervals under-cover or over-cover
at each nominal level.

\begin{figure}[h]
    \centering
    \includegraphics[width=\linewidth]{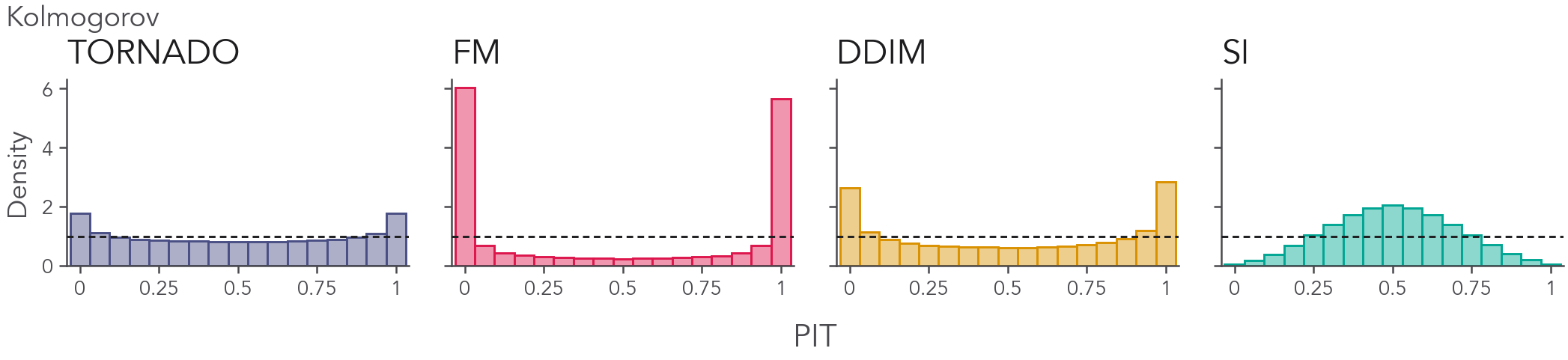}
    \caption{PIT histograms on Kolmogorov flow NFE${=}5$, $K{=}16$, 20 rollout
steps}

    \label{fig:pit-k}
\end{figure}

Figures~\ref{fig:reliability} and~\ref{fig:pit-k} show this difference for SI
and DDIM. Their RMSCE values are similar, but SI's histogram is centre-peaked 
and its PIT CDF lies below the diagonal at low quantiles and above it at high quantiles, consistent with excessive spread. DDIM's histogram
is U-shaped and its PIT CDF exhibits the opposite pattern. TORNADO has the flattest
histogram and the coverage curve closest to the diagonal across the evaluated
levels. Its lower RMSCE thus reflects better calibration across nominal levels,
rather than an improvement confined to a few levels.

\begin{figure}[h]
    \centering
    \includegraphics[width=\linewidth]{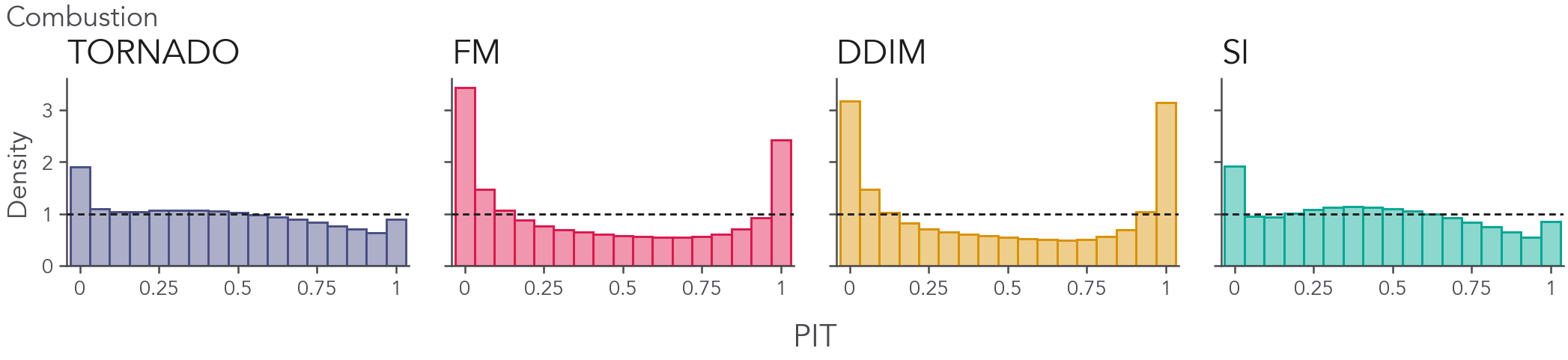}
    \caption{PIT histograms on Combustion NFE${=}5$, $K{=}16$, 20 rollout
steps}

    \label{fig:pit-c}
\end{figure}

\begin{figure}[h]
    \centering
    \includegraphics[width=\linewidth]{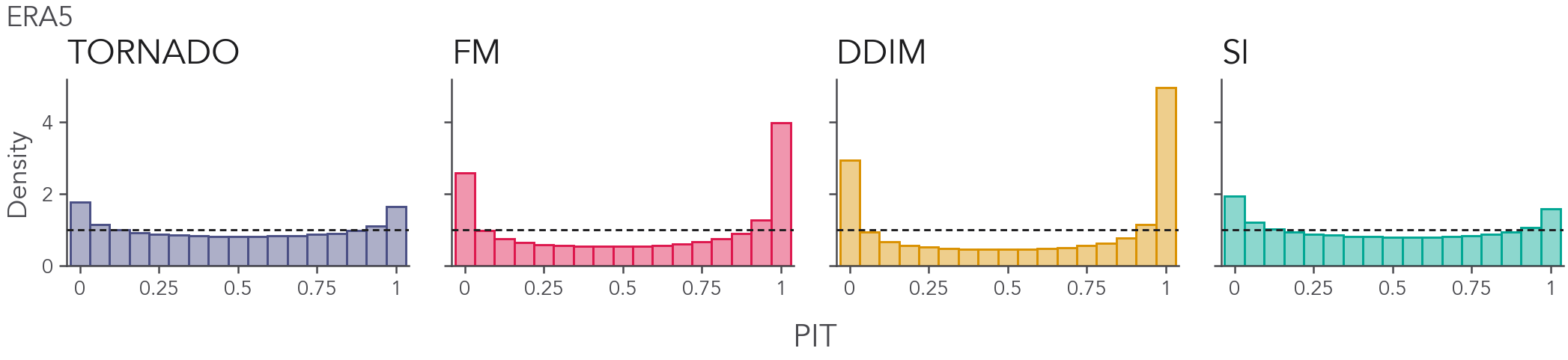}
    \caption{PIT histograms on ERA5 NFE${=}5$, $K{=}16$, 20 rollout
steps}

    \label{fig:pit-e}
\end{figure}

Figures~\ref{fig:pit-c} and~\ref{fig:pit-e} extend this analysis to Combustion and ERA5. On both datasets, FM and DDIM exhibit pronounced boundary peaks, consistent with insufficient ensemble spread, whereas TORNADO and SI have substantially flatter histograms. On Combustion, both retain an excess of low PIT values, indicating that observations fall below their predictive ensembles more often than expected. On ERA5, both show moderate excess mass at the boundaries, suggesting residual underdispersion. Unlike on Kolmogorov flow, SI does not exhibit a pronounced centre peak, and its rank profile is similar to TORNADO's. These results show that the direction and severity of miscalibration depend on the dataset, while TORNADO maintains comparatively balanced rank profiles across all three settings.

\subsection{Error vs Variance allocation}
\label{ap:error-variance}

To assess how each model allocates uncertainty relative to its own errors, we compute pointwise MSE and ensemble variance in latent space over the first 10 rollout steps. For each method, spatial points are grouped into low-, medium-, and high-error terciles based on that model’s MSE.
As shown in Table~\ref{tab:tercile_comparison}, both methods assign higher variance to higher-error regions. However, TORNADO shows a tighter correspondence between variance and error across terciles. SI remains globally over-dispersed, especially in low-error regions, whereas TORNADO keeps the variance-MSE ratio much better aligned with its errors. This suggests that learned diffusion improves  calibration primarily through better magnitude control.

\begin{table}[h!]
\centering
\caption{Within-method tercile analysis of latent-space error and variance for TORNADO and SI.}
\label{tab:tercile_comparison}
\scalebox{0.9}{
\begin{tabular}{lccc ccc}
\toprule
MSE tercile & \multicolumn{3}{c}{TORNADO} & \multicolumn{3}{c}{SI} \\
\cmidrule(lr){2-4} \cmidrule(lr){5-7}
(method specific) & MSE & Var & Var/MSE & MSE & Var & Var/MSE \\
\midrule
low error  & 0.0004 & 0.0061 & 15.0  & 0.0008 & 0.0897 & 111.4 \\
mid error & 0.0040 & 0.0081 & 2.0   & 0.0080 & 0.1180 & 14.8  \\
high error& 0.0628 & 0.0215 & 0.34  & 0.1109 & 0.2998 & 2.70  \\
\bottomrule
\end{tabular}
}
\end{table}

\subsection{ERA5 per-variable results}
\label{ap:era5_pervar}

Table~\ref{tab:megatable} aggregates the four ERA5 variables into a single score.
Table~\ref{tab:era5_pervar} reports the same metrics per variable, computed from
the same rollouts. Metrics in Table~\ref{tab:megatable} are computed in normalised units with $\cos(\text{latitude})$ weighting. MSE and CRPS are averaged over the four variables, so they equal the means of Table~\ref{tab:era5_pervar}, whereas RMSCE and SSR are computed after pooling variables (one rank histogram over all variables, and spread and error pooled across variables), so they are not averages of the per-variable scores.

The pattern is identical on
every field: SI and TORNADO achieve similar MSE and CRPS on all four fields, while TORNADO attains the lowest
calibration error and the $|\log(\mathrm{SSR})|$ closest to zero on all four.
The aggregate is therefore not driven by a single atmospheric quantity.
\begin{table*}[h]
    \centering
    \caption{ERA5 per-variable results on the held-out years (NFE${=}5$, $K{=}16$, steps 1--40, 70 test trajectories). Metrics are in normalised, $\cos(\mathrm{lat})$-weighted units; mean $\pm$ std over the test set. \textbf{Best} and \underline{second-best} per variable and metric are highlighted (\textcolor{BestGreenText}{green}/\textcolor{SecondBlueText}{blue}).}

    \setlength{\tabcolsep}{6pt}
    \renewcommand{\arraystretch}{1.08}
    \resizebox{\columnwidth}{!}{%
    \begin{tabular}{llcccc}
        \toprule
        \textbf{Variable} & \textbf{Model}
        & MSE $(10^{-3})\,(\downarrow)$
        & CRPS $(10^{-2})\,(\downarrow)$
        & RMSCE $(10^{-2})\,(\downarrow)$
        & $|\log(\mathrm{SSR})|\,(\downarrow)$ \\
        \midrule
        \midrule

        \multirow{4}{*}{\textbf{$T_{2m}$}} 
        & DDIM    & \merr{13.116}{2.634} & \merr{5.60}{0.44} & \merr{17.72}{2.16} & \merr{0.787}{0.051} \\
        & FM      & \merr{13.367}{2.569} & \merr{5.52}{0.43} & \merr{15.91}{2.22} & \merr{0.637}{0.046} \\
        & SI      & \best{\textbf{\merr{11.604}{2.214}}} & \best{\textbf{\merr{4.90}{0.38}}} & \second{\underline{\merr{8.44}{1.48}}} & \second{\underline{\merr{0.292}{0.045}}} \\
        & TORNADO & \second{\underline{\merr{11.885}{2.212}}} & \second{\underline{\merr{4.94}{0.35}}} & \best{\textbf{\merr{7.53}{1.21}}} & \best{\textbf{\merr{0.265}{0.042}}} \\
        \midrule

        \multirow{4}{*}{\textbf{$U_{10}$}} 
        & DDIM    & \merr{451.1}{40.7} & \merr{35.06}{1.67} & \merr{12.35}{0.51} & \merr{0.655}{0.044} \\
        & FM      & \merr{460.3}{48.3} & \merr{35.08}{1.89} & \merr{11.00}{0.65} & \merr{0.578}{0.048} \\
        & SI      & \second{\underline{\merr{417.4}{37.0}}} & \second{\underline{\merr{32.16}{1.51}}} & \second{\underline{\merr{5.71}{0.63}}} & \second{\underline{\merr{0.268}{0.035}}} \\
        & TORNADO & \best{\textbf{\merr{416.7}{38.7}}} & \best{\textbf{\merr{32.14}{1.51}}} & \best{\textbf{\merr{5.48}{0.55}}} & \best{\textbf{\merr{0.255}{0.039}}} \\
        \midrule

        \multirow{4}{*}{\textbf{$V_{10}$}} 
        & DDIM    & \merr{641.8}{55.1} & \merr{41.68}{1.83} & \merr{12.04}{0.49} & \merr{0.632}{0.044} \\
        & FM      & \merr{645.3}{53.3} & \merr{41.39}{1.80} & \merr{10.67}{0.47} & \merr{0.564}{0.041} \\
        & SI      & \best{\textbf{\merr{587.5}{46.9}}} & \best{\textbf{\merr{38.16}{1.56}}} & \second{\underline{\merr{5.63}{0.49}}} & \second{\underline{\merr{0.263}{0.036}}} \\
        & TORNADO & \second{\underline{\merr{591.4}{47.1}}} & \second{\underline{\merr{38.25}{1.50}}} & \best{\textbf{\merr{5.44}{0.39}}} & \best{\textbf{\merr{0.255}{0.033}}} \\
        \midrule

        \multirow{4}{*}{\textbf{$Z_{500}$}} 
        & DDIM    & \merr{48.052}{8.057} & \merr{11.13}{1.06} & \merr{24.07}{3.46} & \merr{0.798}{0.069} \\
        & FM      & \merr{51.090}{9.939} & \merr{10.76}{1.15} & \merr{18.76}{4.15} & \merr{0.669}{0.081} \\
        & SI      & \best{\textbf{\merr{43.264}{6.449}}} & \best{\textbf{\merr{9.29}{0.71}}} & \second{\underline{\merr{9.16}{3.08}}} & \second{\underline{\merr{0.250}{0.055}}} \\
        & TORNADO & \second{\underline{\merr{43.940}{7.255}}} & \second{\underline{\merr{9.30}{0.77}}} & \best{\textbf{\merr{7.73}{2.57}}} & \best{\textbf{\merr{0.219}{0.059}}} \\

        \bottomrule
    \end{tabular}%
    }
    \label{tab:era5_pervar}
\end{table*}

\clearpage

\subsection{NFE effect} \label{ap:nfe}

\begin{figure}[h]
    \centering
    \includegraphics[width=\linewidth]{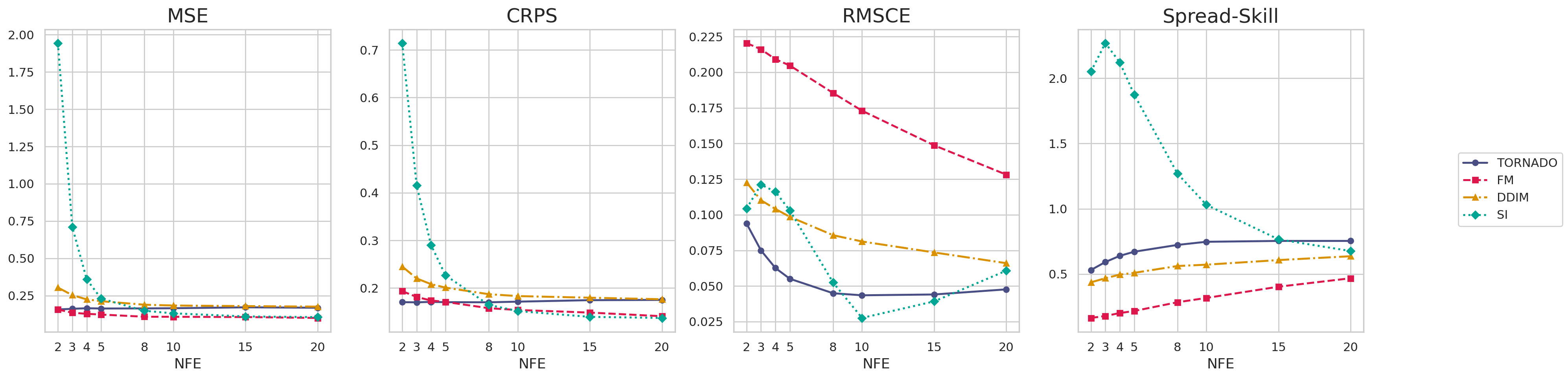}
    \caption{
NFE sensitivity on Kolmogorov Flow. We report rollout MSE, CRPS, RMSCE, and SSR as a function of NFE used by method. Metrics are averaged over the same rollout horizon $T=50$.
}
    \label{fig:sweep}
\end{figure}

Here we show similar plots as in \ref{sec:long} for $\text{NFE}=2$ and $10$. Increased integration steps benefits baselines while TORNADO is comparatively NFE-insensitive in calibration. In Figure~\ref{fig:sweep} we show the $T=50$ time-averaged metrics as a function of NFE. TORNADO remains stable across the tested range, with MSE and CRPS varying only mildly and calibration strongest at intermediate budgets. This supports treating NFE as a finite sampler budget in the compute-constrained regime rather than as a setting that must be increased for reliable performance. 

\begin{figure*}[h]
    \centering
    \includegraphics[width=\linewidth]{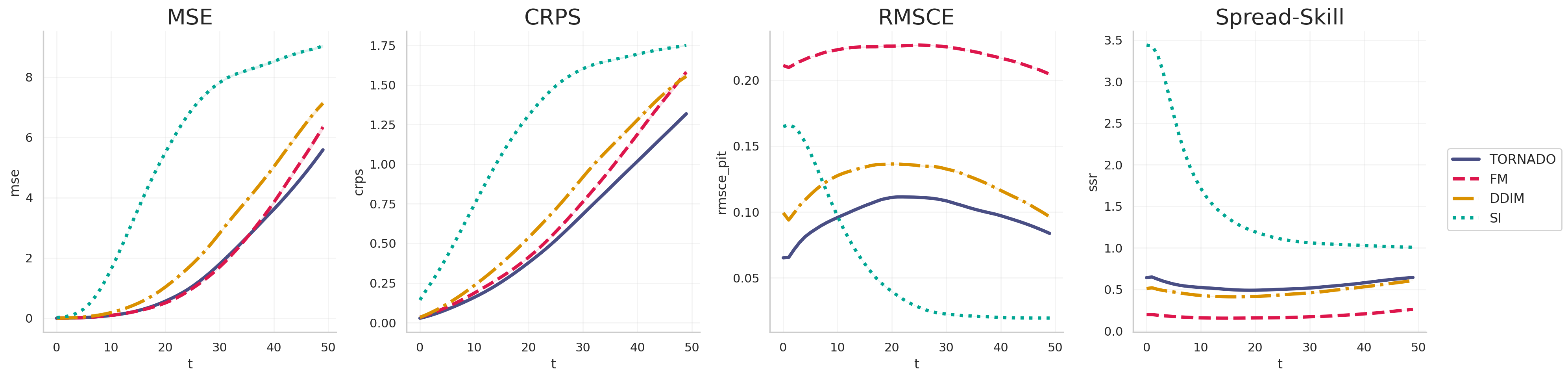}
    \caption{Kolmogorov flow rollout metrics vs.\ horizon $t$ at fixed compute (NFE=2): MSE, CRPS, RMSCE, and SSR.}
    \label{fig:nfe2}
\end{figure*}

\begin{figure*}[h]
    \centering
    \includegraphics[width=\linewidth]{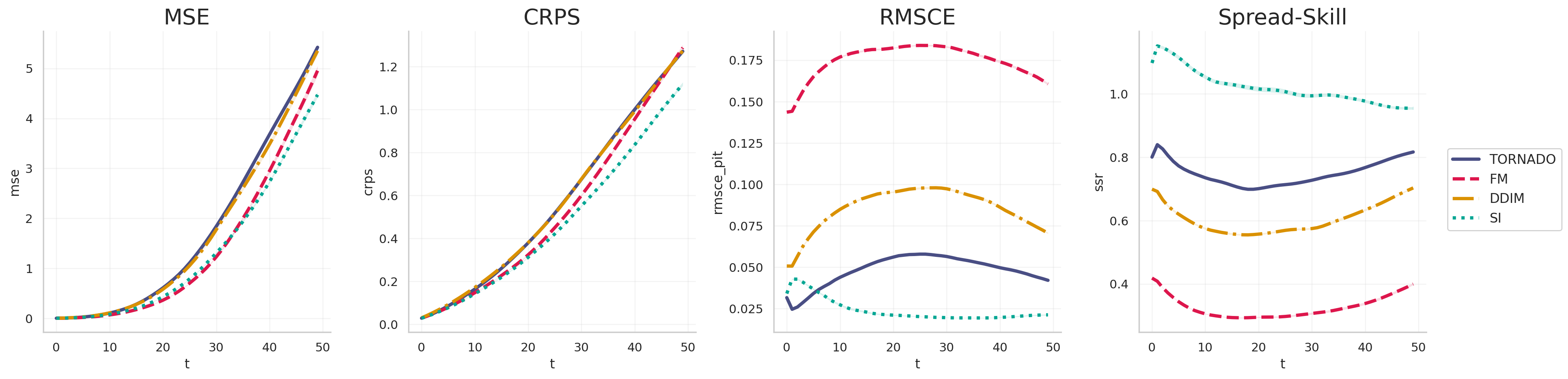}
    \caption{Kolmogorov flow rollout metrics vs.\ horizon $t$ at fixed compute (NFE=10): MSE, CRPS, RMSCE, and SSR.}
    \label{fig:nfe10}
\end{figure*}


\clearpage

\section{Visualizations}
\vspace{4cm}

\begin{figure}[h]
    \centering
    \includegraphics[width=\linewidth]{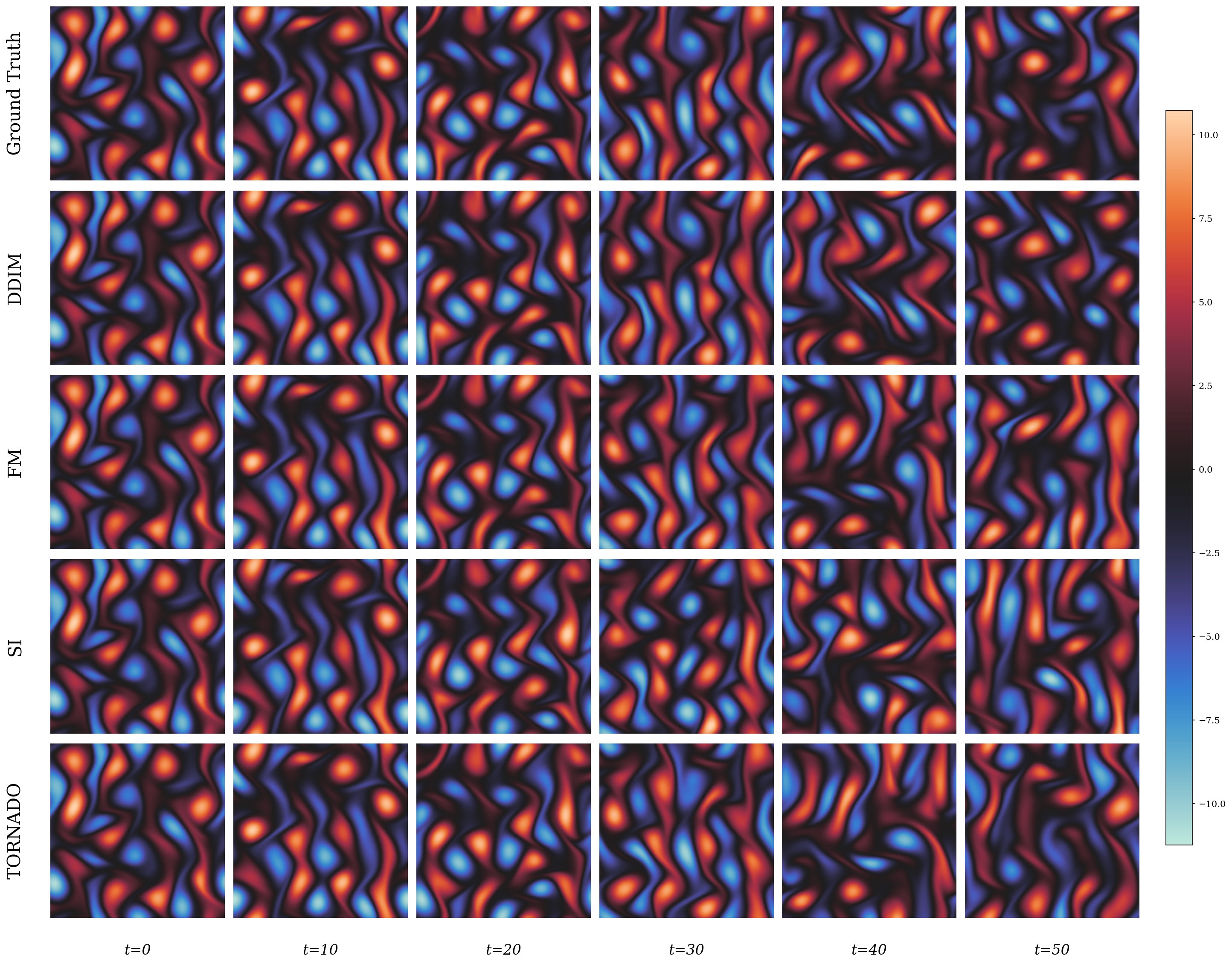}
    \caption{Visualization of rollout predictions at NFE=5 for methods TORNADO, FM, DDIM and SI alongside the ground truth trajectory for the Kolmogorov Flow data set.}
    \label{fig:viz-km}
\end{figure}

\begin{figure}[hp]
    \centering
    \includegraphics[width=\linewidth]{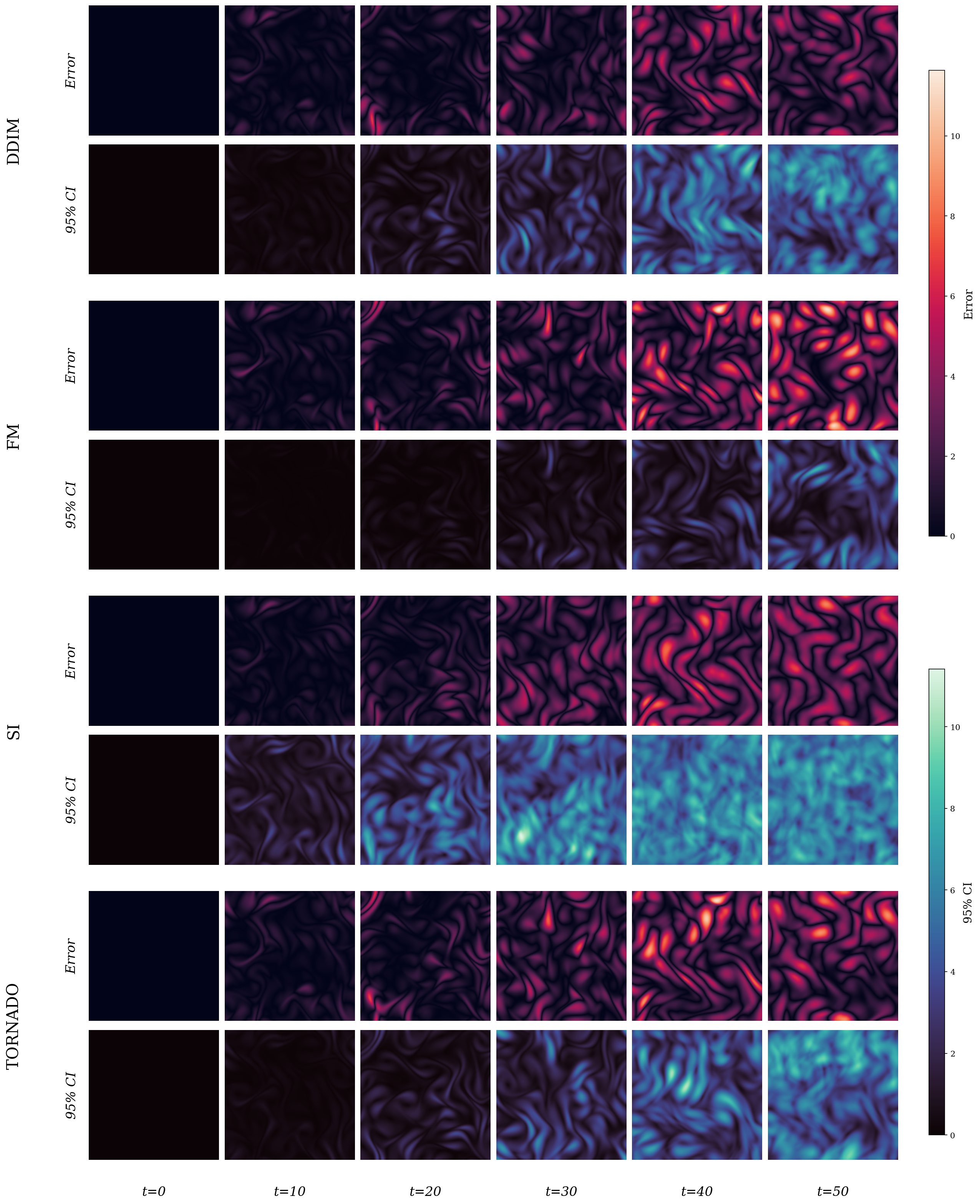}
    \caption{Visualization of the 95\% confidence interval of the predictive ensemble distribution at NFE=5 for methods TORNADO, FM, DDIM and SI alongside the RMSE of the mean of the predictive ensemble distribution with respect to the ground truth for the Kolmogorov Flow data set.}
    \label{fig:viz-km-uq}
\end{figure}

\begin{figure}[hp]
    \centering
    \includegraphics[width=\linewidth]{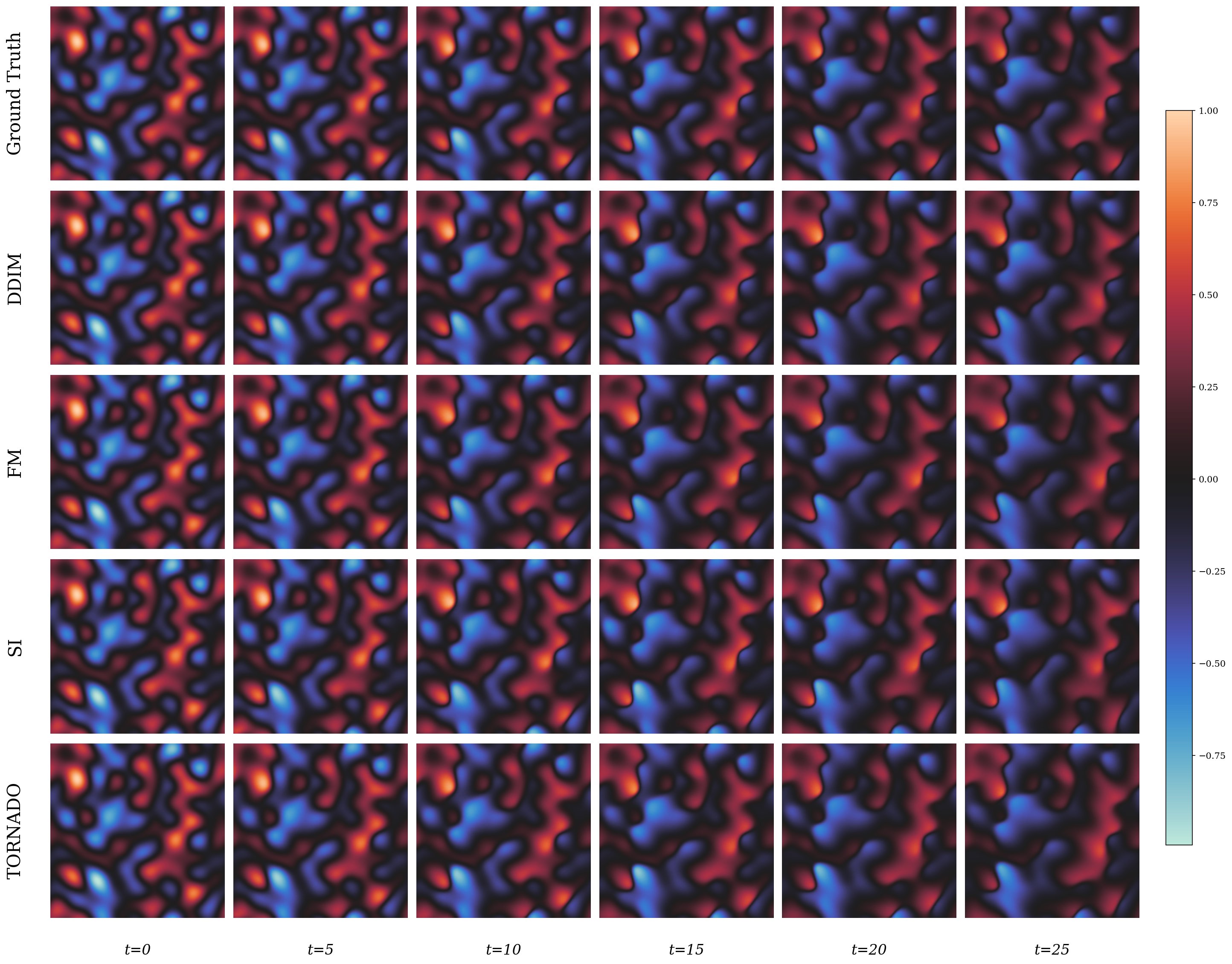}
    \caption{Visualization of rollout predictions at NFE=5 for methods TORNADO, FM, DDIM and SI alongside the ground truth trajectory for the Burgers data set.}
    \label{fig:viz-b}
\end{figure}

\begin{figure}[hp]
    \centering
    \includegraphics[width=\linewidth]{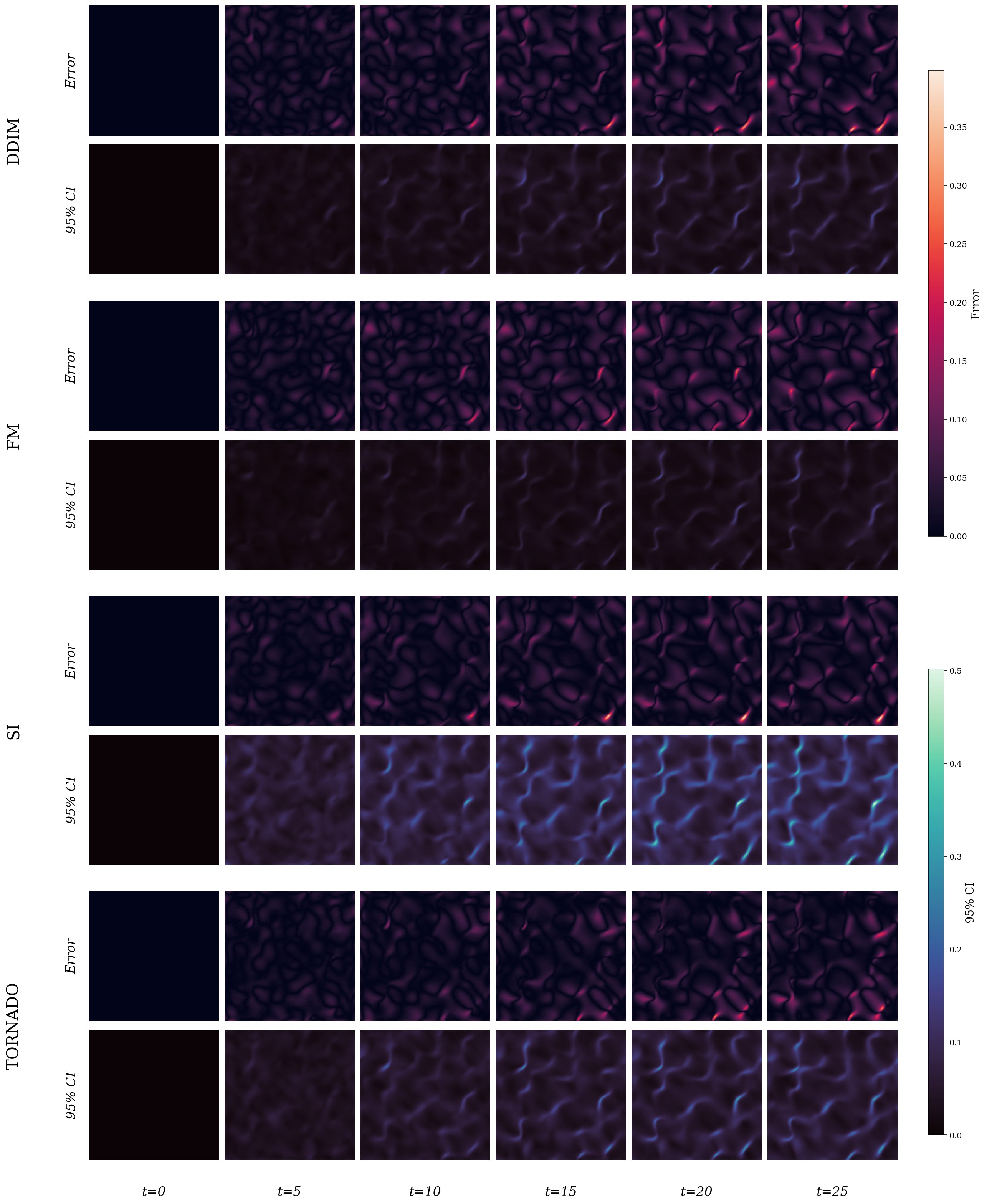}
    \caption{Visualization of the 95\% confidence interval of the predictive ensemble distribution at NFE=5 for methods TORNADO, FM, DDIM and SI alongside the RMSE of the mean of the predictive ensemble distribution with respect to the ground truth for the Burgers data set.}
    \label{fig:viz-b-uq}
\end{figure}

\begin{figure}[hp]
    \centering
    \includegraphics[width=\linewidth]{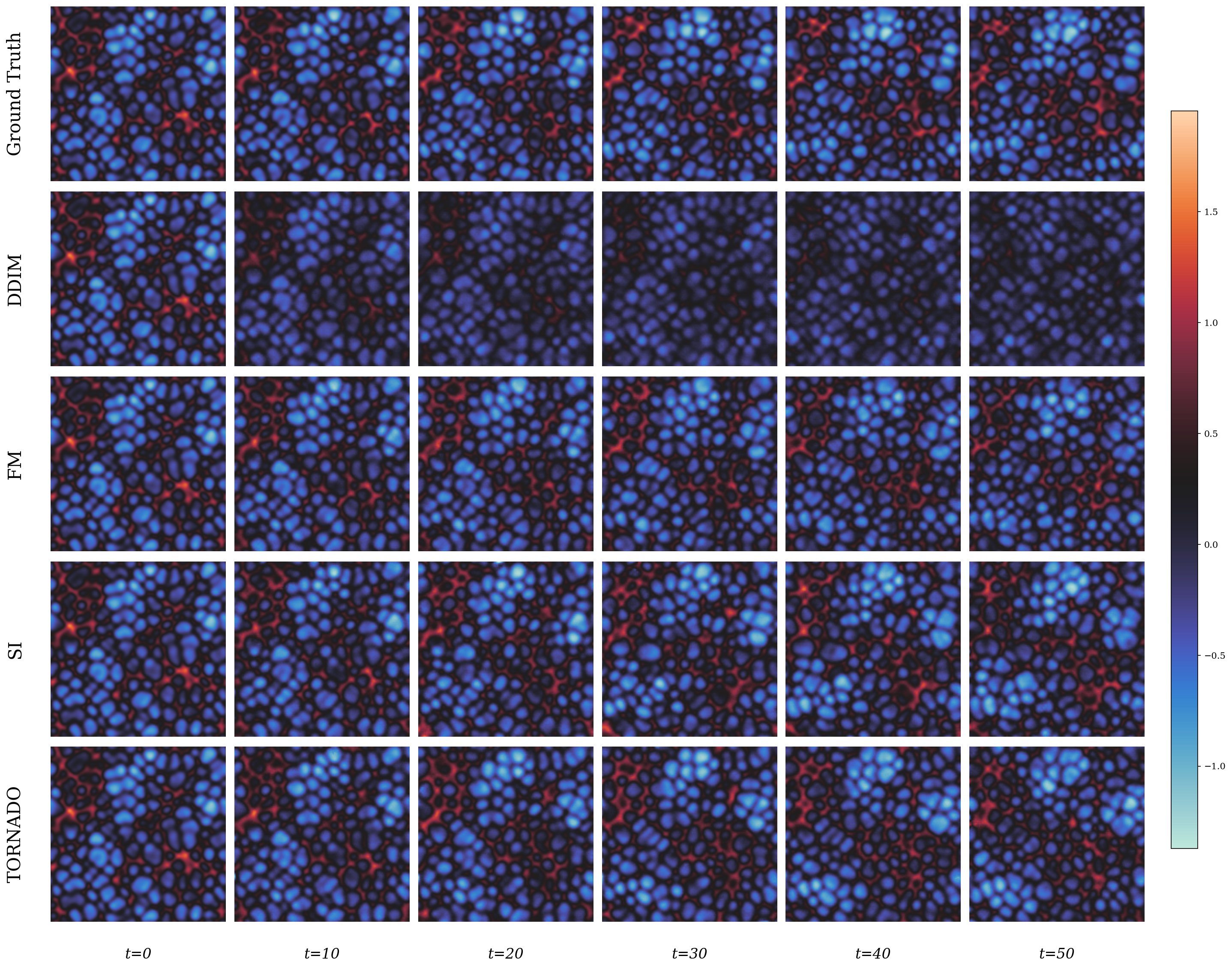}
    \caption{Visualization of rollout predictions at NFE=5 for methods TORNADO, FM, DDIM and SI alongside the ground truth trajectory for the KS data set.}
    \label{fig:viz-ks}
\end{figure}

\begin{figure}[hp]
    \centering
    \includegraphics[width=\linewidth]{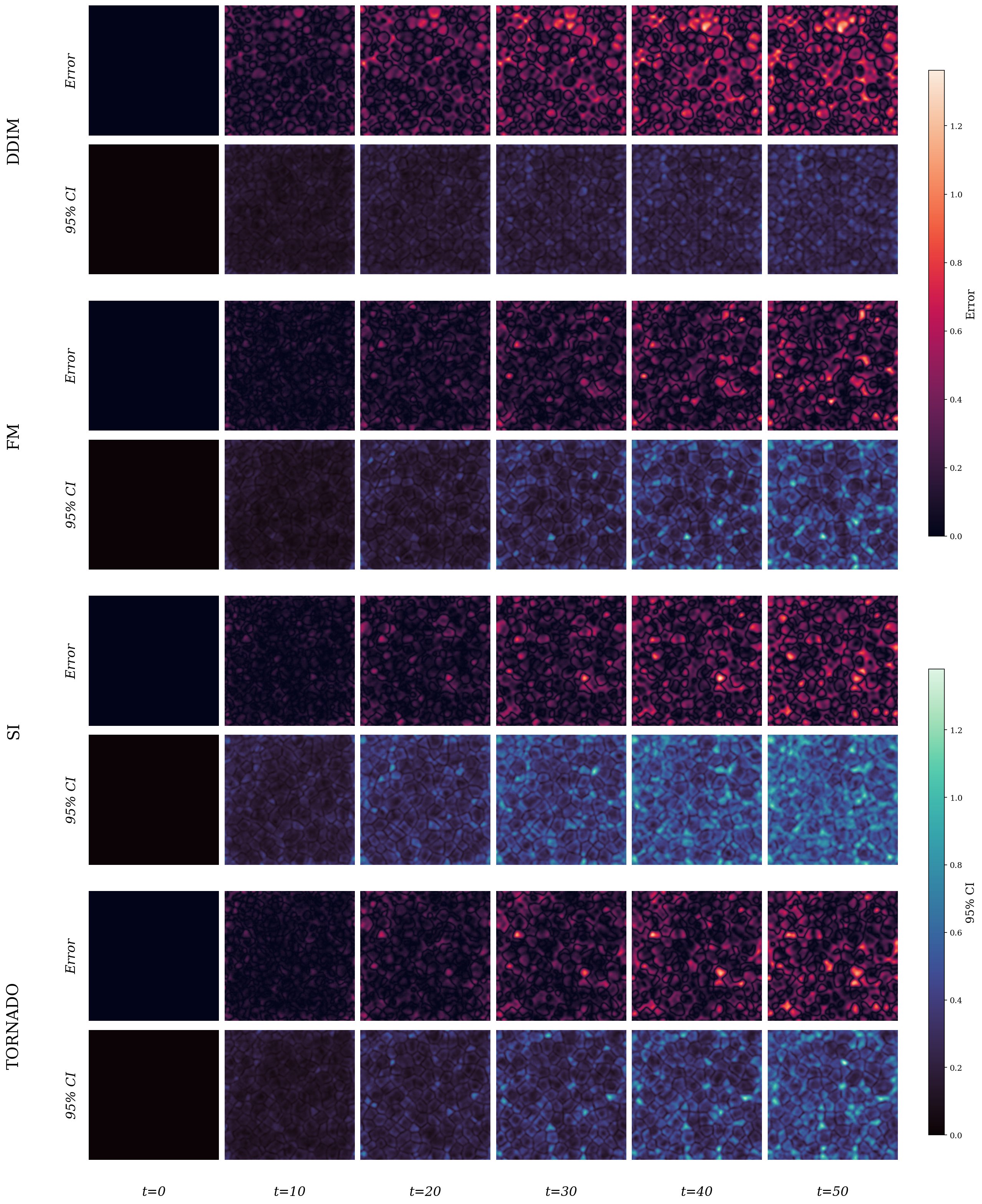}
    \caption{Visualization of the 95\% confidence interval of the predictive ensemble distribution at NFE=5 for methods TORNADO, FM, DDIM and SI alongside the RMSE of the mean of the predictive ensemble distribution with respect to the ground truth for the KS data set.}
    \label{fig:viz-ks-uq}
\end{figure}

\begin{figure}[hp]
    \centering
    \includegraphics[width=\linewidth]{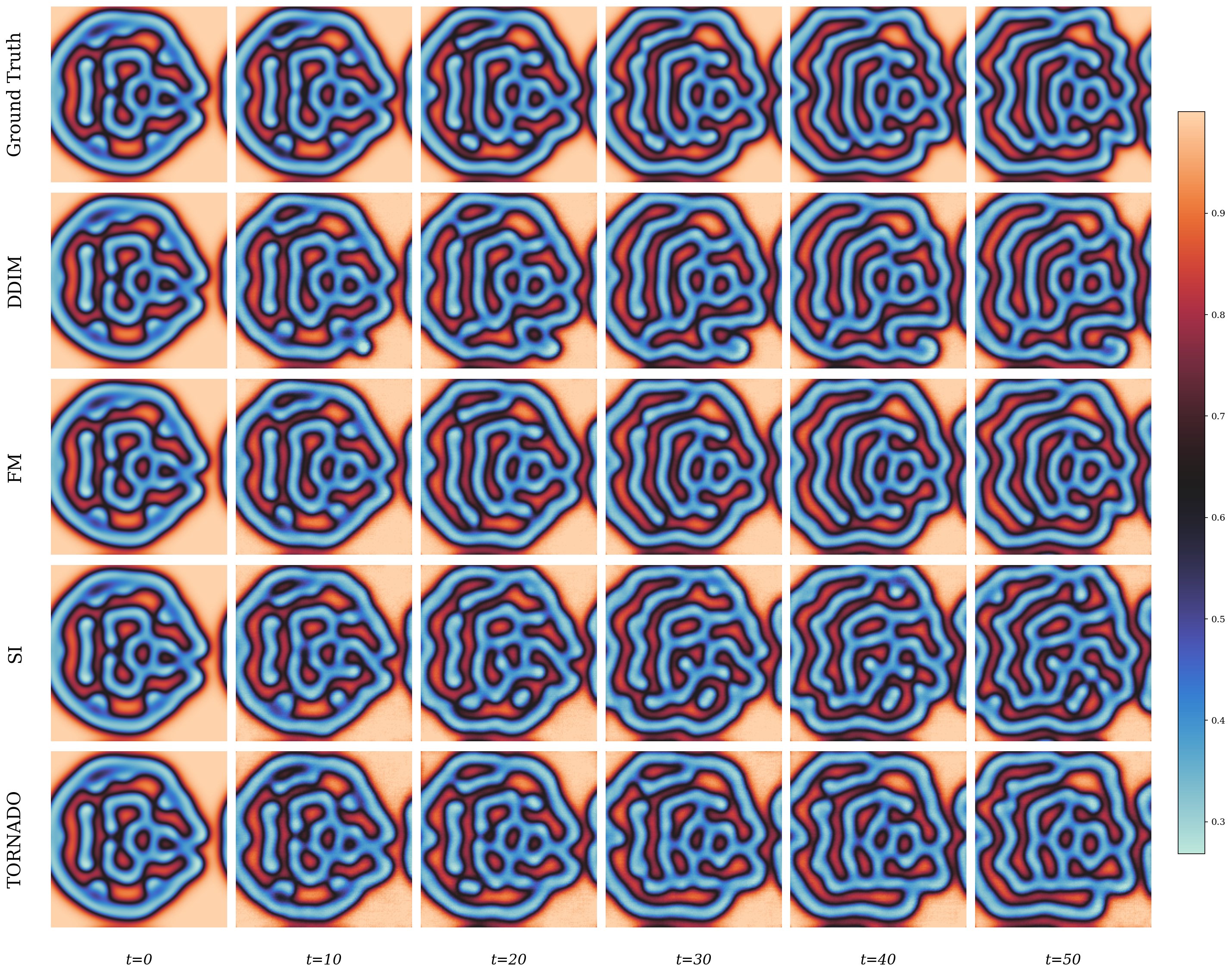}
    \caption{Visualization of rollout predictions at NFE=5 for methods TORNADO, FM, DDIM and SI alongside the ground truth trajectory for the GS data set.}
    \label{fig:viz-gs}
\end{figure}

\begin{figure}[hp]
    \centering
    \includegraphics[width=\linewidth]{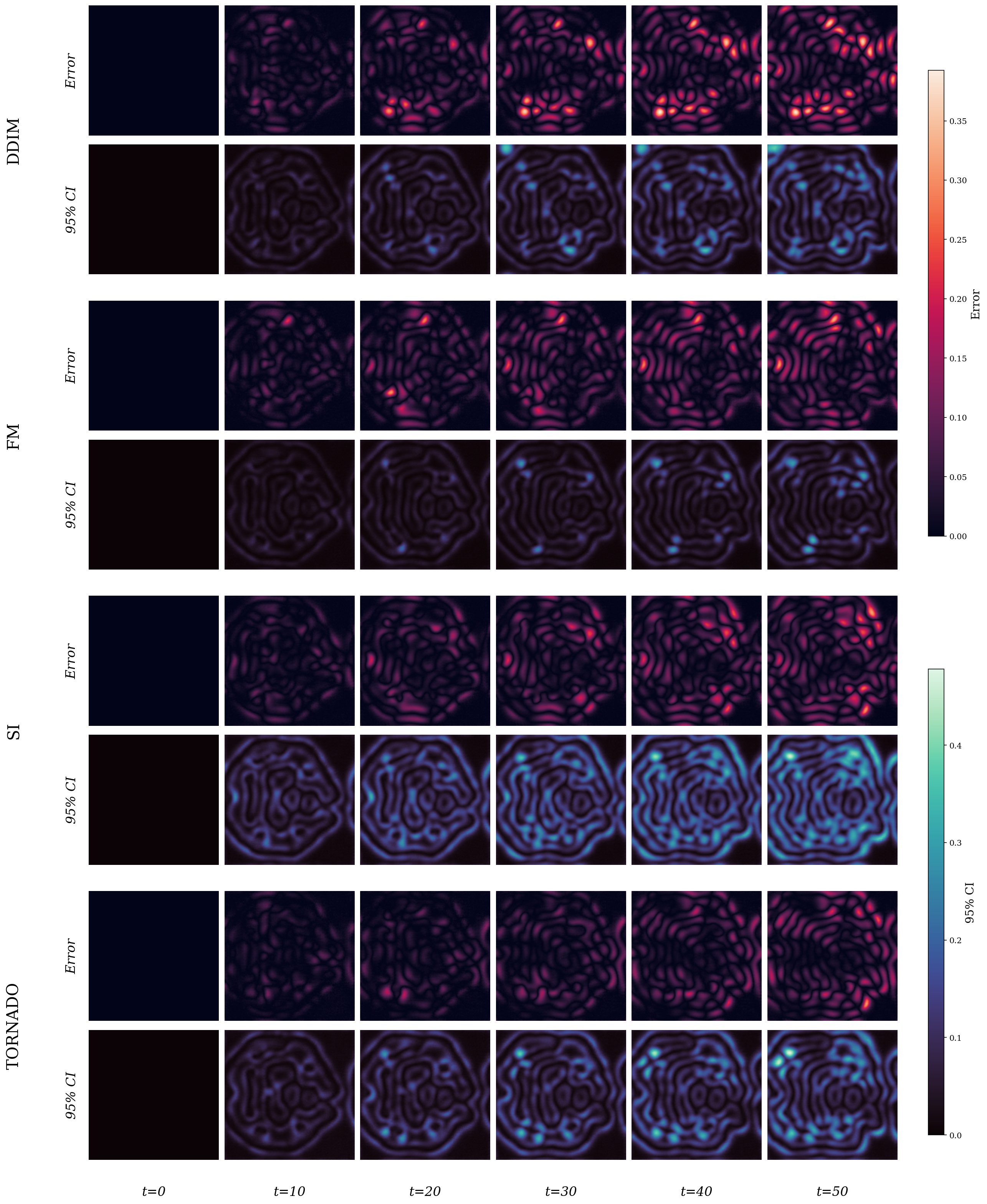}
    \caption{Visualization of the 95\% confidence interval of the predictive ensemble distribution at NFE=5 for methods TORNADO, FM, DDIM and SI alongside the RMSE of the mean of the predictive ensemble distribution with respect to the ground truth for the GS data set.}
    \label{fig:viz-gs-uq}
\end{figure}

\begin{figure}
    \centering
    \includegraphics[width=\linewidth]{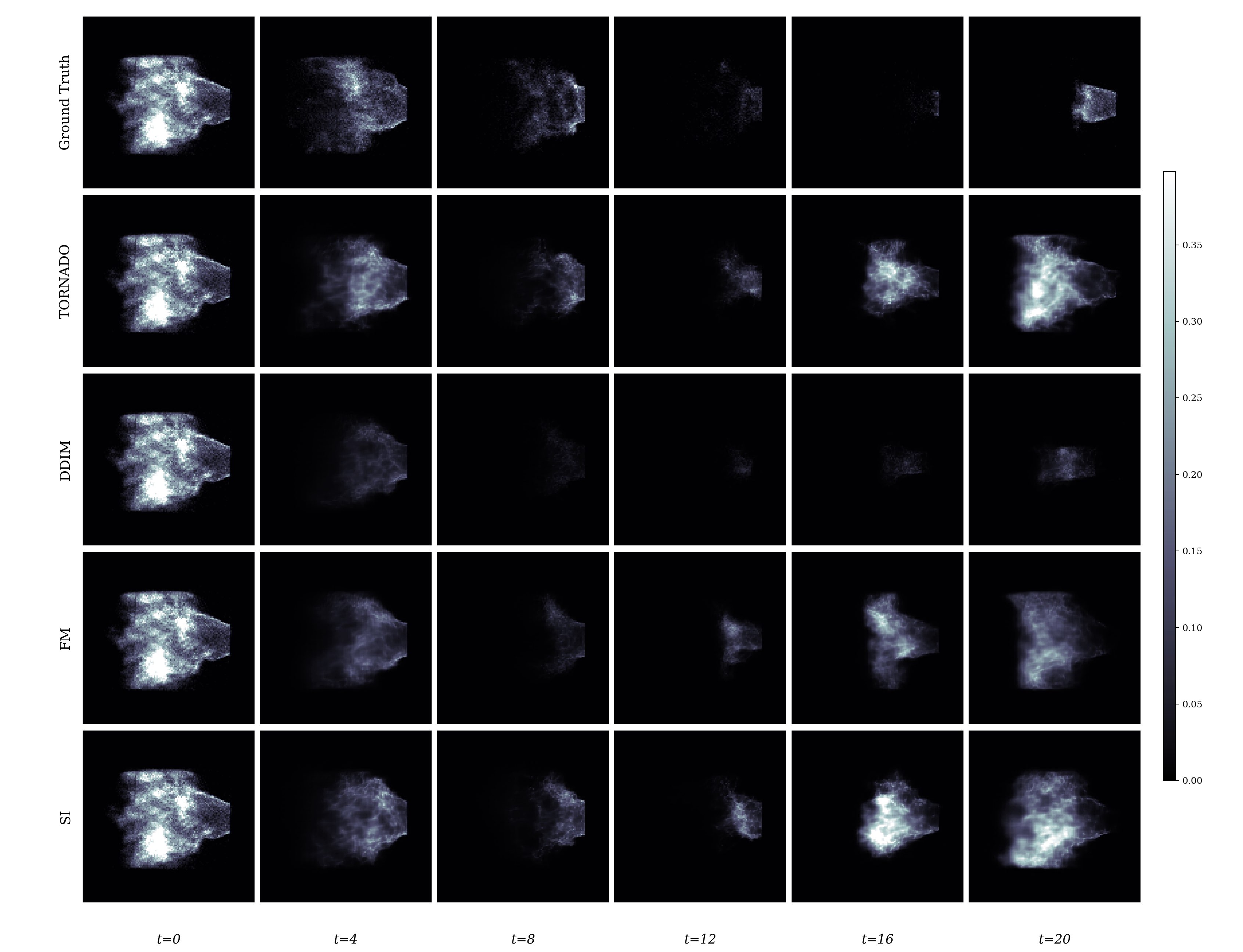}
    \caption{Visualization of rollout predictions at NFE=5 for methods TORNADO, FM, DDIM and SI alongside the ground truth trajectory for the RealPDEBench Combustion data set.}
    \label{fig:viz-com}
\end{figure}

\begin{figure}[hp]
    \centering
    \includegraphics[width=\linewidth]{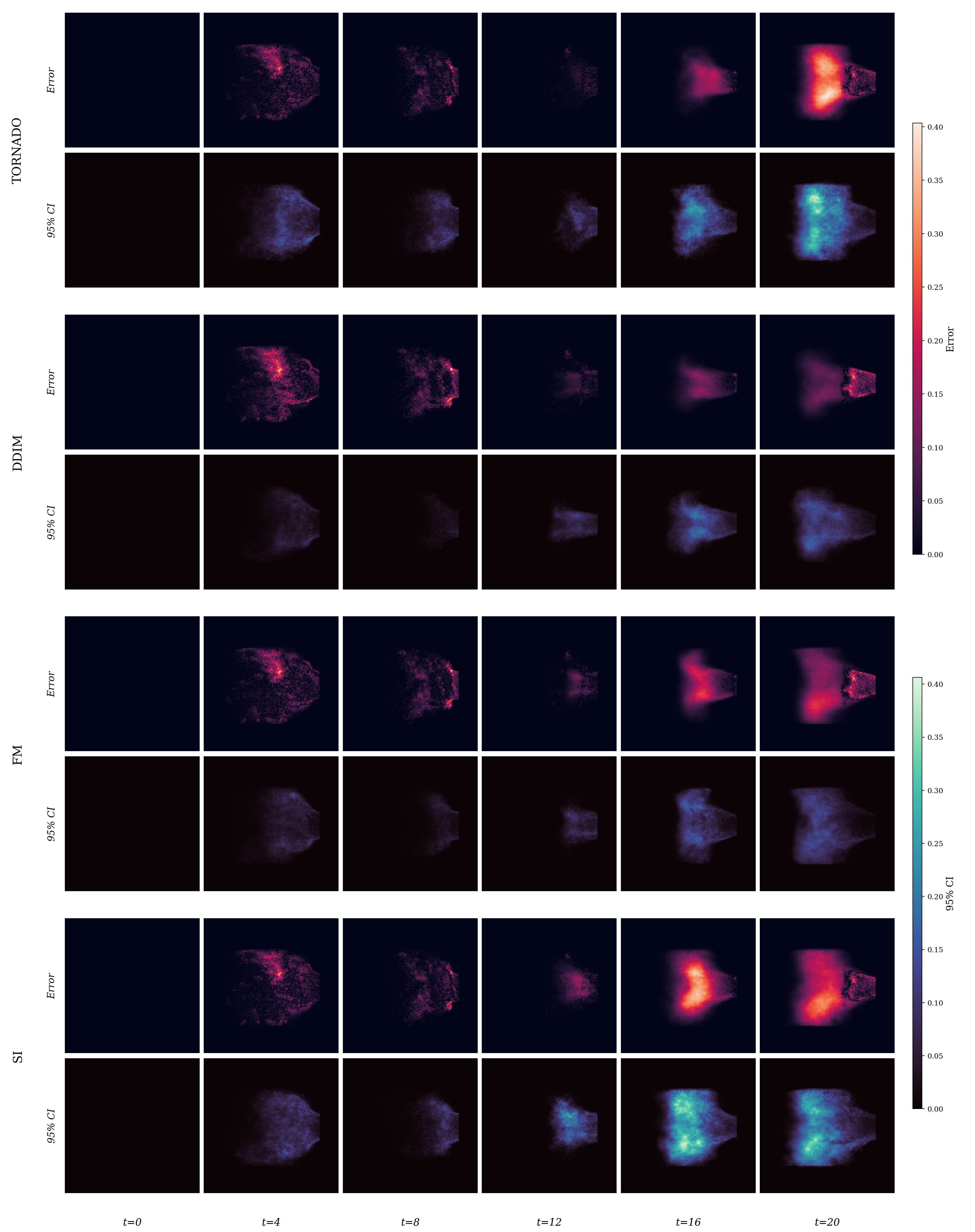}
    \caption{Visualization of the 95\% confidence interval of the predictive ensemble distribution at NFE=5 for methods TORNADO, FM, DDIM and SI alongside the RMSE of the mean of the predictive ensemble distribution with respect to the ground truth for the  RealPDEBench Combustion data set.}
    \label{fig:viz-com-uq}
\end{figure}

\begin{figure}
    \centering
    \includegraphics[width=\linewidth]{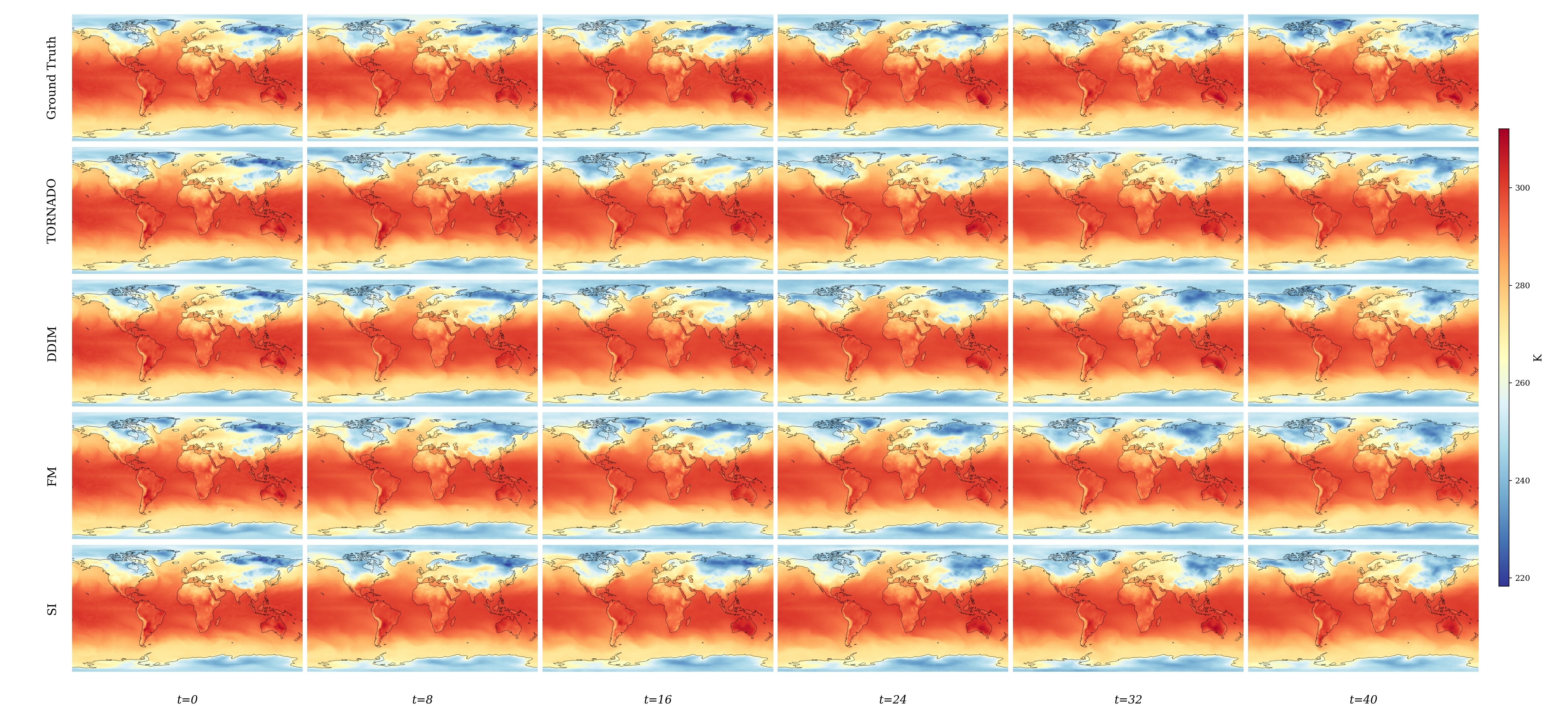}
    \caption{Visualization of rollout predictions at NFE=5 for methods TORNADO, FM, DDIM and SI alongside the ground truth trajectory for the $T_{2m}$ variable WeatherBench ERA5 data set.}
    \label{fig:viz-era5}
\end{figure}

\begin{figure}[hp]
    \centering
    \includegraphics[width=\linewidth]{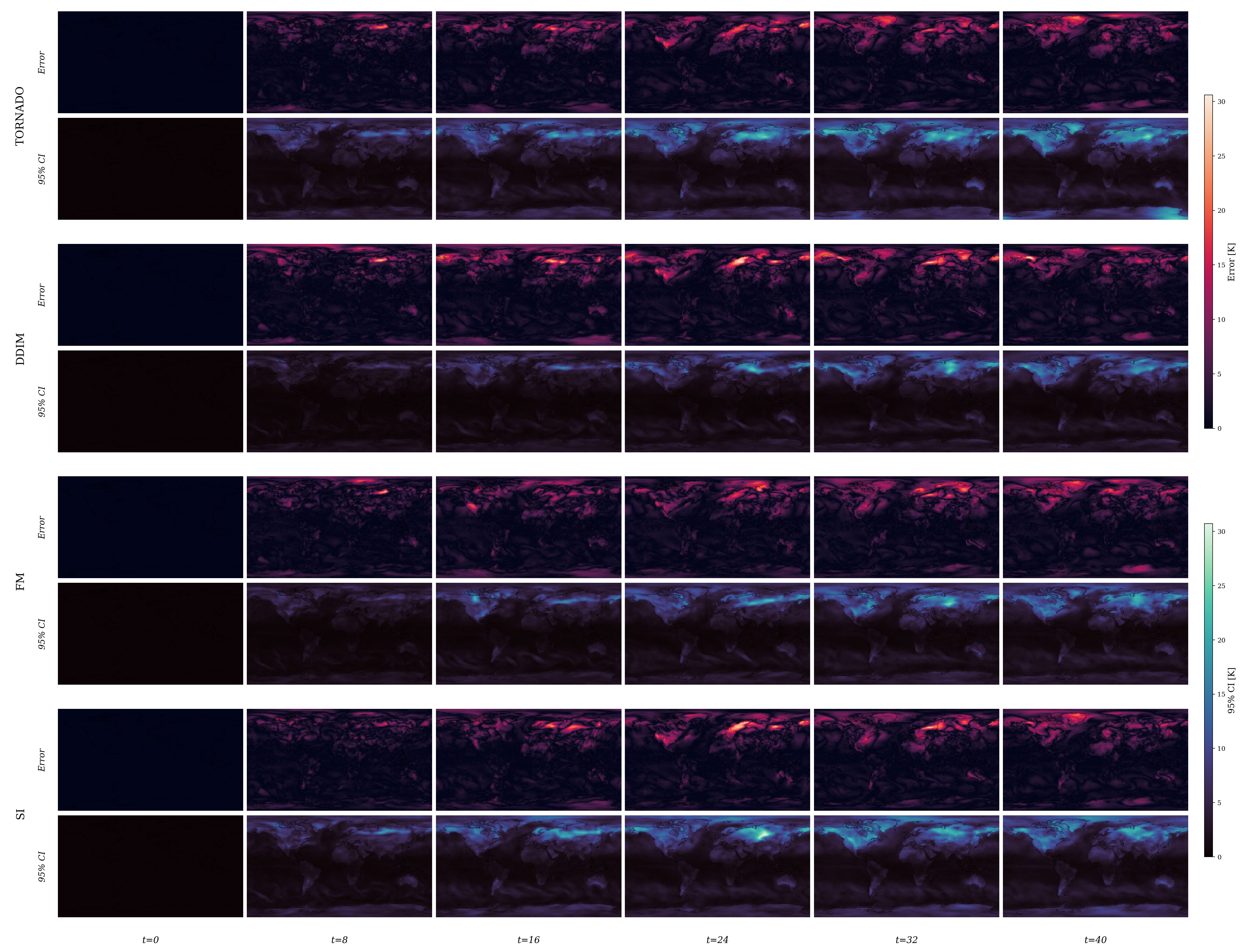}
    \caption{Visualization of the 95\% confidence interval of the predictive ensemble distribution at NFE=5 for methods TORNADO, FM, DDIM and SI alongside the RMSE of the mean of the predictive ensemble distribution with respect to the ground truth for the  $T_{2m}$ variable WeatherBench ERA5 data set.}
    \label{fig:viz-era5-uq}
\end{figure}

\begin{figure}
    \centering
    \includegraphics[width=\linewidth]{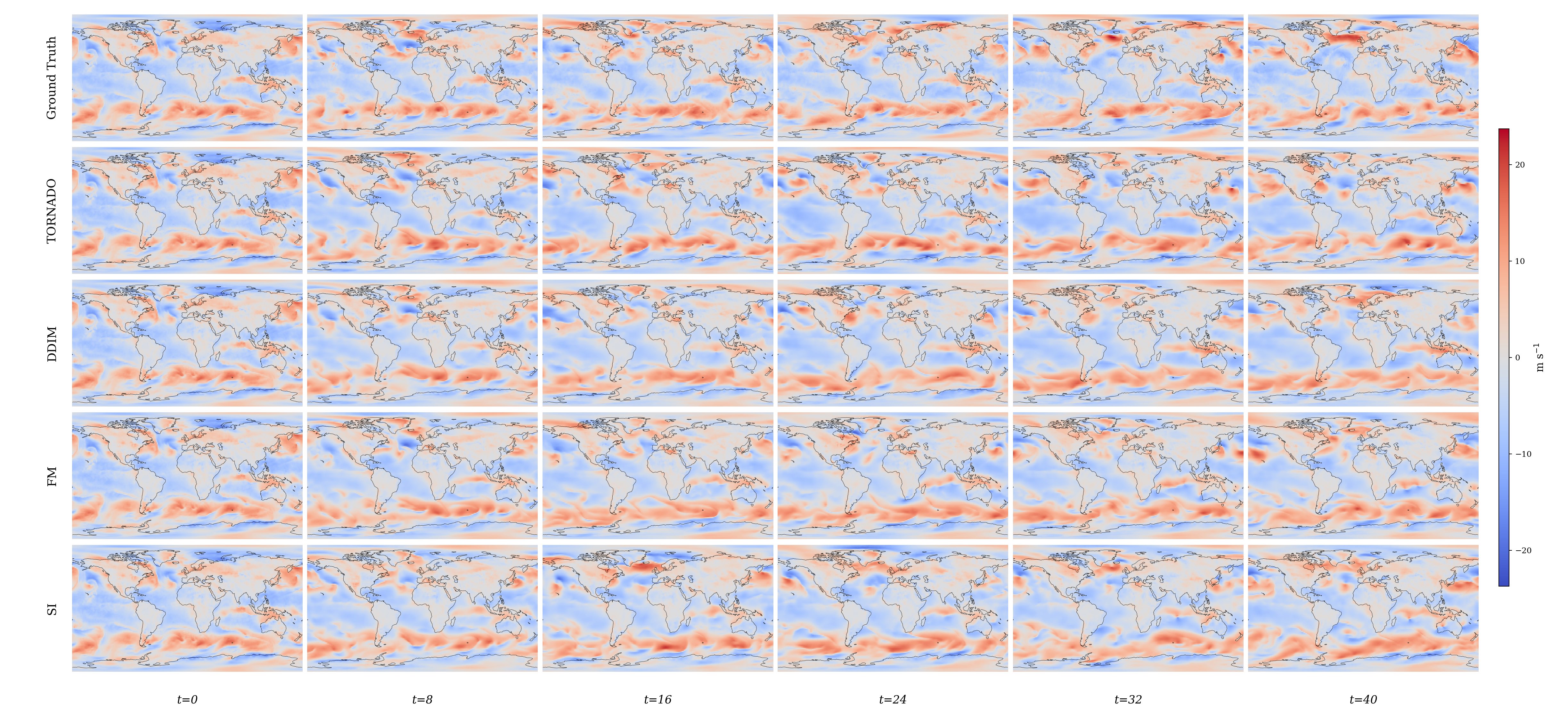}
    \caption{Visualization of rollout predictions at NFE=5 for methods TORNADO, FM, DDIM and SI alongside the ground truth trajectory for the $U_{10}$ variable WeatherBench ERA5 data set.}
    \label{fig:viz-era5-1}
\end{figure}

\begin{figure}[hp]
    \centering
    \includegraphics[width=\linewidth]{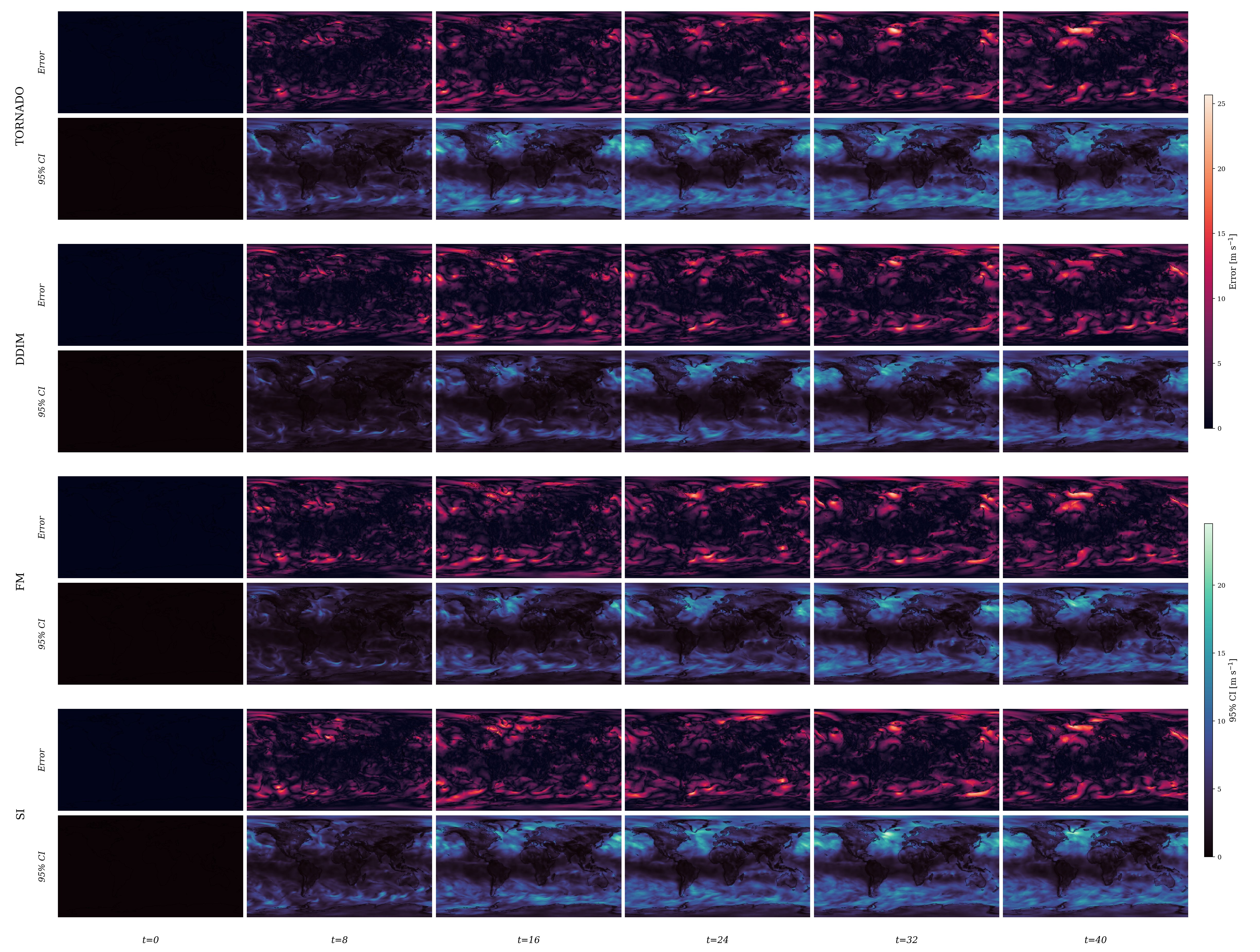}
    \caption{Visualization of the 95\% confidence interval of the predictive ensemble distribution at NFE=5 for methods TORNADO, FM, DDIM and SI alongside the RMSE of the mean of the predictive ensemble distribution with respect to the ground truth for the  $U_{10}$ variable WeatherBench ERA5 data set.}
    \label{fig:viz-era5-uq1}
\end{figure}

\begin{figure}
    \centering
    \includegraphics[width=\linewidth]{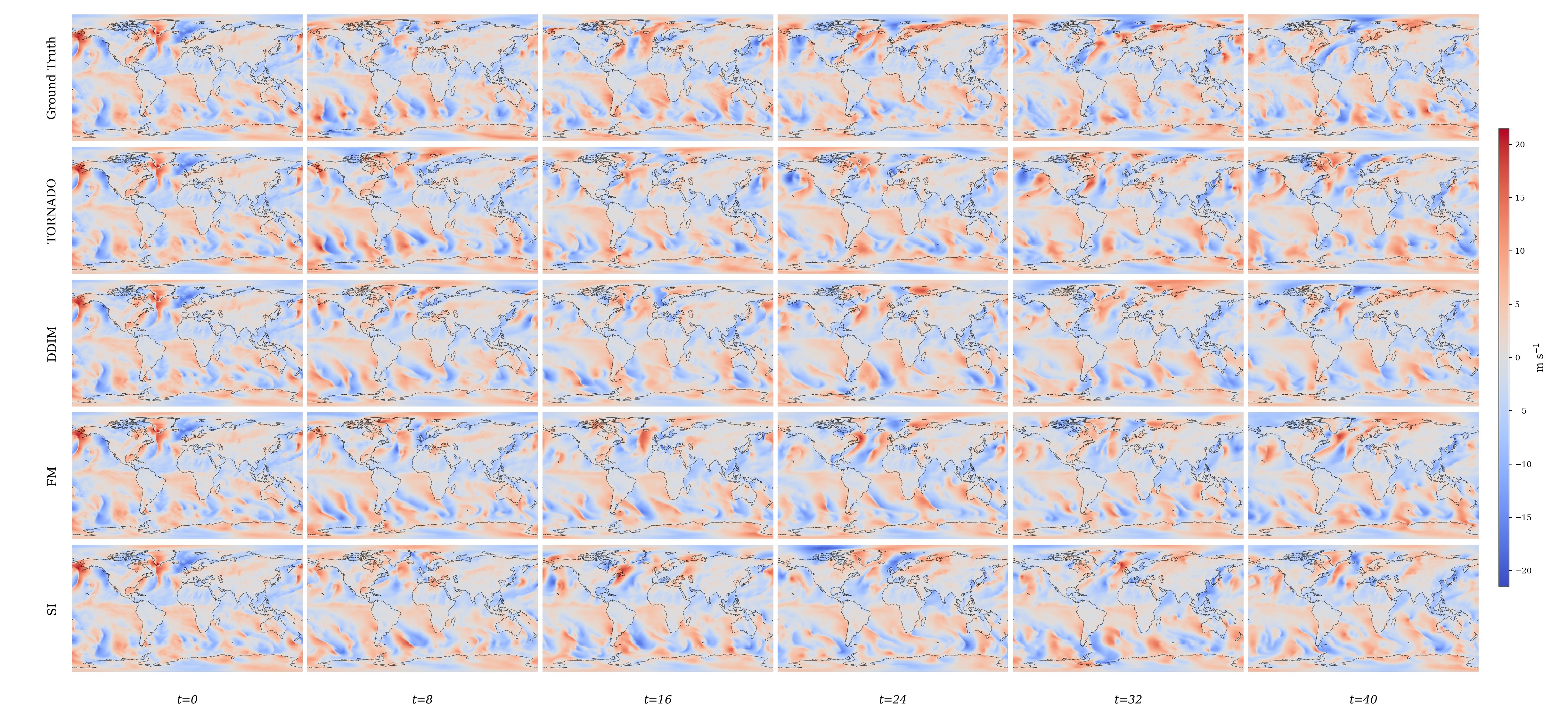}
    \caption{Visualization of rollout predictions at NFE=5 for methods TORNADO, FM, DDIM and SI alongside the ground truth trajectory for the $V_{10}$ variable WeatherBench ERA5 data set.}
    \label{fig:viz-era5-2}
\end{figure}

\begin{figure}[hp]
    \centering
    \includegraphics[width=\linewidth]{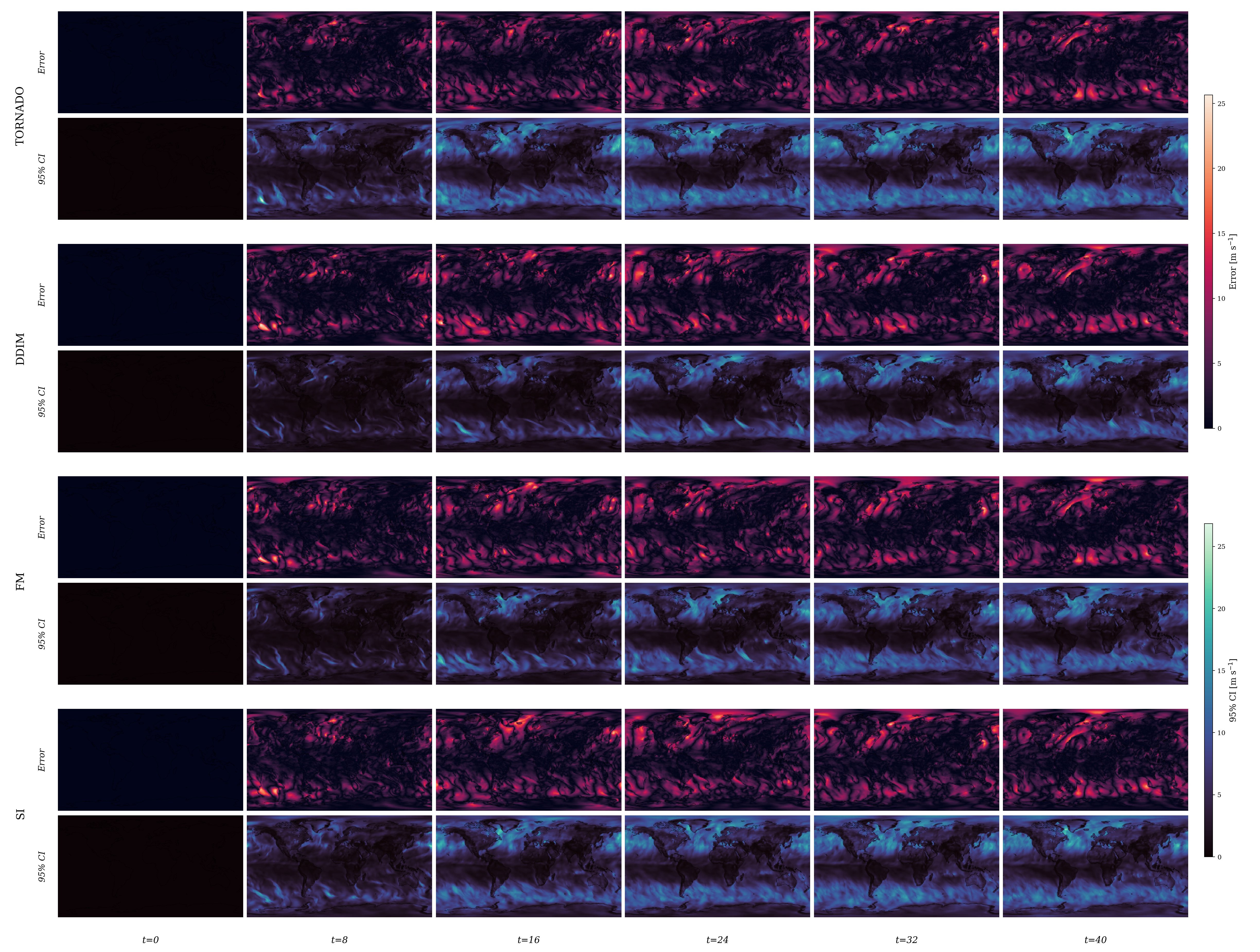}
    \caption{Visualization of the 95\% confidence interval of the predictive ensemble distribution at NFE=5 for methods TORNADO, FM, DDIM and SI alongside the RMSE of the mean of the predictive ensemble distribution with respect to the ground truth for the  $V_{10}$ variable WeatherBench ERA5 data set.}
    \label{fig:viz-era5-uq2}
\end{figure}

\begin{figure}
    \centering
    \includegraphics[width=\linewidth]{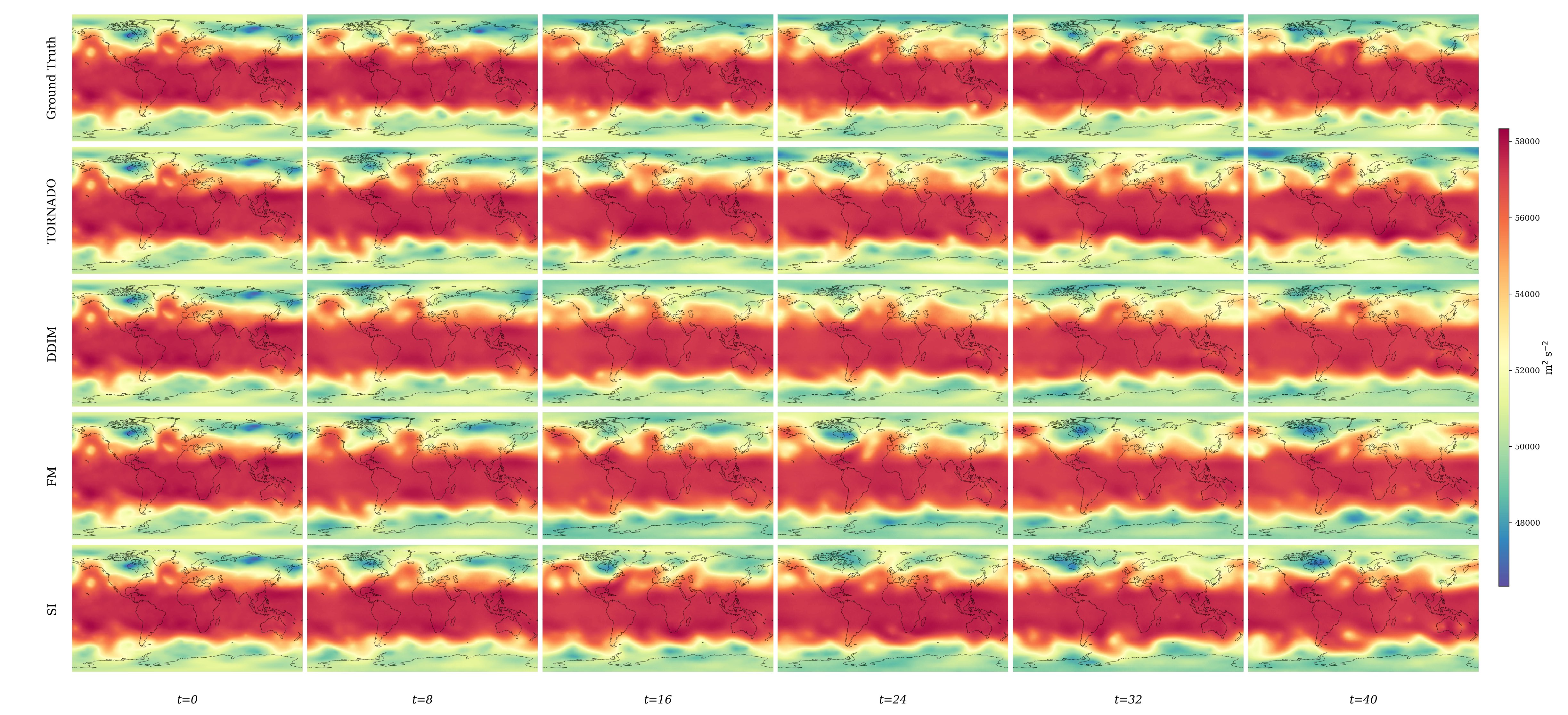}
    \caption{Visualization of rollout predictions at NFE=5 for methods TORNADO, FM, DDIM and SI alongside the ground truth trajectory for the $Z_{500}$ variable WeatherBench ERA5 data set.}
    \label{fig:viz-era5-3}
\end{figure}

\begin{figure}[hp]
    \centering
    \includegraphics[width=\linewidth]{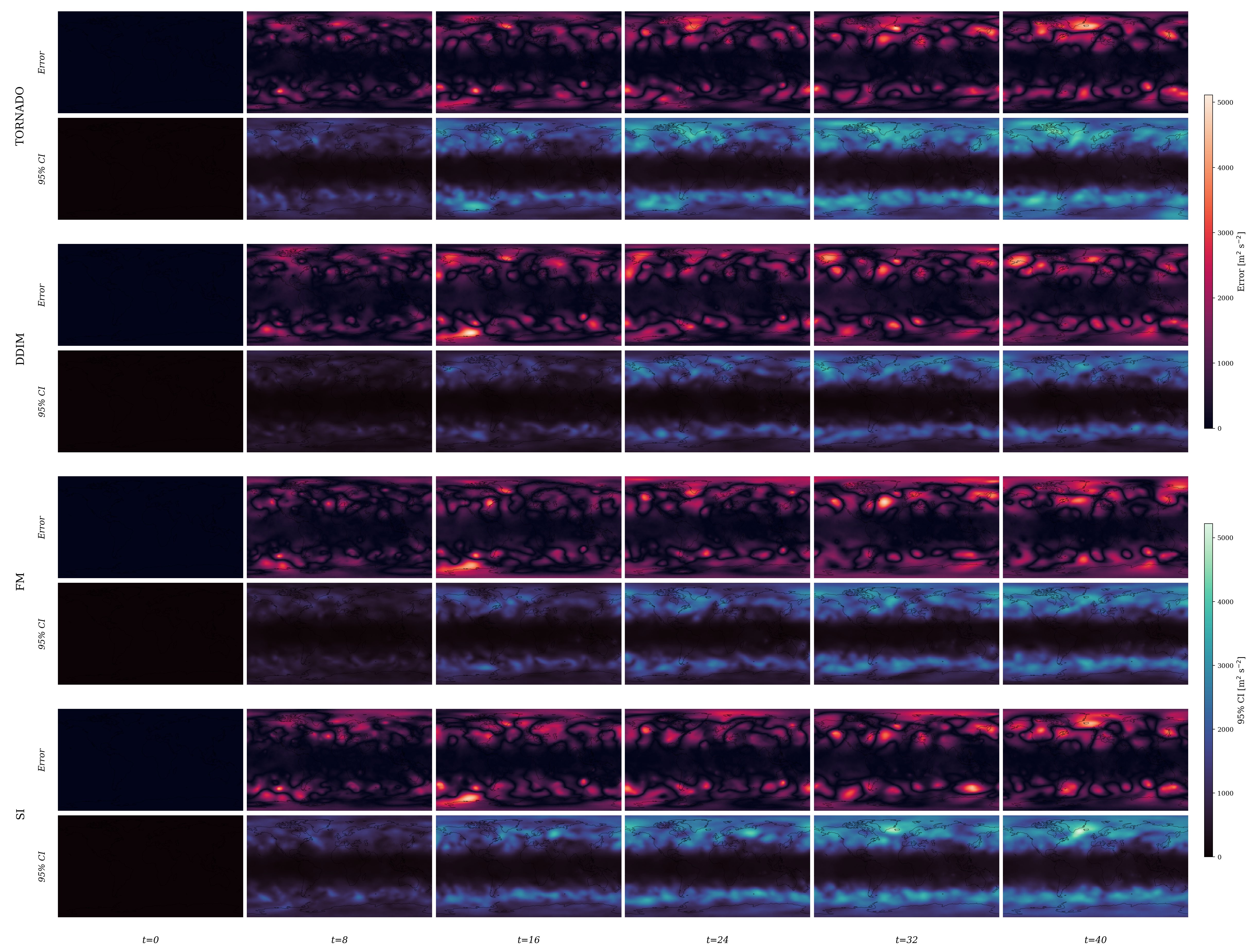}
    \caption{Visualization of the 95\% confidence interval of the predictive ensemble distribution at NFE=5 for methods TORNADO, FM, DDIM and SI alongside the RMSE of the mean of the predictive ensemble distribution with respect to the ground truth for the  $Z_{500}$ variable WeatherBench ERA5 data set.}
    \label{fig:viz-era5-uq3}
\end{figure}


\end{document}